\documentclass[aoas]{imsart}

\RequirePackage{amsthm,amsmath,amsfonts,amssymb}
\RequirePackage[authoryear]{natbib}
\usepackage{amssymb}
\usepackage{soul}
\usepackage{graphicx,epstopdf}
\usepackage{booktabs}
\usepackage{amsthm}
\usepackage{array,multirow}
\usepackage{tikz}
\usepackage{subfig}
\usepackage{caption}
\usepackage{multicol}
\usepackage{paralist}
\usepackage{url}
\usepackage{amsmath}
\usepackage{amsfonts}
\usepackage{color}
\usepackage{afterpage}
\usepackage{float}
\usepackage{soul}
\usepackage[boxruled, linesnumbered]{algorithm2e}
\SetKwInput{KwData}{Input}
\usepackage{hyperref}
\usepackage{verbatim}
\usepackage{array}
\usepackage{tabularx}
\usepackage{enumitem}
\usepackage{bm}
\usetikzlibrary{arrows.meta, positioning}

\newcommand{\V}[1]{\bm{\mathbf{\MakeLowercase{#1}}}} % vector
\theoremstyle{plain}
\theoremstyle{definition}
\newtheorem{theorem}{Theorem}
\newtheorem*{theorem*}{Theorem}
\newtheorem{prop}{Proposition}

\startlocaldefs
\endlocaldefs

\begin{document}

\begin{frontmatter}
\title{Uncertainty Quantification for Named Entity Recognition via Full-Sequence and Subsequence Conformal Prediction}
%\title{A sample article title with some additional note\thanksref{T1}}
\runtitle{UQ for NER}
%\thankstext{T1}{A sample of additional note to the title.}

\begin{aug}
%%%%%%%%%%%%%%%%%%%%%%%%%%%%%%%%%%%%%%%%%%%%%%%
%% Only one address is permitted per author. %%
%% Only division, organization and e-mail is %%
%% included in the address.                  %%
%% Additional information such as            %%
%% identifying the corresponding author must %%
%% be included in in the Acknowledgments     %%
%% section if necessary.                     %%
\author[A]{\fnms{Matthew}~\snm{Singer}\ead[label=e1]{mdsinger@ncsu.edu}},
\author[A]{\fnms{Srijan}~\snm{Sengupta}\ead[label=e2]{ssengup2@ncsu.edu}}
\and
\author[A,B]{\fnms{Karl}~\snm{Pazdernik}\ead[label=e3]{karl.pazdernik@pnnl.gov}}
%%%%%%%%%%%%%%%%%%%%%%%%%%%%%%%%%%%%%%%%%%%%%%
%% Addresses                                %%
%%%%%%%%%%%%%%%%%%%%%%%%%%%%%%%%%%%%%%%%%%%%%%
\address[A]{North Carolina State University \printead[presep={,\ }]{e1,e2}}

\address[B]{Pacific Northwest National Laboratory\printead[presep={,\ }]{e3}}
\end{aug}

\begin{abstract}
Named Entity Recognition (NER) serves as a foundational component in many natural language processing (NLP) pipelines. However, current NER models typically output a single predicted label sequence without any accompanying measure of uncertainty, leaving downstream applications vulnerable to cascading errors. In this paper, we introduce a general framework for adapting sequence-labeling-based NER models to produce uncertainty-aware prediction sets. These prediction sets are collections of full-sentence labelings that are guaranteed to contain the correct labeling with a user-specified confidence level. This approach serves a role analogous to confidence intervals in classical statistics by providing formal guarantees about the reliability of model predictions. Our method builds on conformal prediction, which offers finite-sample coverage guarantees under minimal assumptions. We design efficient nonconformity scoring functions to construct efficient, well-calibrated prediction sets that support both unconditional and class-conditional coverage. This framework accounts for heterogeneity across sentence length, language, entity type, and number of entities within a sentence. Empirical experiments on four NER models across three benchmark datasets demonstrate the broad applicability, validity, and efficiency of the proposed methods.
\end{abstract}


\end{frontmatter}

\section{Introduction}

Named entity recognition (NER) is a key task in natural language processing (NLP) that identifies spans of text corresponding to specific categories of real-world \textit{named entities} such as persons, locations, and organizations. NER underpins sophisticated NLP applications like named entity disambiguation (NED), information extraction, question answering, and text summarization \citep{yamada2016joint,molla2006named,nan2021entity}. For example, the sentence \textit{Sarah is from New York City} would ideally involve tagging ``Sarah'' as a person and ``New York City'' as a location, enabling downstream systems to link entities to a knowledge base.

Despite its importance, most NER systems provide a single label sequence without accounting for uncertainty, which risks error propagation in downstream tasks. For instance, if ``New York'' is incorrectly tagged as the location instead of ``New York City,'' a downstream named entity disambiguation (NED) system could introduce semantic inaccuracies by linking it to the state instead of the city. Such errors are especially problematic in high-stakes domains like biomedical text mining, legal document analysis, and automated knowledge base population. Popular frameworks such as sequence labeling \citep{ma2016end,lample2016neural}, substring classification \citep{zhong2020frustratingly}, and machine reading comprehension \citep{li2019unified} typically ignore uncertainty quantification altogether.

In this paper, we propose a novel uncertainty-aware framework for NER, leveraging conformal prediction to generate confidence-calibrated sets of plausible labels that guarantee finite-sample coverage. Using a conditional random field (CRF) model, we extend conformal prediction techniques to the sequence labeling formulation of NER, creating prediction sets over top-$k$ decoded label sequences. 
% For instance, instead of tagging ``New York'' alone, our method includes both ``New York'' and ``New York City'' as options, mitigating downstream errors.

Conformal prediction \citep{shafer2008tutorial,angelopoulos2023conformal} constructs prediction sets with probabilistic guarantees by estimating quantiles of non-conformity scores either transductively (leave-one-out validation) or inductively (held-out calibration sets). We adopt the inductive paradigm for computational efficiency while ensuring coverage. Our approach captures contextual dependencies among entity labels (e.g., ``if \textit{Sarah} is tagged as a person, then \textit{New York City} is likely a location'') and supports class-conditional calibration based on sentence length, language, and entity distribution.

The main contributions of this paper are as follows:
\begin{enumerate}
    \item \textbf{Full-Sequence Prediction Sets:}
    %\ref{sec:sent level}):}
    We construct prediction sets over complete label sequences that may contain multiple entities, capturing contextual relationships among co-occurring entities.
    To the best of our knowledge, this work provides the first method for constructing finite-sample valid prediction sets over entire NER label sequences.

    \item \textbf{Subsequence-Level Prediction Sets:}
    %\ref{sec:ent level}):}
    We define subsequence-level prediction sets for all possible subsequences that refer to a singular entity, enabling class-conditional coverage.

    \item \textbf{Integrated Prediction Sets:}
    %\ref{sec:sent+ent}):}
    We propose an integrated method that filters full-sequence predictions using entity-level prediction sets to produce full-sequence prediction sets that are well calibrated regardless of the class or number of entities within the full sequence. 
    
    \item \textbf{Covariate Calibration:} We demonstrate how both sentence length and sentence language impact the validity of conformal prediction sets in multilingual NER.  We then develop stratified conformal procedures that achieve valid coverage across language/length strata.
    
    \item \textbf{Combined Nonconformity score metric improvements:} Utilizing the previously mentioned stratified conformal prediction procedure, we demonstrate how multiple non-conformity scores may be combined to improve set efficiency. We propose two methods (Naive and Conditional) and demonstrate how a third combination method (RAPS, \cite{angelopoulos2020uncertainty}) compares and how that third method may be improved for NER.
\end{enumerate}

By combining deep sequence-labeling architectures with distribution-free uncertainty quantification, this work illustrates how formal statistical inference can strengthen the reliability and interpretability of AI systems operating in complex, structured domains, contributing to a deeper integration of the two fields: statistics and AI.

The rest of the paper is organized as follows. Section 2 reviews related work in NER and uncertainty quantification.
Section 3 introduces the sequence-labeling formulation of NER and describes the CRF-based model architecture used throughout the paper.
Section 4 develops conformal prediction methods for full-sequence labelings, including adaptive procedures and several baseline non-conformity scores.
Section 5 presents subsequence-level conformal prediction and establishes class-conditional calibration for individual entity spans.
Section 6 integrates the full-sequence and subsequence approaches, proposing unified prediction sets that account for contextual dependencies across entities and control family-wise error.
Section 7 reports empirical results across multiple benchmarks, models, and calibration strategies, including comparisons of non-conformity scores and covariate-stratified calibration.
Section 8 concludes the paper with a discussion.
In the interest of space, technical proofs and additional details/results from Sections 4-7 are in the supplementary file,  Sections S1-S4.

\section{Related Work and Overview\label{sec:rel_work}}
NER is a foundational task in NLP, originating from the Message Understanding Conferences (MUC) in the 1990s. Early NER systems relied on hand-crafted entity lists and rule-based grammar patterns.
This was followed by feed-forward neural networks in the early 2000s  \citep{hammerton2003named} and recurrent neural networks in the 2010s.
A major breakthrough occurred in 2011, when researchers incorporated CRFs into convolutional neural network (CNN) architectures \citep{collobert2011natural}. By 2016 CRFs became widely adopted as researchers observed their effectiveness when paired with context-aware architectures such as long short-term memory (LSTM) networks \citep{lample2016neural,ma2016end,chiu2016named,clark2018semi}. The CRF layer enabled models to learn structured prediction rules and significantly improved benchmark performance. Although CRF-based decision heads continue to be used in state-of-the-art systems, recent research has shifted toward more expressive embedding methods, reformulating NER as a non-sequential task\citep{fisch2022conformal,liu2022autoregressive}, incorporating external knowledge sources\citep{Wang2022DAMONLPAS}, and improving training methodologies\citep{zhou2021learning,conneau2019unsupervised}.

Uncertainty quantification (UQ) in NER remains a nascent topic. UQ seeks to quantify both aleatoric uncertainty (inherent variability in data) and epistemic uncertainty (stemming from model limitations). The increasing deployment of black-box machine learning models has driven a parallel rise in the importance of UQ techniques, as black-box models are unable to be understood/reasoned with and provide little to no statistical guarantees when compared to more principled statistical techniques. Prior UQ methods for NER include model calibration \citep{ liang2021calibrenet}, ensemble-based approaches \citep{ yang2024uncertainty,akkasi2016improving,he2023uncertainty}, Bayesian inference \citep{akkasi2016improving, maragoudakis2006dealing,he2023uncertainty}, among active learning and other statistical techniques \citep{nguyen2021loss, liu2022ltp, vazhentsev2022uncertainty}.
% \ssg{could you remove the survey-type papers above?}\ms{remvoed survey papers, included a few additional NER examples for each one along with some active learning papers relating to NER.}
Each of these methods has merits but also notable limitations in the NER context. Calibration methods aim to align predicted probabilities with empirical frequencies. Well-calibrated outputs can be used to generate prediction sets, but these methods are highly vulnerable to \textit{miscalibration} since calibration failures can compromise coverage guarantees. Ensemble methods rely on model diversity, such as bootstrapping, dropout, or varying seeds, to estimate uncertainty. However, they are computationally expensive and difficult to deploy at scale. Bayesian methods offer a theoretically sound alternative, but full Bayesian neural networks are challenging to train and scale. Outside of uncertainty quantification, the use of uncertainty of NER is a growing field of research with a myriad of different techniques and applications \cite{shorinwa2025survey,hu2023uncertainty,campos2024conformal}. For example, uncertainty may be utilized in auxiliary NER tasks such as model training \cite{nie2025improving}. Another example is \citet{zhang2024linkner} which performs NER via two models, a local span-identification model and an LLM-based classification model, with uncertainty-aware outputs from the span-identification model being an input to the second classification step.

Conformal prediction offers a statistically principled framework for UQ. It produces prediction sets with finite-sample validity under the exchangeability assumption, which is much weaker than the i.i.d. assumption. Conformal methods can be inductive or transductive. Inductive conformal prediction uses a calibration set to estimate score quantiles and is preferred for scalability. Transductive methods retrain the model for each test input and are more computationally demanding. To date, conformal prediction has seen limited application in NER.

The closest prior work is \citep{fisch2022conformal}, which constructs entity-level prediction sets in a span-classification framework. Their method does not model joint dependencies between entities within a sentence, which is fundamental to sequence-level labeling. Applying their approach to our setting would require combining multiple span-level prediction sets using conservative bounds (e.g., Bonferroni/union bounds), which leads to overly large and inefficient sentence-level sets. 
In contrast, our sentence-level CP formulation directly models and leverages the contextual dependence structure across the 
 entities in a sentence (e.g., person-location co-occurrence patterns). This is a fundamentally different strategy which enables principled and efficient prediction sets for structured NER outputs.

\section{Named Entity Recognition}
\label{sec:NER}
 Given a dataset of $N$ labeled samples $\mathcal{D} = \{(\mathbf{w}_i, \mathbf{y}_i)\}_{i=1}^{N}$, each observation $\mathbf{w}_i = (w_{i,1}, w_{i,2}, \ldots, w_{i,t_i})$ denotes the observed input word sequence (eg: a sentence or paragraph) of length $t_i$ and $\mathbf{y}_i = (y_{i,1}, y_{i,2}, \ldots, y_{i,t_i})$ the associated sequence of class labels, where $y_{i,j} \in \mathcal{L}$. The label space is defined as $\mathcal{L} = \{l_{0}, l_{1}, \ldots, l_{c}, l_{\text{start}}, l_{\text{end}}\}$, consisting of $c$ named entity types along with the non-entity label of $l_0$ and special start and stop indicators of $l_{start}$ and $l_{stop}$. Given that the original input sequence length is from one to $t_i$, we augment $\mathbf{y}_i$ to include $y_{i,0} = l_{\text{start}}$ and $y_{i,t_i+1} = l_{\text{end}}$. The goal of the NER model is to learn a mapping $\mathbf{w} \mapsto \mathbf{y}$. Given $|\mathcal{L}|$ possible class labels and input sequences of varying lengths $t_i$, the space of all possible labelings is $|\mathcal{L}|^{t_i}$. An NER model could be a two-step process where each word is transformed into a vector representation before being fed into the entity classifier model, in such cases $\mathbf{x}_i = (x_{i,1}, x_{i,2}, \ldots, x_{i,t_i})$ represents corresponding vector representations of each word (e.g., contextual embeddings) such that $x_{i,j} \in \mathbb{R}^d$.

Because entities may span multiple words, the model may use the inside-outside-beginning (IOB2) tagging scheme to define entity boundaries. Under this scheme, each label includes a prefix indicating its position within an entity span (\texttt{B} for beginning, \texttt{I} for inside, and \texttt{O} for outside) and a class label. For example, a four-class NER system identifying \texttt{PER}, \texttt{LOC}, \texttt{ORG}, and \texttt{MISC} would define the label space as
\begin{equation*}
    \mathcal{L} 
    = \{\texttt{O}, \texttt{B-PER}, \texttt{I-PER}, \texttt{B-LOC}, \texttt{I-LOC}, \texttt{B-ORG}, \texttt{I-ORG},  \texttt{B-MISC}, \texttt{I-MISC}, \texttt{START}, \texttt{STOP} \}.
\end{equation*}

NLP models often break words down into multiple subword tokens. For the sake of simplicity in notation, we refer to $w_i$ as the $i^{th}$ `word' in the input word sequence and do not recognize any subword tokens. The techniques developed in this paper do, however, work when utilizing subword tokens if an  aggregation step is performed at the end of the predictive model, which resolves any conflicts between two subword labelings. Table~\ref{tab:NER_ex} presents an example sentence annotated using the IOB2 scheme.

\begin{table}[t]
\centering
\begin{tabular}{c|cccccc}
\textbf{Word}  & Sarah & is & from  & New   & York & City \\ \hline
\textbf{Label} & B-PER & O & O & B-LOC & I-LOC & I-LOC
\end{tabular}
\caption{IOB2 NER labels for an English sentence.}
\label{tab:NER_ex}
\end{table}

As discussed earlier, the NER task can be approached using either single-entity or multi-entity methods. In single-entity approaches \citep{zhong2020frustratingly}, the model is provided with one subsequence of words from the input sentence at a time and must determine whether that subsequence corresponds to a named entity class. This procedure is repeated across all possible subsequences up to a specified maximum length, yielding a disjoint set of predictions. In contrast, multi-entity methods predict all entity spans jointly, enabling the model to capture interdependencies between entities. For instance, in the sentence “Sarah is from New York City,” the phrase “is from” often signals a relationship between a person and a location or organization. By exploiting such contextual cues, multi-entity models generally outperform single-entity approaches on state-of-the-art benchmarks. For this reason, this work adopts a multi-entity prediction framework known as a CRF.

\subsection{Conditional Random Fields}
\label{sec:CRF}

\begin{figure}[t]
  \centering
  \includegraphics[width=.8\linewidth]{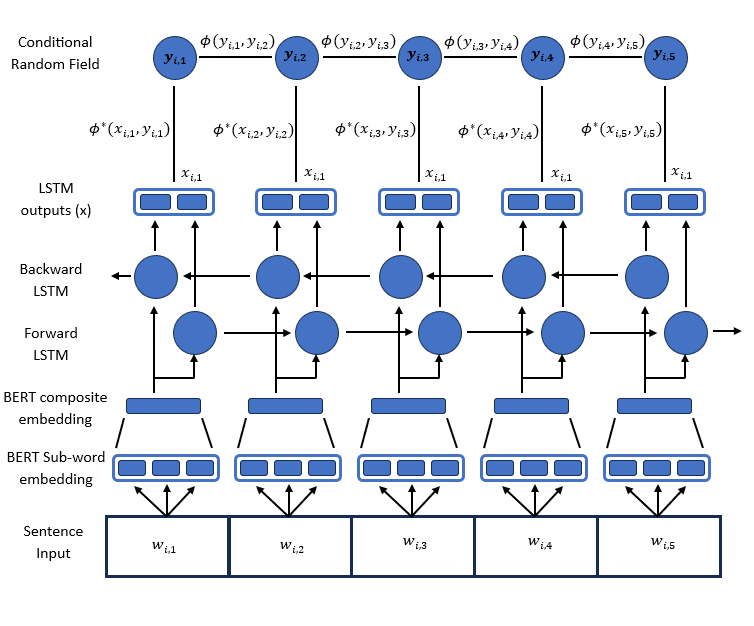}
  \caption{NER BiLSTM-CRF model architecture. $w_{i,1}$, $x_{i,1}$, and $y_{i,1}$ denote the $j^{th}$ word, word-embedding, and predicted label of the $i^{th} $ observation, respectively. CRF transmission and emission feature functions $\phi(Y_{j-1}, Y_j)$ and $\phi^*(X_j, Y_j)$ are defined in Section~\ref{sec:CRF}.}
  \label{fig:ner_diagram}
\end{figure}

The model architecture adopted in this work is a variant of the BERT-BiLSTM-CRF framework~\citep{tedeschi2021wikineural}. A CRF-based model was selected over more complex alternatives such as encoder-decoder architectures, large language models (LLMs), or other multi-entity frameworks due to its strong performance on state-of-the-art NER tasks \citep{akbik2018contextual, wang2014supervised, hu2024deep} and the interpretability of its decision space\citep{agarwal2021interpretability}\footnote{The increased `interpretability' of CRFs is derived from the ability to observe the relationships among the input word embeddings and prediction class labels via the manually specified transmission and emission feature functions.}. 
The structured output of the CRF makes it particularly well-suited for uncertainty quantification. An overview of the adopted CRF architecture is provided in Figure~\ref{fig:ner_diagram}.

The CRF decision head depicted in Figure \ref{fig:ner_diagram} is a type of Markov random field (MRF), where each label, conditioned on the input sequence, satisfies the Markov property with respect to its neighboring labels. For the following equations, the index of the observation is irrelevant and therefore the subscript $i$ is dropped from the notation as described in section \ref{sec:NER}. Following the adapted notation, the conditional independence assumption is expressed as
\begin{equation}
    \mathbb{P}(y_j \mid x_j, \{y_k : k \neq j\}) = \mathbb{P}(y_j \mid x_j, y_{j-1}, y_{j+1}),
\end{equation}
indicating that the probability of label $y_j$ depends only on the numerical representation of the $j^{th}$ word, $x_j$,  and its immediate neighbors $y_{j-1}$ and $y_{j+1}$. In general, inference on full CRFs is NP-hard and may significantly increase the computational cost of a model \citep{cuong2014conditional}. To decrease the computational overhead, in this work, we use a linear-chain CRF with two classes of feature functions that model first-order emission and transition probabilities.

Under the CRF model, the probability of a label subsequence $\mathbf{y}_{a:a+b} = (y_a, y_{a+1}, \ldots, y_{a+b})$  given the input sequence of embedded words $\mathbf{x}_{a-1:a+b+1}$ is modeled as

\begin{align}
   & \mathbb{P}(\mathbf{y}_{a:a+b} \mid \mathbf{x}_{a-1:a+b+1}; y_{a-1}, y_{a+b+1}) \nonumber
   \\
    = 
&\frac{
\phi^*(x_{a-1}, y_{a-1}) \prod_{j=a}^{a+b+1} \phi(y_{j-1}, y_j) \phi^*(x_j, y_j)
}{
\sum_{\mathbf{z} \in \mathcal{L}^m} \phi^*(x_{a-1},z_{a-1}) \prod_{j=a}^{a+b+1} \phi(y_{j-1}, z_j) \phi^*(x_j, z_j)},
\label{eq:base_CRF}
\end{align}

where $\mathcal{L}^b$ is the set of all possible labeling sequences of size b, the model probability described in equation \eqref{eq:base_CRF} is similar to that of a softmax function, where the numerator is the function that denotes the `value' of the given subsequence. The denominator of the probability function denotes the sum of the values of all possible subsequences. The score functions described in this probability statement are related to the set of transmission feature functions ($\phi(y_{j-1}, y_j)$), which denote how probable a label is given its neighboring labels, and the emission feature functions ($\phi^*(x_{a-1}, y_{a-1})$), which denote how probable a label is given the related input vector.
Each feature function is an exponential family parameterized by a weight vector $\lambda$ or $\lambda^*$, yielding:

\begin{equation}\label{eq:score1}
\phi^*(x_j, y_j) = \exp\left( \sum_{k=1}^{e^*} \lambda^*_{k} f^*_{k}(x_j, y_j) \right) = \exp\left( \mathbf{W}^*(y_j, :) \cdot x_j \right)
\end{equation}

\begin{equation}\label{eq:score2}
\phi(y_{j-1}, y_j) = \exp\left( \sum_{k=1}^{e} \lambda_k f_k(y_{j-1}, y_j) \right) = \exp\left( \mathbf{W}(y_{j-1}, y_j) \right)
\end{equation}

Functions $f^*(x_j,y_j)$ and $f(y_{j-1},y_j)$ are any real-valued differentiable functions that take as input either the input vector/label pair or two label/label pairs, respectively. For our application, the feature functions simple single-feature matrices \citep{ma2016end}. where $\mathbf{W}^*(y_j, :) \cdot x_j$ depicts the dot product of the input vector $x_j$ and the emission feature matrix $\mathbf{W}^* \in \mathbb{R}^{|\mathcal{L}| \times d}$. Similarly $\mathbf{W}(y_{j-1}, y_j)$ depicts the transmission feature matrix values corresponding by the labels of $y_{j-1}$ and $y_j$.

\subsection{CRF Decoding via beam search}
Identifying the most probable label sequence is itself a non-trivial task due to the exponential size of the label space (e.g., $|\mathcal{L}|^b$ for a sequence of length $b$).
To address this, CRF models employ the beam search decoding algorithm \citep{meister2020if}.

Beam search begins at the designated \texttt{START} token ($y_0)$ and determines the top-K best transmissions to the next label at time step one ($y_1$) by maximizing the sum of emission and transmission scores from the feature functions depicted in equations \eqref{eq:score1} and \eqref{eq:score2}, i.e., $\mathbf{W}^*(y_1)\cdot x_{y_1} + \mathbf{W}(y_0, y_1)$. Once the top-K transmissions have been determined for time step one, the algorithm evaluates the next $K$ most probable continuations at time step two, resulting in up to $K^2$ sequences. This expanded set is pruned back to the top $K$ sequences, and the process repeats until an \texttt{END} token is reached. If K is larger than the size of the CRF's state space, this process will end in the set of the top-K most probable sequences, of which the sequence with the highest score is the output of the NER model. In this paper, the K values chosen for the beam search algorithm also determine the maximum size of a prediction set before the set of all possible sequences is returned (guaranteeing 100\% accuracy). Because we do not wish to limit the size of our prediction sets to a small number of outputs before returning the full set, we utilize a K value of 100.

\section{Conformal Prediction}
\label{sec:conformal}

Let us define a desired confidence level of $1 - \alpha \in (0,1)$, and consider a dataset of $N$ labeled observations $\mathcal{D} = \{( \mathbf{w}_i, \mathbf{y}_i)\}_{i=1}^N$, as introduced in Section~\ref{sec:NER}. 
{Following the notation in section~\ref{sec:NER}, when defining conformal prediction, we will often suppress the observation index $i$; therefore, each word $j$ in sentence $i$ may be written as} $w_{i,j} = w_j$. This word is mapped to a vector representation $x_{i,j} = x_j \in \mathbb{R}^d$. This vector representation is the output of the Bi-LSTM into the CRF-layer as depicted in Figure \ref{fig:ner_diagram}. The dataset is partitioned into a training set $\mathcal{I}_{\text{train}}$ and a calibration set $\mathcal{I}_{\text{cal}}$.

A CRF model is trained using $\mathcal{I}_{\text{train}}$ to obtain the estimated probability function described previously in Equation~\ref{eq:base_CRF}. 
Let $nc(\mathbf{y} \mid \mathbf{x})$ denote a generic non-conformity score such that higher values of this function correspond to greater non-conformity between the value y and the expected value of y given x from the model. The conformal prediction objective is to determine a threshold $\tau$ such that for a new observation $(\V{x}_{new},\V{y}_{new})$ the conformal prediction set $C(\mathbf{x}_{\text{new}}, \tau)$ satisfies

\begin{equation}\label{base_conformal guarantee}
    \mathbb{P}\left(\mathbf{y}_{\text{new}} \in C(\mathbf{x}_{\text{new}}, \tau)\right) \geq 1 - \alpha.
\end{equation}

The conformal prediction set $C(\mathbf{x},  \tau)$ for classification is defined as:

\begin{equation} \label{eq:conf_pred_cutoff}
  L(\mathbf{x}, \tau) = \min\left\{ r : nc(\mathbf{y}^{(r)} \mid \mathbf{x}) \geq \tau \right\}
\end{equation}
\begin{equation} \label{eq:conf_pred_set}
  C(\mathbf{x}, \tau) = \left\{ \mathbf{y}^{(r)} : r \leq L(\mathbf{x}, \tau) \right\}
\end{equation}

Here, $\mathbf{y}^{(r)}$ represents the $r$-th most {probable} output label sequence under the trained CRF model, ranked by the value function in Equation~\ref{eq:base_CRF}. An acceptable threshold $\tau$ that achieves the desired $1-\alpha$ level of coverage is given by the $\lceil (1 - \alpha)(1 + |\mathcal{I}_{\text{cal}}|) \rceil$-th largest non-conformity score on the calibration set. A proof relating the coverage of the above procedure is provided in the supplementary material. Provided that the calibration and test sets are exchangeable, this threshold guarantees the desired coverage. Notably, Equation~\ref{eq:conf_pred_cutoff} is written for a generic non-conformity score $nc(\mathbf{y}^{(k)} \mid \mathbf{x})$, however, any non-conformity score may be utilized, and all non-conformity scores will achieve at least the desired coverage. Despite this, all non-conformity scores do not achieve the same prediction set efficiency; that is, some non-conformity scores produce a larger-sized prediction set on average than others. 

While the standard conformal prediction procedure outlined above produces valid prediction sets with coverage at least $1 - \alpha$, in this paper, we use the adaptive conformal prediction (ACP) proposed by \cite{romano2020classification}, which produces more efficient prediction sets.
% \ms{this have been moved since you commented}
% \ssg{the order is off here. You should introduce ACP first (fuse this section with section 5.1) and then the specific scores you propose. In fact, you can have a separate subsection for your scores if the combined section is getting too long.}
Assume that a new test observation $(\mathbf{x}_{\text{test}}, \mathbf{y}_{\text{test}})$ is exchangeable with the calibration dataset $\mathcal{I}_{\text{cal}}$. 
% \ssg{this is an assumption rather than a statement of fact, right?} \ms{changed to use `assume' instead of `let'}
The threshold $\tau$ is set to the $(1 - \alpha)$-quantile of the non-conformity scores computed on the calibration set:

\begin{equation}\label{eq:initial quantile}
    \tau = Q_{1-\alpha}\left(\{ nc(\mathbf{y}_i \mid \mathbf{x}_i) \}_{i \in \mathcal{I}_{\text{cal}}} \right) = \lceil (1 - \alpha)(1 + |\mathcal{I}_{\text{cal}}|) \rceil^{th}\text{ largest score in } \mathcal{I}_{\text{cal}}
\end{equation}

In practice, when prediction sets are formed using Equation~\ref{eq:conf_pred_cutoff}, coverage tends to slightly exceed the target level. This is because non-conformity scores increase in discrete steps as each additional sequence is added to the prediction set. Consequently, the final element often causes the total score to overshoot $\tau$, especially when the target coverage is low and each discrete step is large.
To mitigate this, ACP estimates the overshoot and applies a probabilistic rule: the final element is included with probability proportional to how close the cumulative score is to the threshold. The adjusted prediction set is defined as

\begin{equation}
\label{eq:adaptive conformal}
C(\mathbf{x}, \tau, \alpha) = 
\begin{cases}
\{\mathbf{y}^{(1)}, \ldots, \mathbf{y}^{(r)}\} & \text{if } u \leq V(\cdot) \\
\{\mathbf{y}^{(1)}, \ldots, \mathbf{y}^{(r-1)}\} & \text{otherwise}
\end{cases},
\end{equation}
where $nc(\mathbf{y}^{(v-1)} \mid \mathbf{x}) < \tau < nc(\mathbf{y}^{(v)} \mid \mathbf{x})$, and $u \sim \text{Uniform}(0,1)$. The probability of including $\mathbf{y}^{(v)}$ is defined by the overshoot function:

\begin{equation}
\label{eq:cutoff formula}
V(\cdot) = \frac{f(1 - \alpha) - \hat{\pi}(\{\mathbf{y}^{(1)}, \ldots, \mathbf{y}^{(v-1)}\})}{\hat{\pi}(\{\mathbf{y}^{(1)}, \ldots, \mathbf{y}^{(v)}\}) - \hat{\pi}(\{\mathbf{y}^{(1)}, \ldots, \mathbf{y}^{(v-1)}\})}
\end{equation}

where $\hat{\pi}(\cdot)$ is an empirical estimate of the coverage function and $f(1 - \alpha)$ corresponds to the desired coverage level. The `best' choice of $\hat{\pi}$ and $f$ is not known and in practice changes depending on the non-conformity score utilized. Details regarding our choice of $V(\cdot)$ is provided in the supplementary material.

\subsection{Three Baseline Non-conformity Scores}

All equations so far have utilized a generic, undefined non-conformity score. However, set efficiency depends on the choice of the score. Therefore, we define three baseline non-conformity scores:

\begin{equation}\label{score: nc1_sent}
    \mathbf{nc1}(\mathbf{y} \mid \mathbf{x}) = 1 - \hat{\mathbb{P}}(\mathbf{y} \mid \mathbf{x}), \hspace{.2cm} \mathbf{nc2}(\mathbf{y}^{(r)} \mid \mathbf{x}) = \sum_{k=1}^r \hat{\mathbb{P}}(\mathbf{y}^{(k)} \mid \mathbf{x}), \hspace{.2cm}  \mathbf{nc3}(\mathbf{y}^{(r)} \mid \mathbf{x}) = r
\end{equation}

{Where $\hat{\mathbb{P}}(\mathbf{y} \mid \mathbf{x})$ is the predicted probability of label sequence $\mathbf{y}$ given the input embedded word sequence $\mathbf{x}$} arising from the trained CRF model. Previously, all NER and CRF notation was defined using the true probability; however, we will now assume that $\mathbb{P}(\mathbf{y} \mid \mathbf{x}) \approx \hat{\mathbb{P}}(\mathbf{y} \mid \mathbf{x}) $ and will utilize the estimated probability measure for the remainder of our conformal prediction work. All three scores relate the notion of non-conformity to model fit, as on average, $\mathbf{x},\mathbf{y}$ pairs that are common within the training dataset should produce higher predicted probabilities from the fit model when compared to uncommon or absent observation pairs. In \textbf{nc1}, observation pairs that are similar to the training data should contain a high predictive probability and therefore a low non-conformity score. Alternatively, \textbf{nc2} increases as the rank of the observation increases, eventually culminating in a total probability of one. We would expect known observation pairs to occur quickly in the top-k outputs and, therefore, achieve a non-conformity score that is closer to zero. Finally, \textbf{nc3} produces fixed-size observation sets and also relies on the idea that frequently observed observations should be ranked higher than dissimilar/out-of-distribution observations. Although \textbf{nc3} produces sets of constant size, it avoids the scenarios in which \textbf{nc1} and \textbf{nc2} produce large prediction sets with many low probability responses

\subsection{Conformal Prediction for NER}
\label{sec:conformal for NER}

Before any modifications may be made to the conformal prediction techniques for NER, we must first recognize that we modify the underlying probability function presented by Equation~\ref{eq:base_CRF} in order to decrease the computation cost of calculating the denominator. We approximate  the denominator using only the top-$K$ highest-value sequences. Any sequence that does not belong to the top-$K$ highest-value sequences has its value function essentially set to zero. This approximation is common for supervised machine learning methods, which contain a large output space. Let $\mathbf{y}^{(r)}$ denote the $r$-th highest value sequence. The approximate probability becomes:

\begin{equation}
\label{eq:simplified_CRF}
\hat{\mathbb{P}}(\mathbf{y}_{a:a+b} \mid \mathbf{x}) \approx 
\frac{s(\mathbf{y}_{a:a+b})}{\sum_{r=1}^K s(\mathbf{y}_{a:a+b}^{(r)})}
\end{equation}

The choice of K value limits the maximum size of the generated prediction set before the complete set of all possible sequences is returned. This, in turn, limits the effective maximum coverage of the returned prediction sets. We find that in our simulation studies, we utilize K=100 as we find that it is able to achieve an overall maximum desired coverage of approximately 99\% at the sentence level while still being relatively cheap computationally.

\subsubsection{Unconditional Conformal Prediction}
\label{sec:sent level}

Depending on the type of model used, conformal prediction sets may be made for various levels of granularity. For CRF based models, the most direct application of conformal prediction is to form prediction sets over the set of fully labeled word sequences. These conformal prediction sets return a set of full-sequence labelings such that each element in the prediction set is a full-sequence with one label for every input word. The probability of the correct full label sequence being included is calibrated to the desired confidence level.

\begin{table}[t]
\centering
\begin{tabular}{r|cccccc|l}
\textbf{Input:}     & \textit{Sarah} & \textit{is} & \textit{from} & \textit{New} & \textit{York} & \textit{City} & \textbf{nc1}(\textbf{y}$|$\textbf{x})\\
\hline
$\mathbf{y}^{(1)}$: & O            & O               & O           & B-LOC        & I-LOC & I-LOC & .52        \\
$\mathbf{y}^{(2)}$: & B-PER        & O               & O           & B-LOC        & I-LOC & I-LOC & . 72       \\
$\mathbf{y}^{(3)}$: & O            & O               & O           & O            & O  & O & .98         \\
$\mathbf{y}^{(4)}$: & B-PER        & O               & O           & B-ORG        & B-ORG & I-ORG & 1        
\end{tabular}
\caption{Example unconditional prediction set for the sentence ``Sarah is from New York City''.}
\label{tab:sent_pred_set}
\end{table}

Table~\ref{tab:sent_pred_set} provides an example of the top four predicted full-sequence outputs for the sentence `Sarah is from New York City'. 
Given a non-conformity score $nc(\mathbf{y}|\mathbf{x})$ and input sequence $\mathbf{x}$, let $\hat{Q}_{1 - \alpha}$ denote the calibrated threshold estimated from the calibration set. Unconditional conformal prediction is then performed by applying adaptive conformal prediction as described in Equation~\ref{eq:adaptive conformal}. Using table~\ref{tab:sent_pred_set}, if the cutoff threshold were to be .6 then the resulting ACP conformal prediction set would be either $\{\mathbf{y}^{(1)}\}$ or $\{\mathbf{y}^{(1)}, \mathbf{y}^{(2)}\}$ depending on the uniform random variable used in adaptive conformal prediction. As mentioned, there are no separate prediction sets for `Sarah' and `New York', as this method only produces full-label-sequence predictions. The algorithm highlighting the method described in this section is provided in the supplementary material.

\subsection{Stratified Conformal Prediction}
\label{sec:class conditional sentence}

The methodology described in Section~\ref{sec:sent level} will produce prediction sets with an average coverage that is the supplied $1-\alpha$ confidence level. However, when applying the conformal prediction technique, it is readily apparent that there are ways in which you may split the dataset such that a given split may be either under-calibrated or over-calibrated. The most apparent method of splitting the data to induce a miss-calibration in a multilingual model is dividing the data up into different language bins. 

A key property of the exchangeability requirement for conformal prediction is that mixtures of exchangeable sequences remain exchangeable. This enables calibration across confounding or lurking variables, which may impact the nonconformity score, such as sentence length and language. Therefore, we propose a stratified conformal prediction procedure designed to enable precise calibration across such confounding variables.

\begin{theorem}\label{thm:stratified}
    Let $E$ be the sample space for all possible NER inputs and outputs $(\mathbf{x},\mathbf{y})$.
    Consider a partition of $E$ into $m$ mutually exclusive and exhaustive subsets such that $\bigcup_{j=1}^m E_j = E, \mathbb{P}(\bigcup_{j=1}^m E_j) = 1$ and $\forall j \neq k, E_j \cap E_k = \varnothing, \mathbb{P}(E_j \cap E_k) = 0$. 

    Then the prediction set formed by the following equations

    \begin{equation*} \label{eq: quantile}
    \tau_j= Q_{1-\alpha}\left(\{ \mathbf{nc}(\mathbf{y}_i \mid \mathbf{x}_i) \}_{(\mathbf{x_i},\mathbf{y_i})\in E_j} \right)
    \end{equation*}
    \begin{equation*} \label{eq: subset pred}
        C_j(\mathbf{x},\tau) = \{\mathbf{y} : \mathbf{nc}(\mathbf{y}|\mathbf{x_i}) \leq \tau_j \}
    \end{equation*}

    is well calibrated for observations belonging to each subset.

    \begin{equation}\label{eq:conditional prob}
    \mathbb{P}(\mathbf{y}_{n+1} \in C_j(\mathbf{x}_{n+1},\tau_j)|(\mathbf{x_{n+1}},\mathbf{y_{n+1}})\in E_j) \geq 1-\alpha
    \end{equation}
    
\end{theorem}

The proof for Theorem~\ref{thm:stratified} is contained in the supplementary material.

\subsubsection{Length and Language Stratification}

As previously mentioned, the stratified conformal prediction procedure may be applied to confounding variables such as language and length. By partitioning the event space into varying subsections, we can isolate poorly performing sections, such that their bad performance does not increase the average prediction set size of otherwise low-uncertainty sections. To calibrate a model for varying language and sentence lengths, define a partition $\{E_j\}$ of the sample space according to language and length combinations. Since each sentence has a unique language and length, this clearly leads to a valid partition of the sample space with mutually exhaustive and exclusive subsets; therefore, following Theorem~\ref{thm:stratified} leads to valid prediction sets for all languages and length combinations. Data regarding the calibration of full-sequence conformal prediction for language and length are displayed in Tables~\ref{tab: language calibration results} and \ref{tab:length calibratriton}. Further figures and tables are also provided in the supplementary material.

The limit to the number of strata in which you are able to stratify your data set is based on the number of observations and the ability to accurately group each stratum. When splitting our dataset to account for sentence length, due to the amount of data available, it is impractical to separate each `length' in its entirety, as the benchmark calibration set becomes too small for rarely occurring sentence lengths. Therefore, we group similar-length sentences together. This method of grouping similar class structures is motivated by the work of \cite{ding2023class} in the clustering of granular classes for regular classification problems.  For all benchmark datasets, sentence-length strata are grouped into ranges of 1–10, 11–20, 21–30, 31–40, and 40+ words. In applications, as the amount of data available for calibration increases, the size of the sentence-length groupings may be decreased such that each group is more granular.

\subsection{Index-Based Non-Conformity Scores}
\label{sec:index nc score}
As discussed earlier, conformal prediction for NER can be applied in multiple configurations with various non-conformity scores. However, when constructing prediction sets for NER, users often run into a common issue: the creation of large prediction sets populated by low-probability outputs. This is particularly problematic when using the cumulative probability score (\textbf{nc2}). The top few predictions may account for most of the confidence mass, but still fall short of the threshold $\tau$, requiring the inclusion of many low-confidence predictions, inflating the prediction set size.

Notably, this run-on behavior affects both \textbf{nc1} (probability) and \textbf{nc2} (cumulative probability), but not \textbf{nc3} (rank-based). Because \textbf{nc3} enforces a fixed-size threshold, it naturally limits growth. We propose two hybrid strategies and evaluate one existing strategy that combines probability-based and index-based conformal prediction to harness their respective strengths. The effectiveness of each hybrid strategy is evaluated in Section \ref{sec:result-index}.

\subsubsection{Naive Intersection of Sets}
The first hybrid method constructs its hybrid prediction set as the intersection of prediction sets formed by two distinct non-conformity scores. This combination combines the ability of a probability-based non-conformity score (\textbf{nc1} or \textbf{nc2}) to stop prediction set construction early while allowing there to be a maximum set size via \textbf{nc3}.  Let $C_{idx}$ and $C_{prob}$ be prediction sets calibrated at levels $1 - \alpha$ and $1 - \beta$, respectively,
then the probability of a response belonging to their intersection is bounded by:

\begin{equation}\label{eq: naieve prediction set}
    \mathbb{P}(\mathbf{y}_{\text{new}} \in C_{naive}(\mathbf{x}) ) =\mathbb{P}\left( \mathbf{y}_{\text{new}} \in C_{idx}(\mathbf{x}) \cap C_{prob}(\mathbf{x}) \right) \geq 1 - \alpha - \beta.
\end{equation}

The proof of Equation~\ref{eq: naieve prediction set} is provided in the supplementary material. When the parameters $\alpha$ and $\beta$ are selected appropriately, the resulting hybrid method yields a prediction set that is at least as efficient as those produced by either non-conformity score alone as either $\alpha$ or $\beta$ could always be set to zero. 

\subsubsection{Conditional Prediction Sets}
\label{index-conditional prediction sets}
An alternative is to constrain one score to be conditioned on the other being below a threshold. Let: $C_{\text{idx}}(\mathbf{x})$ be a prediction set that utilizes \textbf{nc3} with a desired coverage of $1-\alpha$. We may then define a second set $
C_{\text{prob}}(\mathbf{x})$ using only those calibration samples that fall within this index range:

\begin{equation}
    \mathbb{P}(\mathbf{y}_{\text{new}} \in C_{\text{prob}}(\mathbf{x}) \mid \mathbf{y}_{\text{new}} \in C_{\text{idx}}(\mathbf{x})) \geq 1 - \beta
\end{equation}

The total coverage of the intersection is thus bounded by:

\begin{equation}
    \mathbb{P}(\mathbf{y}_{\text{new}} \in C_{cond.}(\mathbf{x}) ) = \mathbb{P}(\mathbf{y}_{\text{new}} \in C_{\text{idx}}(\mathbf{x}) \cap C_{\text{prob}}(\mathbf{x})) \geq (1 - \alpha)(1 - \beta)
\end{equation}

This approach enforces tighter control than the naive method while reducing unnecessary set growth; the proof of its coverage is in the supplementary material. Like the naive prediction set, the best values of $\alpha$ and $\beta$ depend on the specific application and the non-conformity scores. Although we describe both the conditional and naive techniques as a combination of a probability and an index-based non-conformity score, these techniques are not limited to those choices. Instead, any two non-conformity score combinations may be utilized. 

\subsubsection{Regularized Prediction Sets}

A third approach is the RAPS (Regularized Adaptive Prediction Sets) procedure 
 \citep{angelopoulos2020uncertainty}, which modifies the conformity score to include a linear penalty on the index. Unlike the previous methods, the RAPS procedure is a modified non-conformity score and does not utilize two separate non-conformity scores or two prediction sets. The RAPS procedure was initially proposed using a variation of \textbf{nc2} we have chosen to write its definition utilizing a generic non-conformity score as later in this paper we will show how it may be improved by utilizing \textbf{nc1} instead.

\begin{equation}
    \textbf{nc}_\textbf{raps}(\mathbf{y}^{(r)}|\mathbf{x}) = nc(\mathbf{y}^{(r)}|\mathbf{x}) + \lambda \cdot \max(r - \tau_{\text{idx}}, 0)
\end{equation}

Like the previous two methods, the addition of a regularization term reduces the amount of low-probability results that may be included in the prediction set. While effective, this method assumes a fixed linear relationship ($\lambda$) and does not allow for flexible control of how much the index contributes to total error. In both the naive and conditional approaches, the user can set a maximum set size based on the chosen index and understand approximately how much error is associated with each non-conformity score.  In contrast, the RAPS procedure does allow for the user to set a maximum set size by selecting a large value of $\lambda$, but does not easily allow for the user to know how much error is due to the choice of $\lambda$, $\tau_{\text{idx}}$, and $k$ versus the used non-conformity score.

\section{Subsequence Conformal Prediction}
\label{sec:ent level}

As an alternative to full-sequence prediction sets, we may instead consider conformal prediction on the entity-level where each object in the prediction set corresponds to the labeling of one entity without outside words/secondary entities. This allows for these prediction sets to be well calibrated for each entity-class.

We first redefine the NER classification problem as a non-sequential prediction problem where individual prediction sets may be formed for each entity. Given a sentence, let a sequence of predicted labels be denoted as $\mathbf{y} = \{y_1, y_2, \ldots, y_t\}$. A continuous subsequence is defined as $\mathbf{y}_{a:a+b} = \{y_a, y_{a+1}, \ldots, y_{a+b}\}$.\ Before being able to produce subsequence prediction sets, we first define the probability of a subsequence prediction as the sum of probabilities of full sequences that contain the subsequence. 

Using the example sentence from Table~\ref{tab:sent_pred_set}, the probability that the subsequence \textit{Sarah is from} is labeled as \{\texttt{B-PER} \texttt{O} \texttt{O} \}, is equal to $\mathbf{y}^{(2)} + \mathbf{y}^{(4)}$. More generally:

$$ \hat{\mathbb{P}}(\mathbf{y}_{a:a+b}) = \sum_{\mathbf{y} \in \mathcal{Y}} \hat{\mathbb{P}}(\mathbf{y})\mathbb{I}(\mathbf{y}_{a:a+b} \subseteq \mathbf{y})$$

Because the sentence level prediction sets utilize a simplification in which a positive probability mass is only assigned to the top-K decoded sequence labels and the sum of the top-k probability sequences is equal to 1, the above equation may be simplified to:

\begin{equation}
     \hat{\mathbb{P}}(\mathbf{y}_{a:a+b}) = \sum_{i=1}^k \hat{\mathbb{P}}(\mathbf{y}^{(i)})\mathbb{I}(\mathbf{y}_{a:a+b} \subseteq \mathbf{y^{(i)}})
\end{equation}

Using the above probability distributions, prediction sets may be formed using the original full-sequence prediction methodology described in equations \eqref{eq:conf_pred_cutoff} and \eqref{eq:conf_pred_set}. Despite this, our goal is to construct subsequence prediction sets for subsequences containing singular named entities with entity-class calibration for all entity classes. In order to do so, we must precisely define the notion of 'entity-specific subsequences'. A subsequence $\mathbf{y}_{a:a+b}$ belongs to a entity class if the label of $\mathbf{y}_a$ utilizes the `B' prefix (e.g. \texttt{B-PER}), all subsequent labels $(\mathbf{y}_{a+1}, \cdots,\mathbf{y}_{a+b})$ utilize the `I' prefix with the same class suffix (e.g. \texttt{I-PER}), and the label after the subsequence differs from the previous label ($\mathbf{y}_{a+b} \neq \mathbf{y}_{a+b+1} )$. If the above statement is true, we state that the subsequence is equal to the given class (e.g. $\mathbf{y}_{a:a+b} = \texttt{PER}$).

Using these guidelines, we may see how in Table~\ref{tab:sent_pred_set} the probability of ``New York City'' ($\mathbf{x}_{4:6}$) being labeled as a location is equal to $\hat{\mathbb{P}}(\mathbf{y^{(1)}}) + \hat{\mathbb{P}}(\mathbf{y^{(2)}})$. Additionally, we may also note that for the full sentence to be considered `correctly labeled' two distinct $1-\alpha$ subsequence prediction sets must be satisfied, leading to a family-wise error problem. 

Using the above example, the probability of the subsequence $\mathbf{y}_{a:a+b}$ belonging to a specific entity class $w \in \mathcal{W}$ is estimated using the top-$K$ decoded sequences as:

\begin{equation}
\label{eq:ent probability def}
\hat{\mathbb{P}}_{ent}(\mathbf{y}_{a:a+b} = w) = \sum_{r=1}^k \hat{\mathbb{P}}(\mathbf{y}^{(r)}) \mathbb{I}(\mathbf{y}_{a:a+b} = w)
\end{equation}

Using this definition, we extend the standard non-conformity scores \textbf{nc1}, \textbf{nc2}, and \textbf{nc3} (Equation \eqref{score: nc1_sent}) to the class-specific subsequence setting:

\begin{equation}
\label{eq:subseq_class_1}
\mathbf{nc1_{ent}}(\mathbf{y}\mid \mathbf{x}, w,a,b) = 1-\hat{\mathbb{P}}_{ent}(\mathbf{y}_{a:a+b} = w),
\end{equation}

\begin{equation}
    \label{eq:subseq_class_2}
     \mathbf{nc2_{ent}}(\mathbf{y} \mid  \mathbf{x}, w,a,b) = \sum_{ w^* \in \mathcal{W}} \hat{\mathbb{P}}_{ent}(\mathbf{y}_{a:a+b} = w^*) \cdot \mathbb{I}\left(\hat{\mathbb{P}}_{ent}(\mathbf{y}_{a:a+b} = w^*) \leq \hat{\mathbb{P}}_{ent}(\mathbf{y}_{a:a+b} = w)\right),
\end{equation}

\begin{equation}
    \label{eq:subseq_class_3}
    \mathbf{nc3_{ent}}(\mathbf{y} \mid \mathbf{x}, w,a,b) =  \sum_{w^* \in \mathcal{W}} \mathbb{I}\Big(\hat{\mathbb{P}}_{ent}(\mathbf{y}_{a:a+b} = w^*) \leq \hat{\mathbb{P}}_{ent}(\mathbf{y}_{a:a+b} = w)\Big).
\end{equation}

From the modified non-conformity scores, the subsequence conformal prediction sets may be defined for each entity class such that:

\begin{equation}
    \mathbb{P}\left(w \in C_{w,ent}(\mathbf{x}_{\text{new}}, \tau_w,a,b)| \mathbf{y}_{a:a+b} = w\right) \geq 1 - \alpha
\end{equation}

Where $\tau_w$ is the class-specific cutoff found by finding the $1-\alpha$ quantile of the calibration dataset's nonconformity scores for subsequences belonging to class $w$. Utilizing this cutoff, the subsequence conformal prediction set of class $w$, $C_{w,ent}(\mathbf{x}, \tau_w,a,b)$ is given by:

\begin{equation} \label{eq:ent pred set}
    C_{w,ent}(\mathbf{x}, \tau_w,a,b) = \{ w \text{ if } nc_{ent}(\mathbf{y} \mid \mathbf{x},w,a,b) \leq \tau_w \}
\end{equation}

The class-specific cutoff is defined as the $1-\alpha$ quantile of $nc_{ent}$ on the subset of entities whose true value is $w$ in the calibration partition. We define the set of all $w$ class subsequences in the calibration dataset as $\mathcal{I}_{cal}^w$. Each entity prediction set is either the null set $\{\}$, indicating that the subsequence does not belong to entity class $w$, or the singular set $\{ w \}$, indicating the entity may belong to the given class.  We now define the union of entity class prediction sets:

\begin{equation}\label{eq: ent pred set comb}
     C_{ent}(\mathbf{x},\mathbf{\tau}_{\mathcal{W}},a,b) = \bigcup_{w\in \mathcal{W}} C_{w,ent}(x,\tau_w,a,b)
\end{equation}

Where $\mathbf{\tau}_{\mathcal{W}}$ is the set of all class specific cutoffs: $\{\tau_{per},\tau_{loc},\tau_{org},\tau_{misc} \} $. This union of all class prediction sets achieves the desired coverage level, that is:

\begin{equation}
    \forall \text{ }  w \in \mathcal{W}{}, \text{    } \mathbb{P}\big(\mathbf{y}_{a:a+b} \in C_{ent}(x,\tau_{\mathcal{W}},a,b)| \mathbf{y}_{a:a+b} = w\big) \geq 1-\alpha
\end{equation}

 The proof of the above statement is provided in the supplementary material alongside a full algorithm depicting the subsequence conformal prediction procedure defined here.

\section{Integrated Conformal Prediction}
\label{sec:sent+ent}

Subsequence conformal prediction provides class-conditional prediction sets for all contiguous subsequences of the input sentence. However, these sets do not capture contextual dependencies across different entities within a full sentence. That is, although each subsequence prediction is individually calibrated, the subsequence framework does not specify how these predictions should be combined to form a coherent full-sequence labeling. To bridge this gap, we introduce an integrated approach that merges subsequence prediction sets into full-sentence predictions.

Let $C_{ent}(\mathbf{x},\mathbf{\tau}_{\mathcal{W}},a,b)$ be the prediction set of all entity types for subsequence $\mathbf{y}_{a:a+b}$. For a given candidate full-sequence labeling $\mathbf{y} = \{y_1,y_2,...y_t\}$, let $E(\mathbf{y})$ be the set of $s$ subsequences within $\mathbf{y}$, $(y_{a_i:a_i+b_i}  \text{ for }1\leq i \leq s)$ that map to a given class in $\mathcal{W}$, as defined in Section~\ref{sec:ent level}. If all $y_{a_i:a_i+b_i}\in E(\mathbf{y})$ exist within their respective prediction sets $C_{ent}(\mathbf{x},\mathbf{\tau}_{\mathcal{W}},a_i,b_i)$ then we claim that the candidate full-sequence labeling is included in the integrated prediction set. More specifically, for any candidate sequence $y^{(i)}$:

$$ y^{(i)}_{a:a+b} \in E(\mathbf{y}^{(i)}) \text{ iff } \mathbf{y}^{(i)}_{a:a+b} \neq \texttt{Non-Entity}$$

Let the set of all continuous subsequences be given by: 

\begin{equation}\label{eq: nontrivial bounds}
    S(\mathbf{x}) = \{(a,b) : a+b \leq len(x), a \geq 1, b \geq 0 \} 
\end{equation}  

Using these definitions, the integrated prediction set may be defined as: 

\begin{equation} \label{eq:integrated predictions}
    C_{int}(\mathbf{x},\tau_{\mathcal{W}}) = \{\mathbf{y}^{(i)}: E(\mathbf{y}^{(i)}) \subseteq \bigcup_{(a,b) \in S(\mathbf{x})} C_{ent}(\mathbf{x},\mathbf{\tau}_{\mathcal{W}},a,b)\}
\end{equation}

The key challenge in this aggregation is the risk of family-wise error. Specifically, if a candidate sequence labeling contains $s$ total entitien then the probability of all entities being jointly found by their independent $1 - \alpha$ confidence prediction sets has a lower bound of $(1-\alpha)^s$. We do note that the true probability should be higher than this lower bound (as the prediction sets are not independent). To mitigate the family-wise error of our combined prediction sets, we apply the Šidák correction to estimate the required per-entity confidence level. Let $\hat{s}$ be the number of predicted entities in the top-ranked model output $\mathbf{y}^{(1)}$. Then, the per-entity confidence level is adjusted to:
\[
1 - \alpha_{\text{Šidák}} = (1 - \alpha)^{1 / \hat{s}}
\]
Results regarding the coverage of the integrated methodology before and after the sidak correction are provided later in Table~\ref{tab: n_ent_results} and the supplementary material.

To prevent low-probability sequences from being included, in these integrated prediction sets we can follow a similar indexing procedure to the full-sequence prediction sets. By combining the integrated probability-based non-conformity score with an index-related non-conformity score.
Define an index-entity conformity score as follows.

\begin{equation}
    \label{eq:subseq_class_idx}
    \mathbf{nc3_{ent,idx}}(\mathbf{y} \mid \mathbf{x}, w,a,b) = min(i|\mathbf{y}^{(i)}_{a:a+b} = w)
\end{equation}

Utilizing the above non-conformity score an index-referential conformal prediction set may be constructed on the entity level:

\begin{equation} \label{eq:ent pred set idx}
    C_{w,ent,idx}(\mathbf{x}, \tau_w,a,b)  = \{min(i| \mathbf{y}_{a:a+b} = w) 
     \text{ if } nc_{ent}(\mathbf{y}\mid \mathbf{x},w,a,b) \leq \tau_w \}
\end{equation}

\begin{equation}
     C_{ent,idx}(\mathbf{x},\mathbf{\tau}_{\mathcal{W}},a,b) = \bigcup_{w\in \mathcal{W}} C_{w,ent,idx}(x,\tau_w,a,b)
\end{equation}

When adapted for the integrated output level, the prediction set formed by $\mathcal{C}_{ent,idx}$ may be given by:

\begin{equation} \label{eq:integrated predictions idx}
    C_{idx}(\mathbf{x},\tau_{\mathcal{W}}) = \{\mathbf{y}^{(i)}: i \leq max\bigg[ \bigcup_{(a,b) \in S(\mathbf{x})} C_{ent,idx}(\mathbf{x},\mathbf{\tau}_{\mathcal{W}},a,b)\bigg]\}
\end{equation}

After construction, the above prediction set may be combined with the initial $\mathcal{C}_{int}$ prediction set in order to limit the number of low-probability/high index prediction directly sets that are included within the integrated prediction set method.

\begin{equation} \label{eq:integrated predictions int idx}
    C_{idx,int}(\mathbf{x},\tau_{\mathcal{W}},\tau^*_{\mathcal{W}}) =  C_{int}(\mathbf{x},\tau_{\mathcal{W}}) \cap C_{idx}(\mathbf{x},\tau^*_{\mathcal{W}}) 
\end{equation}

When calibrating the above equation, if  $\mathcal{C}_{int}$ is set to achieve a desired coverage of $1-\alpha$ and $\mathcal{C}_{idx}$ is set to achieve a desired coverage of $1-\beta$, then the overall coverage of the unified index-integrated methodology may be given by:

\begin{align}
     \mathbb{P}(\mathbf{y}_{new} \in C_{idx,int}(\mathbf{x},\tau_{\mathcal{W}},\tau^*_{\mathcal{W}})) = \mathbb{P}(\mathbf{y}_{new} \in  C_{int}(\mathbf{x},\tau_{\mathcal{W}}) \cap C_{idx}(\mathbf{x},\tau^*_{\mathcal{W}})) \geq 1- \alpha - \beta
\end{align}

The proof of the above statement follows from the Naive hybrid prediction sets defined earlier. Proofs regarding the coverage of $C_{int}$ and $C_{idx}$ as defined above are contained in the supplementary material.

\section{Experiments\label{sec:Exp}}

We evaluate our methods across three benchmarks and four base models, comparing full-sequence, subsequence, and integrated prediction sets, as well as different non-conformity scores and calibration strategies.

\subsection{Benchmark Datasets}

We evaluate our methods on three primary benchmark datasets: CoNLL++ \citep{wang2019crossweigh}, CoNLL-reduced, and WikiNEuRal \citep{tedeschi2021wikineural}.
CoNLL++ is a refined version of the original CoNLL-2003 benchmark, featuring a reduced number of labeling errors. It includes four entity classes: \texttt{PER}, \texttt{LOC}, \texttt{ORG}, and \texttt{MISC}. We use the English partition of CoNLL++, which consists of annotations from Reuters newswire articles collected between August 1996 and August 1997.
CoNLL-Red is a reduced and simplified version of CoNLL++, in which all entity classes are merged into a single `Entity' category. This benchmark provides a way to evaluate the uncertainty around entity identification without classification. This is important as models which were trained on a different benchmark may be successful at entity identification but not entity classification. 
WikiNEuRal is a multilingual benchmark built from Wikipedia articles. Named entities in WikiNEuRal are generated through a silver-standard data creation process and align with Wikipedia’s own entity annotations. It adopts the CoNLL class schema and spans nine languages: Dutch, English, French, German, Italian, Polish, Portuguese, Russian, and Spanish.  We utilize the multilingual form of WikiNEuRal throughout sections 4-6 to demonstrate various techniques for generating language-specific coverage and report English-only results for model comparisons, since not all models are multilingual.

\subsection{Base Models}

We evaluate four different base models, all of which were chosen due to their effectiveness at performing NER and their public availability at \href{huggingface.co}{huggingface.com}.

\href{https://huggingface.co/Babelscape/wikineural-multilingual-ner}{Babelscape} \citep{tedeschi2021wikineural} is a multilingual NER model trained on the WikiNEuRal dataset. It is utilized throughout this paper to demonstrate language and length calibration.
\href{https://huggingface.co/dslim/bert-base-NER}{Dslim} \citep{DBLP:journals/corr/abs-1810-04805} is the smallest model among those considered (110 million parameters) and was trained on the CoNLL-2003 benchmark. 
\href{https://huggingface.co/Jean-Baptiste/roberta-large-ner-english}{Jean-Baptiste} \citep{JeanBaptistePolle}is the second model in our evaluation trained on the CoNLL-2003 dataset and unlike other models utilizes RoBERTa-based embeddings. 
\href{https://huggingface.co/tner/roberta-large-ontonotes5}{TNER} \citep{ushio-camacho-collados-2021-ner} is a large NER model trained on the OntoNotes dataset \citep{weischedel2013ontonotes} and is the only OntoNotes-based baseline included in our evaluation. We selected TNER to illustrate how effectively a model that was trained with a different ontology and trained on out-of-distribution data can be fine-tuned for conformal prediction on the CoNLL dataset. 
% \ssg{provide citations for these models}\ms{all models are now cited}
%Furthermore, we compare its adaptability to CoNLL with that of CoNLL- or WikiNEuRal-based models adapting to OntoNotes.

\subsection{Calibrating Non-conformity scores}
\label{sec: nc calibration results}

As described in Sections~\ref{sec:class conditional sentence}–\ref{sec:sent+ent}, we introduced methods for calibrating conformal prediction sets to account for language, sequence length, and number of entities. The following tables detail the resulting changes in empirical coverage and average set size for a desired coverage of 95\%. More detailed results for other coverage values are provided in the supplementary material. For consistency, all results use the Babelscape model on the multilingual WikiNEuRal dataset.

\begin{table*}[t]
\centering
\scriptsize
\resizebox{\linewidth}{!}{%
\begin{tabular}{llcccc}
\toprule
\textbf{Length Range} & \textbf{Desired Cov.} 
& \multicolumn{2}{c}{\textbf{No Calibration}} 
& \multicolumn{2}{c}{\textbf{Length Calibration}} \\
\cmidrule(lr){3-4}\cmidrule(lr){5-6}
& &  \textbf{Emp. - Desired Cov.} & \textbf{Average Set Size} & \textbf{Emp. - Desired Cov.} & \textbf{Average Set Size} \\
\midrule
1–10   & 0.95  & 0.0341 & 1.8184 & 0.0150 & \textbf{1.3032} \\
\midrule
11–20   & 0.95  & 0.0284 & 1.8396 & 0.0174 & \textbf{1.5684} \\
\midrule
21–30   & 0.95  & 0.0162 & 1.8445 & 0.0157 & 1.8191 \\
\midrule
31–40   & 0.95  & 0.0016 & 1.8406 & 0.0119 & \textbf{\textit{2.2697}} \\
\midrule
40+ & 0.95  & \textbf{-0.0170} & 1.8415 & 0.0076 & \textbf{\textit{6.0854}} \\
\bottomrule
\end{tabular}
}
\caption{Model calibration for five sequence length bins on the WikiNEuRal benchmark. Coverage columns are bolded if the empirical coverage (Emp. Cov.) is less than the desired coverage (Desired Cov.). Set size columns are bolded if there is a decrease of greater than .1 in average set size and bolded + italicized if there is an increase of greater than .1 in average set size.}
\label{tab:length calibratriton}
\end{table*}

Table~\ref{tab:length calibratriton} summarizes calibration by sentence length. For short sequences (bins 1–10 and 11–20), calibration reduces the degree of over-coverage and substantially decreases set sizes. Lengths 21–30 remain relatively unchanged, while longer bins (31–40 and 40+) show increased coverage and larger sets. Notably, calibration corrects the under-coverage of the 40+ bin, raising it above the target. Because non-conformity score distributions vary by length, perfect per-length calibration would require single-length bins; however, bin granularity is limited by the amount of available calibration data. 

\begin{table*}[t]
\centering
\scriptsize
\resizebox{.8\linewidth}{!}{%
\begin{tabular}{llcccc}
\toprule
\textbf{Language} & \textbf{Desired Cov.} 
& \multicolumn{2}{c}{\textbf{No Calibration}} 
& \multicolumn{2}{c}{\textbf{Language Calibration}} \\
\cmidrule(lr){3-4}\cmidrule(lr){5-6}
 &  & \textbf{Emp. - Desired Cov.} & Average Set Size & \textbf{Emp. - Desired Cov.}& Average Set Size \\
\midrule
Dutch       & 0.95  & 0.0389  & 1.8212 & 0.0152 & \textbf{1.2703} \\
\midrule
English    & 0.95  & \textbf{-0.0118} & 1.8617 & 0.0124 & \textbf{\textit{3.3362}} \\
\midrule
French      & 0.95  & 0.0156  & 1.8462 & 0.0173 & 1.8648 \\
\midrule
German       & 0.95  & 0.0354  & 1.8307 & 0.0175 & \textbf{1.3981} \\
\midrule
Italian     & 0.95  & 0.0184  & 1.8390 & 0.0148 & \textbf{1.7171} \\
\midrule
Polish       & 0.95  & 0.0003  & 1.8344 & 0.0108 & \textbf{\textit{2.5685}} \\
\midrule
Portuguese  & 0.95  & 0.0136  & 1.8326 & 0.0132 & 1.8205 \\
\midrule
Russian     & 0.95  & \textbf{-0.0269} & 1.8571 & 0.0110 & \textbf{\textit{6.6575}} \\
\midrule
Spanish     & 0.95  & 0.0419  & 1.8133 & 0.0131 & \textbf{1.1570} \\
\midrule
\end{tabular}
}
\caption{Model calibration for all languages in the WikiNEuRal benchmark. Coverage columns are bolded if the empirical coverage is less than the desired coverage. Set size columns are bolded if there is a decrease of greater than .1 in average set size and bolded + italicized if there is an increase of greater than .1 in average set size.}
\label{tab: language calibration results}
\end{table*}

Table~\ref{tab: language calibration results} shows analogous results for language calibration. Under-coverage in English and Russian is corrected, though often at the cost of larger prediction sets (e.g., English 95\% coverage increases from 1.8617 to 3.3362 predictions, with coverage improving from 93.82\% to 96.24\%). Other languages, like German, Italian, and Spanish, see reduced set sizes and over-coverage improvements.

\begin{table*}[t]
\centering
\scriptsize
\resizebox{\linewidth}{!}{%
\begin{tabular}{lrrrrrrrrrrrr}
\toprule
\textbf{Prediction type}   
& \multicolumn{2}{c}{\textbf{1 Entity}}
& \multicolumn{2}{c}{\textbf{2 Entities}}
& \multicolumn{2}{c}{\textbf{3 Entities}}
& \multicolumn{2}{c}{\textbf{4 Entities}}
& \multicolumn{2}{c}{\textbf{5 Entities}} \\
\cmidrule(lr){2-3}\cmidrule(lr){4-5}\cmidrule(lr){6-7}\cmidrule(lr){8-9}\cmidrule(lr){10-11}
& \textbf{Cov.} & \textbf{Set Size} & \textbf{Cov} & \textbf{Set Size} & \textbf{Cov} & \textbf{Set Size} & \textbf{Cov} & \textbf{Set Size} & \textbf{Cov} & \textbf{Set Size} \\
\midrule
Full-sequence  & 0.9731 & 1.4527 & \textbf{0.9411} & 1.6003 & \textbf{0.9102} & 1.7559 & \textbf{0.9019} & 1.8736 & \textbf{0.9068} & 2.0519 \\

\midrule
Int. without Šidák  & 0.9766 & 7.5755 & \textbf{0.9293} & 9.2887 & \textbf{0.8859} & 10.1569 & \textbf{0.8748} & 10.9781 & \textbf{0.8682} & 11.7518 \\

\midrule
Int. with Šidák & 0.9773 & 7.6536 & 0.9623 & 16.3684 & 0.9635 & 24.0292 & 0.9629 & 26.1613 & 0.9565 & 28.6008 \\
\bottomrule
\end{tabular}
}

\caption{Model calibration per number of named entities within a sentence for three methods, Full-Sequence, Integrated without a Šidák family-wise error control correction, and Integrated with said correction. All prediction sets are calculated on the WikiNEuRal benchmark utilizing the Babelscape model and \textbf{nc1} non-conformity score. Coverage columns are bolded if the empirical coverage is less than the desired coverage. }
\label{tab: n_ent_results}
\end{table*}

Table~\ref{tab: n_ent_results} compares full-sequence and integrated methods across sentences containing differing numbers of entities. Both full-sequence and Integrated-without-Šidák approaches fail to maintain valid coverage for multi-entity inputs. Applying Šidák correction within the integrated method yields valid coverage for all five entities at 95\% coverage. While full-sequence calibration could be improved by partitioning data by entity count, doing so along with existing partitions by language and length would produce bins too small for reliable calibration. In contrast, the integrated method avoids binning based on the number of entities by dynamically increasing subsequence-level coverage as the estimated number of entities grows, leading to valid combined sets.

Results for entity-class calibration are presented separately in Table~\ref{tab:hybrid comparisons} (Section~\ref{subsec:class cond}).

\subsection{Comparing Index-based Methods}
\label{sec:result-index}

As alluded to in previous sections, we compared the efficiency of all proposed non-conformity scores for full-sequence, subsequence, and integrated prediction sets. We begin with Table ~\ref{tab:index comparisons}, which shows the efficiency of \textbf{nc1}, \textbf{nc2}, and \textbf{nc3} when calculated with the Babelscape model on the multilingual WikiNEuRal benchmark. 

\begin{table*}[t]
\centering
\begin{tabular}{c|cc|cc|cc}
\toprule
\multicolumn{7}{c}{\textbf{full-sequence}} \\
\midrule
\multirow{2}{*}{\textbf{Prediction }} 
& \multicolumn{2}{c|}{\textbf{nc1}} 
& \multicolumn{2}{c|}{\textbf{nc2}} 
& \multicolumn{2}{c|}{\textbf{nc3}} \\
\cmidrule(r){2-7}
\textbf{Type} & Cov. & Size & Cov. & Size & Cov. & Size \\
\midrule
full-sequence & 0.9679 & 1.8400 & 0.9948 & 45.4994 & 0.9438 & 1.3345 \\
\midrule
subsequence & 0.9895 & 1.1815 & 0.9994 & 1.9008 & 0.9812 & 1.1812 \\
\midrule
Integrated & 0.9796 & 13.1685 & 0.994 & 34.0406 & 0.9665 & 20.8824 \\
\bottomrule
\end{tabular}
\caption{Average coverage and set size for \textbf{nc1}, \textbf{nc2}, and \textbf{nc3} as evaluated on the multi-lingual WikiNEuRal benchmark with the Babelscape model } 
\label{tab:index comparisons}
\end{table*}

In Table \ref{tab:index comparisons}, \textbf{nc1} performs significantly better than \textbf{nc2} and \textbf{nc3} for subsequence and integrated prediction sets. Regarding full-sequence predictions, \textbf{nc3} produces sets of smaller size; however, the coverage of \textbf{nc3} is below the desired coverage of 95\%. For all prediction types, \textbf{nc2} performs significantly worse than both \textbf{nc1} and \textbf{nc3}, producing extremely large prediction sets while significantly overstepping the desired coverage levels. Further results for coverage levels other than 95\% are provided in the supplementary material.

\begin{table*}[t]
\centering
\resizebox{\linewidth}{!}{%
\begin{tabular}{cc|cc|cc|cc|cc|cc}
\multicolumn{4}{c}{Full-sequence} & \multicolumn{4}{c}{Subsequence}& \multicolumn{4}{c}{Integrated} \\
\toprule
\multicolumn{2}{c|}{\textbf{Naive + nc1}} 
& \multicolumn{2}{c|}{\textbf{Naive + nc2}}
& \multicolumn{2}{c|}{\textbf{Naive + nc1}} 
& \multicolumn{2}{c|}{\textbf{Naive + nc2}}
& \multicolumn{2}{c|}{\textbf{Naive + nc1}} 
& \multicolumn{2}{c}{\textbf{Naive + nc2}}\\
Cov. & Size & Cov. & Size & Cov. & Size & Cov. & Size & Cov. & Size & Cov. & Size  \\
\midrule
0.9504 & 1.3626 & 0.9569 & 2.7537 & 0.98 & 1.1585 & 0.9945 & 1.2402 & 0.965 & 16.7349 & 0.9893 & 26.1235 \\

\\
\multicolumn{2}{c|}{\textbf{RAPS + nc1}} 
& \multicolumn{2}{c|}{\textbf{RAPS + nc2}} 
& \multicolumn{2}{c|}{\textbf{RAPS + nc1}} 
& \multicolumn{2}{c|}{\textbf{RAPS + nc2}}
& \multicolumn{2}{c|}{\textbf{RAPS + nc1}} 
& \multicolumn{2}{c}{\textbf{RAPS + nc2}}\\
Cov. & Size & Cov. & Size & Cov. & Size & Cov. & Size & Cov. & Size & Cov. & Size  \\
\midrule
 0.9496 & 1.6049 & 0.9495 & 2.1582 & 0.9698 & 1.0443 & 0.9762 & 1.0552 & 0.9504 & 2.6704 & 0.9641 & 4.9603 \\

\\
\multicolumn{2}{c|}{\textbf{Cond. + nc1}} 
& \multicolumn{2}{c|}{\textbf{Cond. + nc2.}}
& \multicolumn{2}{c|}{\textbf{Cond. + nc1}} 
& \multicolumn{2}{c|}{\textbf{Cond. + nc2.}}
& \multicolumn{2}{c|}{\textbf{Cond. + nc1}} 
& \multicolumn{2}{c}{\textbf{Cond. + nc2.}}\\
Cov. & Size & Cov. & Size & Cov. & Size & Cov. & Size & Cov. & Size & Cov. & Size  \\
\midrule
0.9518 & 1.42 & 0.9546 & 3.5607 & 0.9817 & 1.1221 & 0.9872 & 1.2178 & 0.969 & 5.9108 & 0.9799 & 6.6276 \\

\bottomrule
\end{tabular}
}
\caption{Average coverage and set size for the naive, RAPS, and conditional methods when utilizing either \textbf{nc1} or \textbf{nc2} as evaluated on the multilingual WikiNEuRal benchmark with the Babelscape model across all types of prediction sets}
\label{tab:indexcomparisons2}
\end{table*}

 Table~\ref{tab:indexcomparisons2} displays the efficiency and coverage of the six index-based integrated non-conformity scores when evaluated on full-sequence, subsequence, and integrated prediction sets for the WikiNEuRal Benchmark with the Babelscape model. In general, we observe that all integrated methods, when utilizing \textbf{nc1}, perform better than their \textbf{nc2} counterparts in almost all situations. The RAPS procedure also performs slightly better than the conditional and naive methods for most situations except for the full-sequence \textbf{nc1} prediction set. Some performance gaps may be due to coarse grid searches for estimating the conditional method’s $(\alpha, \beta)$ and the RAPS penalty $\lambda$, though the extent of this effect is unclear. Although the Naive method performs worse than both RAPS and conditional, it still improves on the initial \textbf{nc1} and \textbf{nc2} results from Table~\ref{tab:index comparisons}, with the greatest improvement being on \textbf{Naive + nc2}.

\subsubsection{Comparing Class Conditional Coverage} \label{subsec:class cond}

A key motivation for the integrated method is achieving entity-class conditional coverage regardless of sentence structure. Table~\ref{tab:hybrid comparisons} reports average set size and coverage by class using length and language-controlled \textbf{nc1} and $\textbf{nc1}_{\text{ent}}$ scores. Full-sequence sets fail to meet class-conditional coverage for the Miscellaneous class but remain smaller overall. Subsequence sets remain small for all classes because the output space contains only five labels, whereas full-sequence and integrated sets may contain up to $k$ explicit sequences before returning the full set of all possible sequences.

\begin{table*}[t]
\centering
\resizebox{\linewidth}{!}{%
\begin{tabular}{c|cc|cc|cc|cc}
\textbf{Prediction Type} & \multicolumn{2}{c|}{\textbf{PER}} & \multicolumn{2}{c|}{\textbf{LOC}} & \multicolumn{2}{c|}{\textbf{ORG}} & \multicolumn{2}{c}{\textbf{MISC}} \\
& Coverage & Set Size & Coverage & Set Size & Coverage & Set Size & Coverage & Set Size \\
\midrule
\textbf{Full Sequence} & 0.9685 & 1.8325 & 0.9637 & 1.8347 & 0.9572 & 1.8417 & \textbf{0.9303} & 1.861 \\

\textbf{Subsequence} & 0.9895 & 1.0994 & 0.9884 & 1.2159 & 0.9882 & 1.2669 & 0.9917 & 1.1198 \\

\textbf{Integrated} & 0.9798 & 15.415 & 0.9756 & 15.285 & 0.9785 & 16.2023 & 0.9732 & 18.23911 \\
\end{tabular}}
\caption{Average coverage and prediction set size of each entity class by conformal prediction methodology with the \textbf{nc1} non-conformity score. Non-conformity scores were calculated using the Babelscape model on the multilingual WikiNEuRal benchmark dataset. Empirical coverages below the desired coverage threshold are bolded.  \label{tab:hybrid comparisons}}
\end{table*}

\subsection{Benchmark Comparisons}\label{sec:res_benchmark_comparisons}

The following subsection presents the performance of four NER models evaluated across the CoNLL, CoNLL\_Red, and WikiNEuRal\_en datasets. For ease of comparison, we report only the results obtained using the \textbf{nc1} non-conformity score and 95\% coverage. This confidence level exceeds the best one-shot accuracy of the base models and thus yields non-trivial prediction sets.

Table~\ref{tab:sentence results} reports the average coverage and prediction set size for full-sequence prediction sets. As expected, all models achieved empirical coverage close to the target of 95\%. However, the Jean-Baptiste model outperformed others on both the CoNLL and CoNLL reduced datasets, likely due to its original training on CoNLL. In contrast, the Dslim model underperformed on the CoNLL dataset despite also being trained on it. Babelscape, being trained on the WikiNEuRal dataset, exhibited superior performance on it while performing poorly on CoNLL and CoNLL reduced. Interestingly, Dslim outperformed Jean-Baptiste on WikiNEuRal, and is only slightly behind Jean on CoNLL reduced, suggesting that it may be struggling more with entity-classification and not entity-identification. TNER performed the worst on the WikiNEuRal task while performing better than Dslim/Babelscape on Conll and better than Babelscape on CoNLL reduced.

\begin{table}[t]
\centering
\resizebox{\linewidth}{!}{%
\begin{tabular}{c|cc|cc|cc}
\toprule
 & \multicolumn{2}{c|}{\textbf{CoNLL}} & \multicolumn{2}{c|}{\textbf{CoNLL Reduced}} & \multicolumn{2}{c|}{\textbf{WikiNEuRal\_en}} \\
 \textbf{Benchmark} & \textbf{Coverage} & \textbf{Set Size} & \textbf{Coverage} & \textbf{Set Size} & \textbf{Coverage} & \textbf{Set Size} \\
\midrule
\textbf{Babelscape} & 0.942 $\pm$ 0.007 & 25.19 $\pm$ 0.969 & 0.941 $\pm$ 0.009 & 4.025 $\pm$ 0.096 &  0.955 $\pm$ 0.003 & \textbf{2.465 $\pm$ 0.022}\\
\textbf{Dslim} & 0.947 $\pm$ 0.007 & 23.021 $\pm$ 0.882 & 0.951 $\pm$ 0.004 & 2.089 $\pm$ 0.015 & 0.951 $\pm$ 0.004 & 4.927 $\pm$ 0.076  \\
\textbf{Jean} & 0.948 $\pm$ 0.006 & \textbf{4.556 $\pm$ 0.38} & 0.960 $\pm$ 0.008 & \textbf{1.786 $\pm$ 0.023} & 0.953 $\pm$ 0.003 & 5.676 $\pm$ 0.052 \\
\textbf{TNER} & 0.943 $\pm$ 0.004 & 18.492 $\pm$ 0.678 & 0.954 $\pm$ 0.001 & 2.557 $\pm$ 0.006 & 0.950 $\pm$ 0.002 & 6.621 $\pm$ 0.134 \\
\bottomrule
\end{tabular}
}
\caption{Coverage and Set Size by Model across three benchmarks for full-sequence prediction sets at 95\% target coverage. Average coverage and prediction set size are presented with a 95\% prediction set formed from twenty differing calibration and test splits.
% \ssg{To improve readability, put the datasets in columns and models in rows}\ms{Is something like this what you were thinking?} 
}
\label{tab:sentence results}
\end{table}

Table~\ref{tab:non-entity} presents prediction set sizes for spans incorrectly identified as entities (false positives). We observe that the average prediction set size for these false positive spans is extremely small. Babelscape achieves the smallest sets on the WikiNEuRal benchmark. The  Dslim, Jean-Baptiste, and TNER models produce considerably larger prediction sets on WikiNEuRal, suggesting that these models may yield a substantial number of false positives.  For CoNLL and CoNLL reduced, the Jean model achieved the smallest prediction sets. All base models produce significantly larger prediction sets for CoNLL compared to CoNLL Reduced, suggesting once again that most mistakes are with entity classification and not entity identification.

\begin{table*}[t]
\centering

\label{tab:o_entity_setsize}
\resizebox{\linewidth}{!}{%
\begin{tabular}{llcccc}
\toprule
\textbf{Benchmark} & \textbf{Metric} & \textbf{Babelscape} & \textbf{Dslim} & \textbf{Jean} & \textbf{Tner} \\
\midrule
\multirow{1}{*}{CoNLL}
  & Set Size & $0.087 \pm 0.006$ & $0.339 \pm 0.008$ & $\mathbf{0.059 \pm 0.008}$ & $0.200 \pm 0.010$ \\
\midrule
\multirow{1}{*}{CoNLL Reduced}
  & Set Size & $\mathbf{0.003 \pm 0.000}$ & $0.019 \pm 0.000$ & $\mathbf{0.002 \pm 0.000}$ & $0.035 \pm 0.002$ \\
\midrule
\multirow{1}{*}{Wikineural\_en}
  & Set Size & $\mathbf{0.117 \pm 0.002}$ & $0.662 \pm 0.008$ & $0.922 \pm 0.010$ & $0.604 \pm 0.002$ \\
\bottomrule
\end{tabular}
}
\caption{Set Size of subsequence prediction sets by model for incorrectly identified named entities (false positives) at a 95\% confidence level.}
\label{tab:non-entity}
\end{table*}

Table~\ref{tab:entity_performance} presents results for the subsequence prediction set method. In this approach, prediction sets are generated for all identified spans, with each set containing up to five possible labels: one for each entity class and an additional label for non-entity spans. We observe that the average set size is nearly one for all models. Despite this, the models are not over-generating NER entities, as we can see from Table~\ref{tab:non-entity} that the average set size of non-entity spans is near zero, indicating that the conformal prediction procedure is able to differentiate between valid and erroneous entity spans. 

For the CoNLL and WikiNEuRal benchmarks, the subsequence prediction sets consistently achieve coverage above the desired threshold while maintaining an average prediction set size of approximately 1.2 to 2 NER types per identified entity span. Consistent with previous results, the Babelscape model continues to outperform its counterparts on the WikiNEuRal benchmark while underperforming on the CoNLL benchmark compared to all other methods. Meanwhile, Jean performs best for CoNLL while being equivalent to Dslim for both CoNLL reduced and WikiNeuRal. TNER, while not being the best at any benchmark, outperforms Babelscape on CoNLL and CoNLL reduced.

\begin{table*}[t]
\centering
\resizebox{\linewidth}{!}{%
\begin{tabular}{llllll}
\toprule
\textbf{Benchmark} & \textbf{Metric} & \textbf{Babelscape} & \textbf{Dslim }& \textbf{Jean} & \textbf{Tner} \\
\midrule
\multirow{2}{*}{CoNLL}
& Coverage & $0.981 \pm 0.005$ & $0.983 \pm 0.001$ & $0.983 \pm 0.001$ & $0.98 \pm 0.002$ \\
& Set Size & $1.959 \pm 0.026$ & $1.363 \pm 0.013$ & $\mathbf{1.208 \pm 0.003}$ & $1.675 \pm 0.02$ \\
\midrule
\multirow{2}{*}{CoNLL\_red}
& Coverage & 0.994 $\pm$ 0.002 & 0.986 $\pm$ 0.001 & 0.988 $\pm$ 0.001 & 0.993 $\pm$ 0.001 \\
& Set Size & 0.992 $\pm$ 0.002 & $\mathbf{0.961 \pm 0.003}$ & $\mathbf{0.964 \pm 0.004}$ & 0.986 $\pm$ 0.002 \\
\midrule
\multirow{2}{*}{WikiNeuRal\_en}
& Coverage & $0.984 \pm 0.002$ & $0.984 \pm 0.002$ & $0.985 \pm 0.002$ & $0.982 \pm 0.001$ \\
& Set Size & $\mathbf{1.28 \pm 0.004}$ & $1.407 \pm 0.007$ & $1.399 \pm 0.007$ & $1.431 \pm 0.01$\\
\bottomrule
\end{tabular}
}
\caption{Coverage and Set Size of subsequence prediction sets with 95\% confidence for each model and entity type in CoNLL and the English partition of WikiNEuRal.The best overall prediction set size is bolded in each benchmark.}
\label{tab:entity_performance}

\end{table*}

Finally, Table~\ref{tab:hybrid method} presents the results for the integrated method prediction sets for a desired coverage of 95\%. Consistent with earlier results, Babelscape performs best on WikiNEuRal, while Jean and Dslim yield the strongest results on CoNLL and CoNLL reduced, respectively. Interestingly, TNER slightly outperforms Jean on the reduced CoNLL benchmark. The substantial increase in average prediction set size for the Babelscape model may indicate that its top-ranked outputs include many plausible combinations of NER entities in the CoNLL and reduced CoNLL benchmarks.

\begin{table*}[t]
\centering

\resizebox{\linewidth}{!}{%
\begin{tabular}{llcccc}
\toprule
\textbf{Model} & \textbf{Metric} & \textbf{Babelscape} & \textbf{Dslim} & \textbf{Jean} & \textbf{Tner} \\
\midrule
\multirow{2}{*}{CoNLL} 
  & Coverage & $0.946 \pm 0.021$ & $0.947 \pm 0.006$ & $0.951 \pm 0.014$ & $0.928 \pm 0.004$ \\
  & Set Size & $25.835 \pm 0.760$ & $16.518 \pm 0.530$ & $\mathbf{13.854 \pm 0.089}$ & $27.993 \pm 0.810$ \\
\midrule
\multirow{2}{*}{CoNLL\_red} 
  & Coverage & $0.948 \pm 0.002$ & $0.953 \pm 0.002$ & $0.964 \pm 0.008$ & $0.962 \pm 0.005$ \\
  & Set Size & $14.780 \pm 0.460$ & $\mathbf{5.833 \pm 0.053}$ & $7.215 \pm 0.024$ & $6.448 \pm 0.027$ \\
\midrule
\multirow{2}{*}{WikiNEuRal\_en} 
  & Coverage & $0.966 \pm 0.005$ & $0.963 \pm 0.003$ & $0.965 \pm 0.004$ & $0.948 \pm 0.009$ \\
  & Set Size & $\mathbf{7.678 \pm 0.091}$ & $19.052 \pm 0.107$ & $19.169 \pm 0.043$ & $19.866 \pm 0.082$ \\
\bottomrule
\end{tabular}
}
\caption{Coverage and Set Size by Benchmark and Model for Integrated prediction sets (conditional + \textbf{nc1} w/ Šidák correction) at a desired coverage of 95\%. The best overall prediction set size is bolded in each benchmark.}
\label{tab:hybrid method}
\end{table*}

\section{Conclusion\label{sec:conc}}

This paper introduces a framework for applying conformal prediction to NER models, enabling finite-sample uncertainty quantification through prediction sets rather than point estimates. We evaluated full-sequence, subsequence, and integrated conformal methods using several non-conformity scores across several benchmark datasets and models.

Our results highlight the importance of conditioning on language and sentence length to obtain well-calibrated sets across diverse inputs. We also address the challenge of excessive, low probability, prediction sets by evaluating one established (RAPS) and two new (naive and conditional) index–probability combination strategies built to combat this issue. Methods that rely on model-probability scores (\textbf{nc1}) consistently outperform cumulative-probability scores (\textbf{nc2}), including in our modified RAPS implementation, which delivered the most efficient sets. Because the original formulation of RAPS uses \textbf{nc2}, we recommend that future work test the adapted RAPS + \textbf{nc1} formulation in other domains. Furthermore, although the RAPS procedure produces slightly more efficient prediction sets than the conditional method we proposed, we note that the conditional method is more interpretable, as it allows for the specific specification of how much error is due to the maximum index bound being reached versus the maximum probability bound being reached.

Our benchmarking further shows that any NER model can achieve target coverage when calibrated by language and length, though efficiency is strongly model-dependent. Models perform best on benchmarks aligned with their training data; for example, Babelscape on WikiNEuRal and Jean-Baptiste on CoNLL. To compare how models perform when only identifying entity spans and not entity classes, we introduced the CoNLL-Reduced benchmark, which collapses entity types into entity/non-entity tags. Results show minimal differences in average set size across models on the reduced benchmark, even when large differences appear under the original CoNLL ontology, suggesting that out-of-domain models mainly struggle with class-type distinctions rather than span identification.

Uncertainty quantification for NER remains a relatively nascent area of research, particularly as modern architectures shift toward larger LLMs and encoder–decoder models. The techniques and considerations introduced in this paper can be adapted and extended to these emerging architectures, offering significant potential for future work. Moreover, despite the growing interest in uncertainty quantification for NER, research remains sparse on how this uncertainty can be effectively propagated through downstream NLP tasks. Addressing this gap would enable the development of multi-stage NLP pipelines that account for interconnected sources of uncertainty, ultimately supporting more informed decision-making and reducing the risk of error propagation. Regarding future directions of work, we did not analyze the optimization aspect of the conditional + \textbf{nc1} score versus the RAPS + \textbf{nc1} score and the effect of the hyperparameter grid-search used in this paper.

\bibliographystyle{imsart-nameyear} % Style BST file
\bibliography{ref_merged}       % Bibliography file (usually '*.bib')

%% or include bibliography directly:
% \begin{thebibliography}{}
% \bibitem[\protect\citeauthoryear{???}{???}]{b1}
% \end{thebibliography}

\clearpage
\makeatletter
\renewcommand{\thesection}{S\arabic{section}}
\makeatother

% To restart numbering at 1
\setcounter{section}{0}
\section{Supplement for Section 4: Conformal Prediction}

\subsection{Proof of Conformal Prediction (Equation~\ref{base_conformal guarantee})}

The construction of the following proposition follows from \cite{gupta2022nested}; we simply repeat their proof with our differing notation.

Suppose we observe data $\mathcal{D} = \{( \mathbf{x}_i, \mathbf{y}_i)\}_{i=1}^N$. Let this dataset be partitioned into two subsections,  $\mathcal{I}_{\text{train}}$ and $\mathcal{I}_{\text{cal}}$. A CRF model is trained using $\mathcal{I}_{\text{train}}$. Let  $nc(\mathbf{y}|\mathbf{x})$ denote a generic non-conformity score based on the outputs of the trained CRF model. To construct a prediction set based on the above non-conformity score, calculate the $(1-\alpha)$-quantile on the calibration dataset $\mathcal{I}_{\text{cal}}$.

\begin{align} \label{eq: quantile_ref}
    \tau=& Q_{1-\alpha}\left(\{ \mathbf{nc}(\mathbf{y}_i \mid \mathbf{x}_i) \}_{i \in \mathcal{I}_{\text{cal}}} \right) \nonumber \\
    =& \lceil (1 - \alpha)(1 + |\mathcal{I}_{\text{cal}}|) \rceil^{th}\text{ largest score in } \mathcal{I}_{\text{cal}}
\end{align}

Next, the prediction set may be constructed using $nc(\cdot,\mathbf{x}_i)$ and $\tau$:

\begin{equation} \label{eq: prediction set}
    C(\mathbf{x},\tau) = \{\mathbf{y} : \mathbf{nc}(\mathbf{y}|\mathbf{x_i}) \leq \tau \}
\end{equation}

\begin{prop} \label{prop:basic_conformal}
    Assuming $\{( \mathbf{x}_i, \mathbf{y}_i)\}_{i\in \cup\mathcal{I}_{\text{cal}}\cup[n+1] }$ are exchangeable, then the prediction set $C(\cdot)$ in Equation~\ref{eq: prediction set} satisfies:
    
    \begin{equation}
        \mathbb{P}(\mathbf{y}_{n+1} \in C(\mathbf{x}_{n+1},\tau)) \geq 1-\alpha
    \end{equation}
\end{prop}

\begin{proof}
    By the construction of the confidence interval $C(\mathbf{x},\tau)$:

    $$ \mathbf{y}_{n+1} \in C(\mathbf{x}_{n+1},\tau) \text{ iff } \mathbf{nc}(\mathbf{y}|\mathbf{x_i}) \leq \tau $$
    
    Therefore
    
    $$ \mathbb{P}(\mathbf{y}_{n+1} \in C(\mathbf{x}_{n+1},\tau)) = \mathbb{P}(nc(\mathbf{y}_{n+1}|\mathbf{x}_{n+1}) \leq \tau) $$
    
    The exchangeability of $\{( \mathbf{x}_i, \mathbf{y}_i)\}_{i\in \cup\mathcal{I}_{\text{cal}}\cup[n+1] }$, implies that $\{\mathbf{nc}(\mathbf{y}_i|\mathbf{x}_i)\}_{i\in \cup\mathcal{I}_{\text{cal}}\cup[n+1] }$ is also exchangeable. Thus, Lemma 2 of \citet{romano2019conformalized} yields
    
    $$ \mathbb{P}(\mathbf{nc}(\mathbf{y}_{n+1}|\mathbf{x}_{n+1}) \leq \tau) \geq 1-\alpha $$
\end{proof}

\subsection{Choice of Overshoot Function}

To the best of our knowledge, there has been no detailed review of the impact of the overshoot function on ACP. Indeed, the original paper \citep{romano2020classification} defines the overshoot function only for the oracle classifier. In our limited exploration of overshoot functions, we found that when evaluating equation~\eqref{eq:cutoff formula} with probabilistic scores such as \textbf{nc1} and \textbf{nc2}, the following two approximations work well:

\begin{equation}
    V(\cdot) = \frac{\hat{\mathbb{P}}(\mathbf{y}^{(r-1)}) - \tau}{\hat{\mathbb{P}}(\mathbf{y}^{(r-1)}) - \hat{\mathbb{P}}(\mathbf{y}^{(r)})} \text{ or }  V(\cdot) = \frac{\sum^{r-1}_{v = 1}\hat{\mathbb{P}}(\mathbf{y}^{(v)}) - \tau}{\hat{\mathbb{P}}(\mathbf{y}^{(v)})}
\end{equation}

where $\tau$ is the prediction set cutoff, and $\hat{\mathbb{P}}$ is the estimated model probability for the given value of $y$. For the rank-based score \textbf{nc3}, an approximation is obtained using the calibration quantile function:

\begin{equation}
    V(\cdot) = \frac{(1 - \alpha) - Q(nc(\mathbf{y}^{(v-1)}))}{Q(nc(\mathbf{y}^{(v)}) - Q(nc(\mathbf{y}^{(v-1)}))}
\end{equation}

where $Q(nc(\mathbf{y}^{(v)}))$ is the empirical CDF value (percentile) of the score within the calibration set $\mathcal{I}_{\text{cal}}$. 

Table \ref{tab: adaptive ex} we demonstrate the effectiveness of the adaptive coverage method when utilizing the three baseline non-conformity scores on the multilingual WikiNEuRal benchmark dataset and the Babelscape model. More information regarding the WikiNEuRal benchmark and babelscape model may be found in section~\ref{sec:Exp}.  The results demonstrate that the adaptive coverage method achieves empirical coverage levels that more closely align with the desired threshold of 90\%. Additionally, the table highlights a key characteristic of the \textbf{nc3} method: its use of a fixed set size. As a result, when early stopping is applied, an empty prediction set is consistently returned. Table \ref{tab: adaptive ex} also demonstrates how \textbf{nc2} constructs significantly larger prediction sets than \textbf{nc1} and \textbf{nc3} while also overshooting the desired coverage.

\begin{table}[t]
\centering
\resizebox{.45\textwidth}{!}{%

\begin{tabular}{llcc}
\textbf{Dataset} & \textbf{Method} & \textbf{Confidence} & \textbf{Set Size} \\
\hline
\multirow{3}{*}{NC1}
  & Early Stop      & 0.8983 & 0.927 \\
  & Full Prediction & 0.9525 & 1.927 \\
  & ACP            & 0.9267 & 1.287 \\
\hline
\multirow{3}{*}{NC2}
  & Early Stop      & 0.8981 & 36.64 \\
  & Full Prediction & 0.9951 & 37.64 \\
  & ACP            & 0.9946 & 37.02 \\
\hline
\multirow{3}{*}{NC3}
  & Early Stop      & 0.0000 & 0.0000 \\
  & Full Prediction & 0.9351 & 1.0000 \\
  & ACP             & 0.8885 & 0.9546 \\
\end{tabular}%
}
\caption{Average coverage and set size of unconditional prediction sets for \textbf{NC1}, \textbf{NC2}, and \textbf{NC3} across the early stopping, full prediction, and ACP conformal prediction methods when constructing prediction sets at 90\% coverage. Results are reported from the Babelscape model as evaluated on the WikiNEuRal benchmark.}
\label{tab: adaptive ex}
\end{table}

\subsection{Algorithm for Sequence-level conformal prediction}

The following algorithm provides a baseline for sequence-level conformal prediction without language/length stratification. To include language/length stratification, simply divide the sample space into a language/length partition $\{E_j\}$, calculate $\tau_j$ for each partition, and apply Algorithm~\ref{algo:sent} to each partition. 

\begin{algorithm}[t]
\DontPrintSemicolon
\KwIn{Training/calibration data $\mathcal{I}_{\text{train}}, \mathcal{I}_{\text{cal}}$; test input $(\mathbf{x}_{\text{test}}, \mathbf{y}_{\text{test}})$; confidence level $1-\alpha$; non-conformity score $nc$; beam width $k$}
\KwOut{Prediction set $C(\mathbf{x}_{\text{test}}, \tau)$}
Train CRF model $M$ on $\mathcal{I}_{\text{train}}$\;
Compute $\tau = \hat{Q}_{1 - \alpha}( \{ nc(\mathbf{y}_i \mid \mathbf{x}_i) \}_{i \in \mathcal{I}_{\text{cal}}})$\;
Predict top-$k$ sequences $\mathbf{y}^{(1)}, \ldots, \mathbf{y}^{(k)} = M(\mathbf{x}_{\text{test}})$\;
Initialize $C \gets \{\}$\;
$i \gets 1$\;
\While{$nc(\mathbf{y}^{(i)} \mid \mathbf{x}_{\text{test}}) < \tau$ \textbf{and} $i \leq k$}{
    $C \gets C \cup \{\mathbf{y}^{(i)}\}$\;
    $i \gets i + 1$\;
}
$u \sim \text{Unif}(0,1)$\;
$v \gets [  \hat{\mathbb{P}}(\mathbf{y}^{(i)}) - \tau] /\text{ } [\hat{\mathbb{P}}(\mathbf{y}^{(i)}) - \hat{\mathbb{P}}(\mathbf{y}^{(i+1)})]$\;
\If{$u \leq v$}{
    $C \gets C \cup \{\mathbf{y}^{(i)}\}$\;
}
\Return{$C$}
\caption{{\sc Unconditional Conformal Prediction}}
\label{algo:sent}
\end{algorithm}

\subsection{Proof of Stratified Conformal Prediction}

Theorem~\ref{thm:stratified} is reiterated here for ease of reference.

\begin{theorem*}
    Let $E$ be the sample space for all possible NER inputs and outputs $(\mathbf{x},\mathbf{y})$.
    Consider a partition of $E$ into $m$ mutually exclusive and exhaustive subsets such that $\bigcup_{j=1}^m E_j = E, \mathbb{P}(\bigcup_{j=1}^m E_j) = 1$ and $\forall j \neq k, E_j \cap E_k = \varnothing, \mathbb{P}(E_j \cap E_k) = 0$. 

    Then the prediction set formed by the following equations

    \begin{equation*} \label{eq: quantile_ref2}
    \tau_j= Q_{1-\alpha}\left(\{ \mathbf{nc}(\mathbf{y}_i \mid \mathbf{x}_i) \}_{(\mathbf{x_i},\mathbf{y_i})\in E_j} \right)
    \end{equation*}
    \begin{equation*} \label{eq: subset pred_ref}
        C_j(\mathbf{x},\tau) = \{\mathbf{y} : \mathbf{nc}(\mathbf{y}|\mathbf{x_i}) \leq \tau_j \}
    \end{equation*}

    is well calibrated for observations belonging to each subset such that:

    \begin{equation}\label{eq:conditional prob_ref}
    \mathbb{P}(\mathbf{y}_{n+1} \in C_j(\mathbf{x}_{n+1},\tau_j)|(\mathbf{x_{n+1}},\mathbf{y_{n+1}})\in E_j) \geq 1-\alpha
    \end{equation}
    
\end{theorem*}

\begin{proof}
    Let $E$ be the sample space for all possible NER inputs and outputs $(\mathbf{x},\mathbf{y})$.
    Consider a partition of $E$ into $m$ mutually exclusive and exhaustive subsets such that $\bigcup_{j=1}^m E_j = E, \mathbb{P}(\bigcup_{j=1}^m E_j) = 1$ and $\forall j \neq k, E_j \cap E_k = \varnothing, \mathbb{P}(E_j \cap E_k) = 0$.

    Now consider the calibration dataset $ \mathcal{D}_{cal} = \{( \mathbf{x}_i, \mathbf{y}_i)\}_{i\in \mathcal{I}_{\text{cal}} } $. Assuming that the calibration data is exchangeable, we may partition the calibration dataset based on the sample space partition $\{E_j\}_{j=1}^m$. Let  
    $$\mathcal{D}^j_{cal} = \{( \mathbf{x}_i, \mathbf{y}_i) \in \mathcal{D}_{cal} : ( \mathbf{x}_i, \mathbf{y}_i) \in E_j\} =\{( \mathbf{x}_i, \mathbf{y}_i)\}_{i\in \mathcal{I}^j_{\text{cal}}},$$   
    where $\mathcal{I}^j_{\text{cal}}$ is the set of indices corresponding to observations belonging to the $E_j$. Clearly, each partition $D^j_{cal}$ of the calibration data inherits the exchangeable property.
    
    We will now define a conformal prediction set, conditional on the true observation belonging to the partition $E_j$.   
    \begin{align*} \label{eq: quantile_ref3}
        \tau_j=& Q_{1-\alpha}\left(\{ \mathbf{nc}(\mathbf{y}_i \mid \mathbf{x}_i) \}_{(\mathbf{x_i},\mathbf{y_i})\in E_j} \right) \nonumber \\
        =& Q_{1-\alpha}\left(\{ \mathbf{nc}(\mathbf{y}_i \mid \mathbf{x}_i) \}_{i \in \mathcal{I}^j_{cal}} \right) \nonumber\\
        = & \lceil (1 - \alpha)(1 + |\mathcal{I}^j_{cal}|) \rceil^{th}\text{ largest} \nonumber \\
        &\text{score in } \mathcal{I}^j_{cal}
    \end{align*}
    
    A prediction set may then utilize this cutoff to generate a conditionally-calibrated prediction set:
    \begin{equation*} \label{eq: subset pred_ref2}
        C_j(\mathbf{x},\tau) = \{\mathbf{y} : \mathbf{nc}(\mathbf{y}|\mathbf{x_i}) \leq \tau_j \}
    \end{equation*}
        
    Given the above, let a new observation $(\mathbf{x}_{n+1},\mathbf{y_{n+1}})$ be exchangeable with the calibration data $\{ (\mathbf{x}_i, \mathbf{y}_i) \}_{i \in \mathcal{I}_{cal}}$. 
    We must have $(\mathbf{x_{n+1}},\mathbf{y_{n+1}})\in E_j$ for some $1 \le j \le m$.
    Then, $\{ \mathbf{nc}(\mathbf{y}_i \mid \mathbf{x}_i) \}_{i \in \mathcal{I}^j_{cal}\cup [n+1]}$ is exchangeable. Therefore, via Proposition~\ref{prop:basic_conformal}:
    
    \begin{equation}\label{eq:conditional prob proof}
        \mathbb{P}(\mathbf{y}_{n+1} \in C_j(\mathbf{x}_{n+1},\tau_j)|(\mathbf{x_{n+1}},\mathbf{y_{n+1}})\in E_j) \geq 1-\alpha
    \end{equation}

\end{proof}

\subsection{Language and Length Calibration Figures}

\begin{figure}[t]
    \centering
    \includegraphics[width=0.8\linewidth]{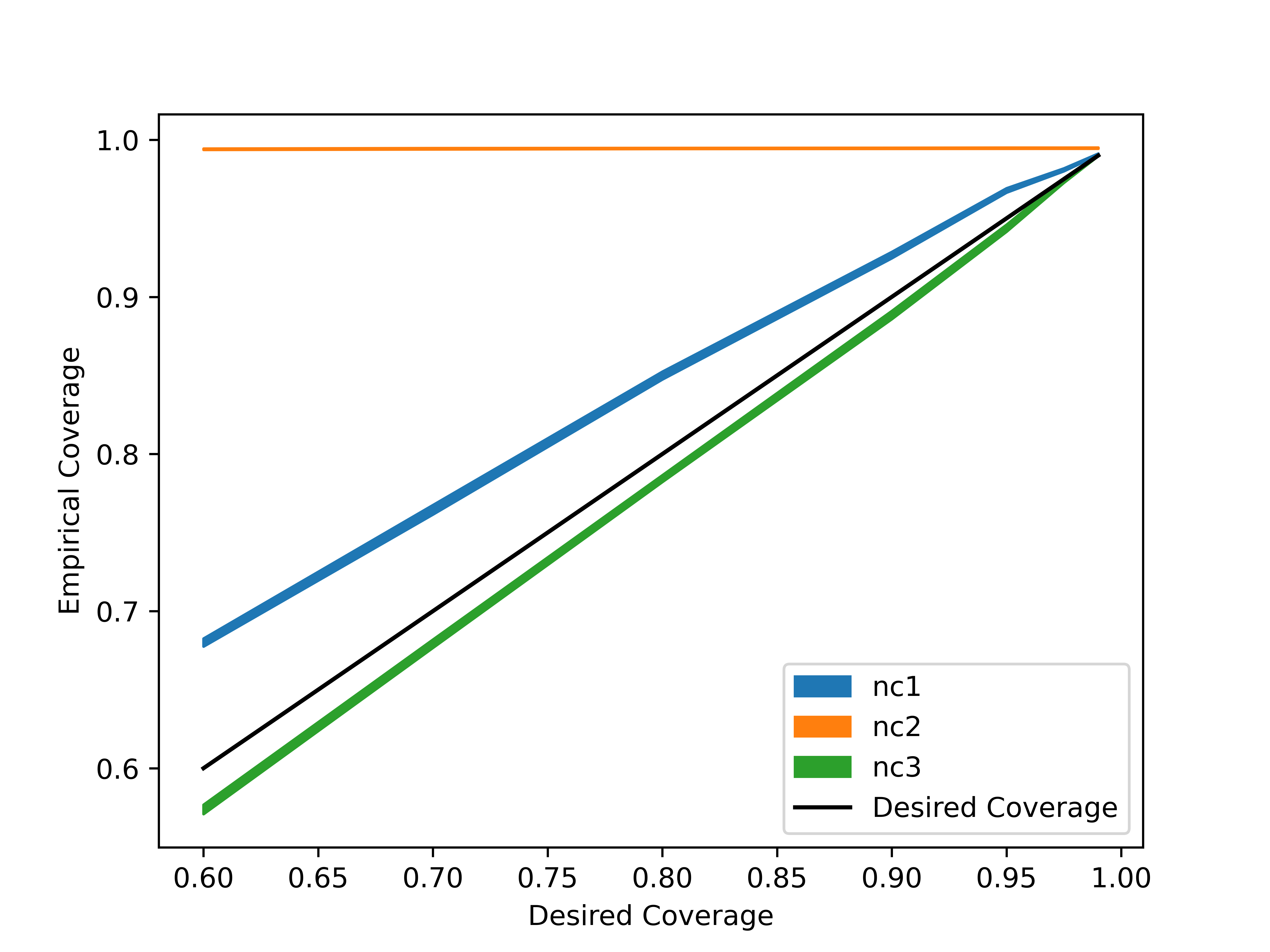}
    \caption{Overall calibration of full-sequence prediction sets for \textbf{nc1}, \textbf{nc2}, and \textbf{nc3} when computed with the Babelscape model on the multilingual WikiNEuRal benchmark dataset. The coverage depicted is calculated utilizing the \textbf{nc1} nonconformity score and displays the 95\% confidence interval produced by 20 iterations.}
    \label{fig:initial calibration}
\end{figure}

Figure~\ref{fig:initial calibration} shows that the full-sequence conformal prediction methods achieve the desired coverage when aggregated over all inputs regardless of their language or sentence length. This coverage figure was generated for the multilingual WikiNEuRal benchmark dataset with the Babelscape model. For ease of comparison, all future figures displayed in this section rely on the \textbf{ nc1} non-conformity score.  

\begin{figure}[t]
    \centering
    \includegraphics[width=0.8\linewidth]{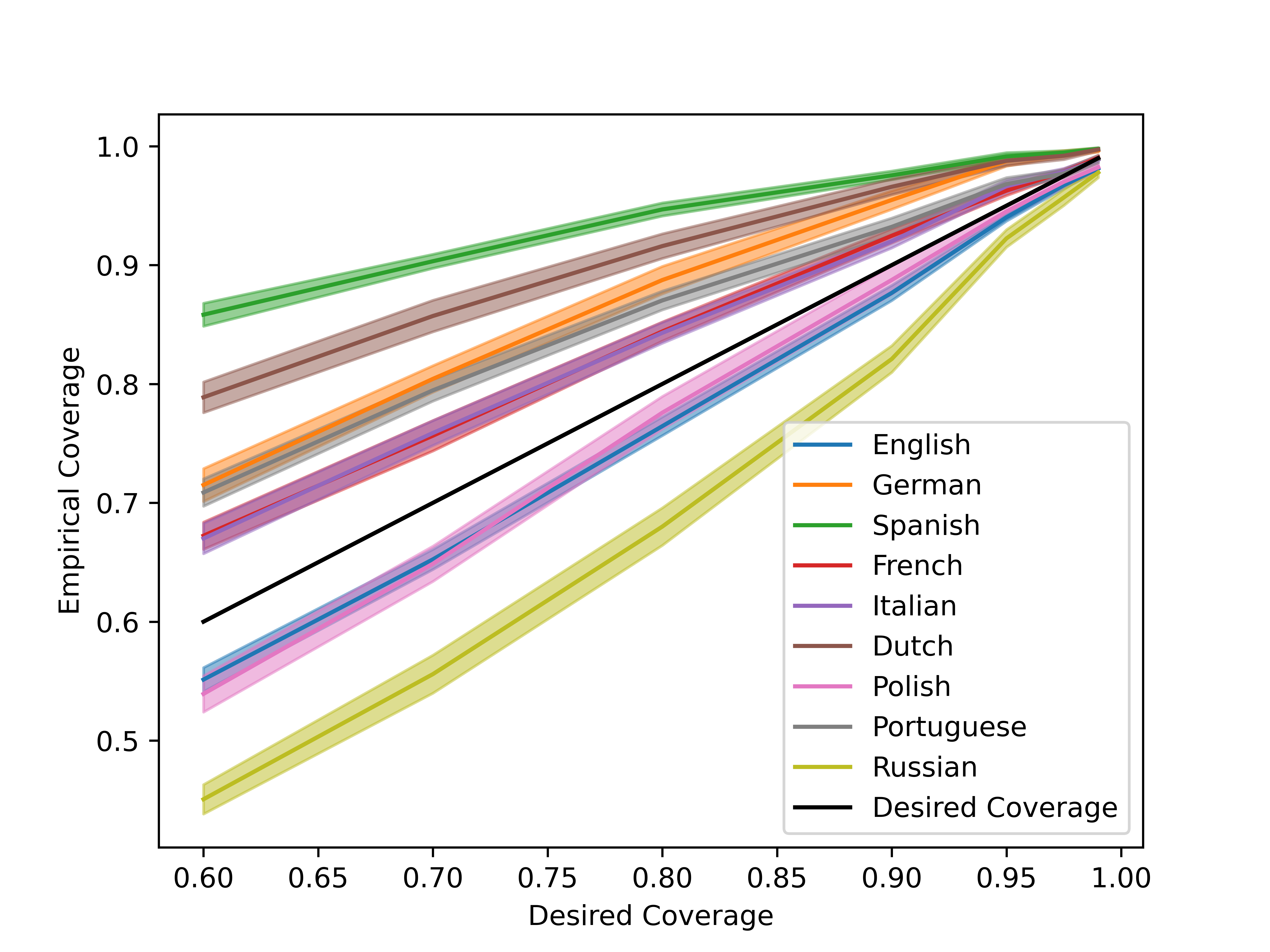}
    \caption{Initial calibration per language without adjustments for the Babelscape model on the multilingual WikiNEuRal dataset. The coverage depicted is calculated utilizing the \textbf{nc1} nonconformity score and displays the 95\% confidence interval produced by 20 iterations.}
    \label{fig:initial-language}
\end{figure}

However, despite good unconditional calibration, conditional miscalibration can occur when the data is stratified by confounding variables such as sequence length and language. Figure~\ref{fig:initial-language} illustrates how, when unaccounted for, different language groups exhibit systematic over- or under-coverage in the constructed prediction sets. As seen in Figure~\ref{fig:initial-language}, the Babelscape model, when trained on the Multilingual dataset and calibrated without language separation, tends to construct prediction sets with poor coverage for the Russian language while producing excessively large prediction sets for the Spanish language. Similarly, Figure~\ref{fig:initial-language} demonstrates how, when length is unaccounted for, sentence groups steadily trend from over-calibration to under-calibration as length increases.

\begin{figure}[t]
    \centering
    \includegraphics[width=0.8\linewidth]{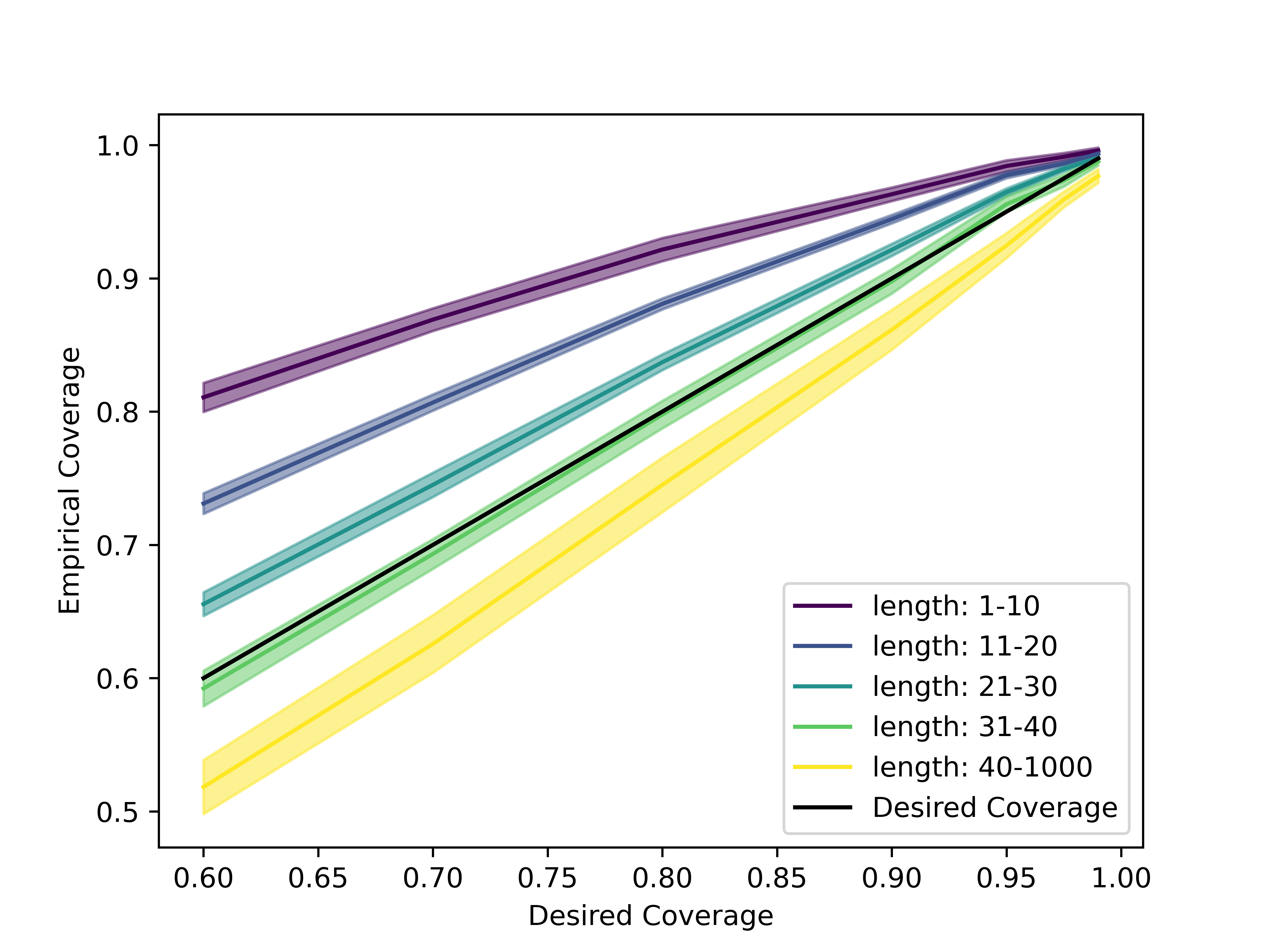}
    \caption{Initial calibration per sentence-length bin without adjustments for the Babelscape model on the multilingual WikiNEuRal dataset. The coverage depicted is calculated utilizing the \textbf{nc1} nonconformity score and displays the 95\% confidence interval produced by 20 iterations.}
    \label{fig:initial-length}
\end{figure}

\begin{figure}[t]
    \centering
    \includegraphics[width=0.8\linewidth]{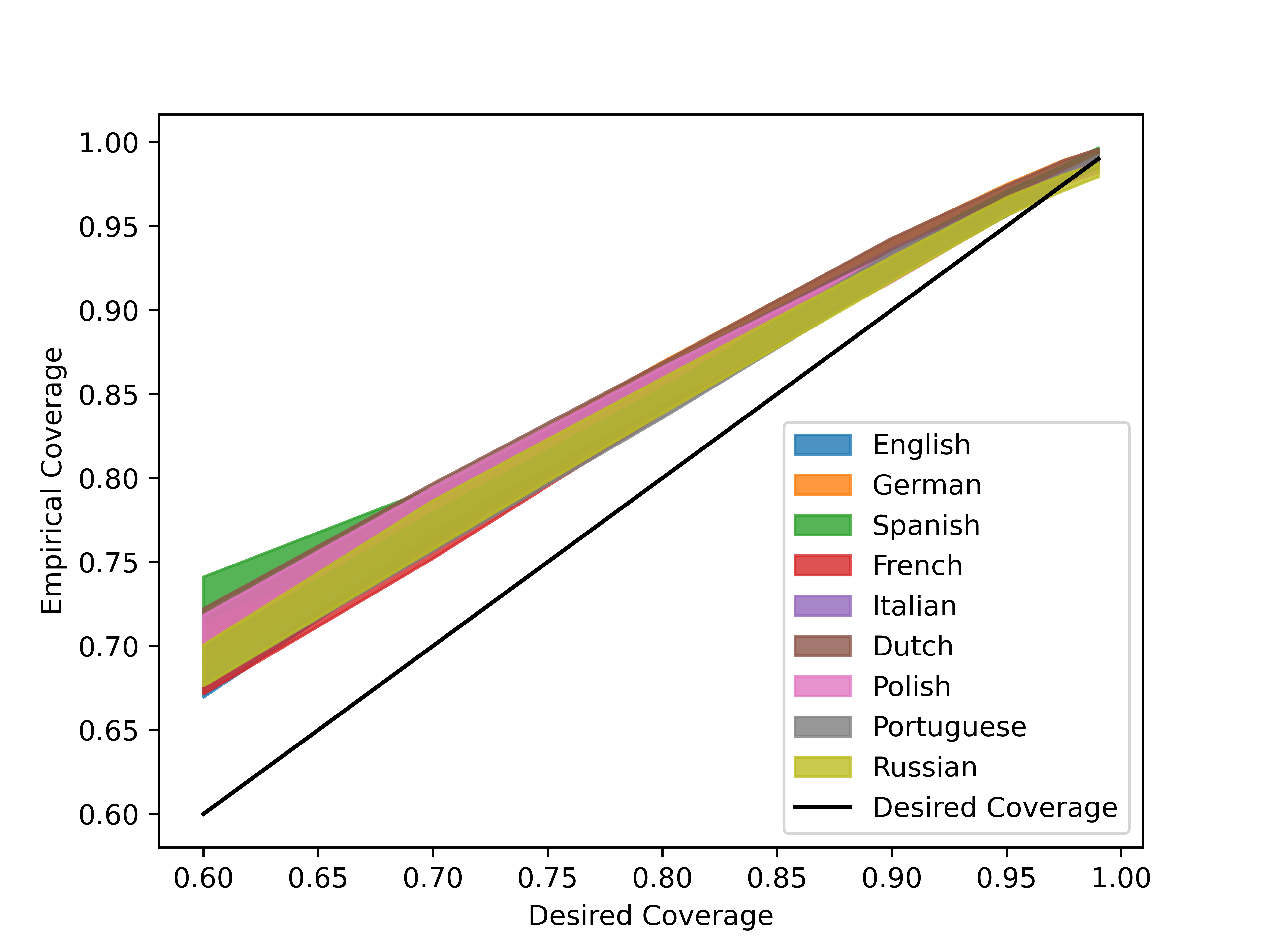}
    \includegraphics[width=0.8\linewidth]{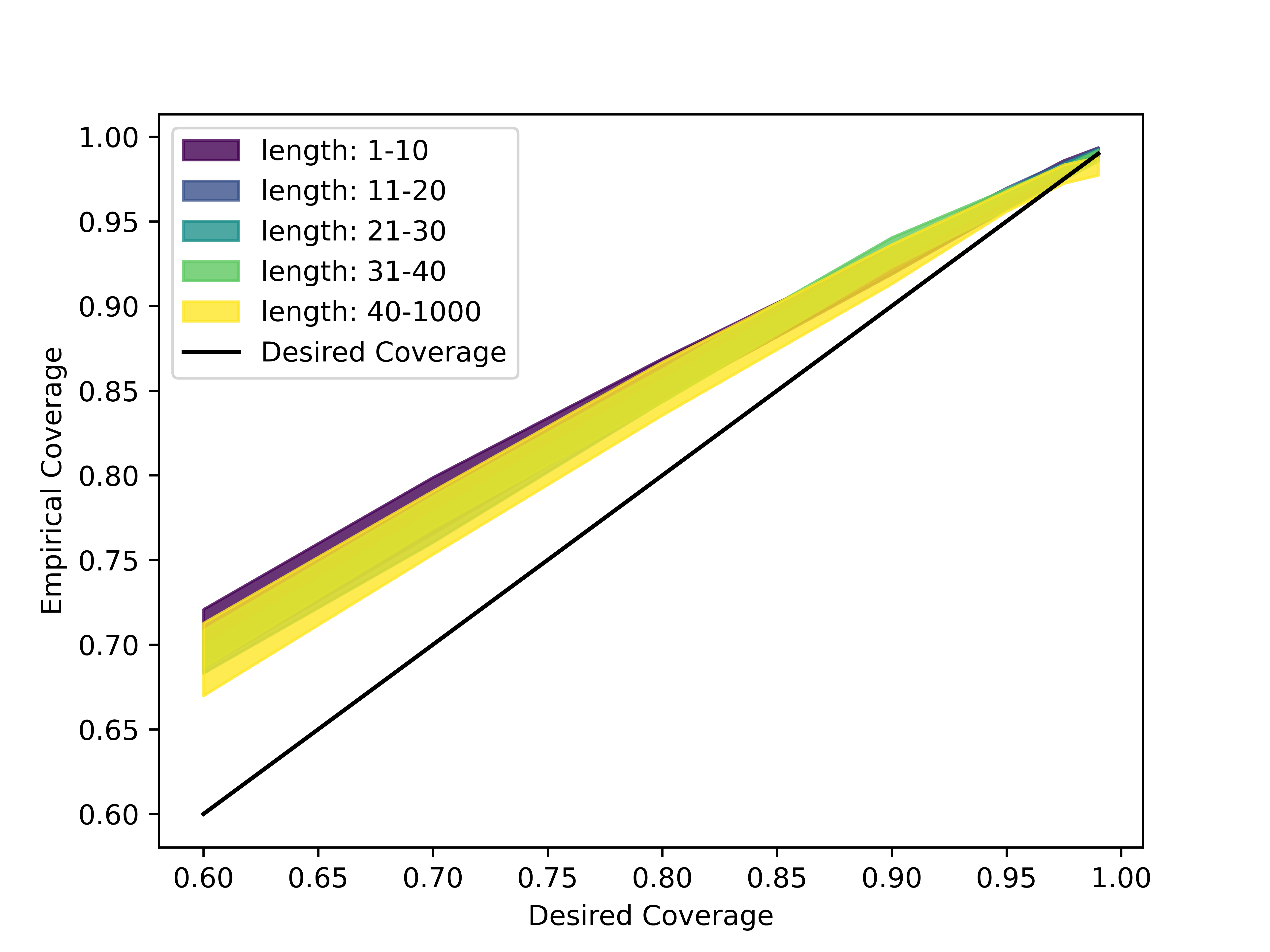}
    \caption{Stratified coverage across all languages (TOP) and sentence length bins (BOTTOM) in the multilingual WikiNEuRal benchmark dataset. The coverage depicted is calculated utilizing the conditional non-conformity score with \textbf{nc1} and displays the 95\% confidence interval produced by 20 iterations.} 
    %Right: KS test p-value matrix comparing non-conformity score distributions between length groups.}
    \label{fig:class-language}
\end{figure}

Although Figure~\ref{fig:initial-language} and Figure~\ref{fig:initial-length} display the misclassifications of language and length independently, when accounting for such confounding variables, it is possible to construct prediction sets that account for both language and length at the same time. Figure~\ref{fig:class-language} shows the recalibrated coverage across different sentence lengths and languages. As you can see, unlike Figures \ref{fig:initial-language} and \ref{fig:initial-length}, each sentence length and language groupings are well calibrated. We do note that the grouping procedure performed on the sentence length bins does ensure conditional coverage for each language-length bin combination, but does not ensure proper calibration within the bin. For example, the input length 19 may be under-calibrated because its cutoff threshold is based on all sentences from length 11-20, each of which has a slightly different non-conformity score distribution.

\begin{figure}[t]
    \centering
    \includegraphics[width=0.8\linewidth]{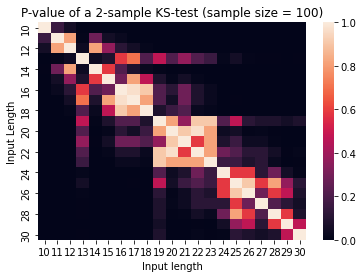}
    \caption{Kolmogorov-Smirnov between multiple 100-sample non-conformity scores from different sentence-length groups within the English language partition of the multilingual WikiNEuRal benchmark.} 
    %Right: KS test p-value matrix comparing non-conformity score distributions between length groups.}
    \label{fig:ks-tests}
\end{figure}

In order to visualize the difference in distributions between different sentence lengths, we plot a heatmap of p-values obtained from Kolmogorov–Smirnov (KS) tests comparing distributions of \textbf{nc1} non-conformity scores between sentence length groups in Figure \ref{fig:ks-tests}. Lighter cells indicate similar distributions (high p-values), while darker cells indicate statistically significant differences. To perform the Kolmogorov–Smirnov (KS) test, a sample of 100 observations was drawn from each input length group of size 10 to size 30. For each observation, the nonconformity score corresponding to the correct answer was computed. These scores were then used to conduct the KS test, and the resulting p-value was recorded. This entire process was repeated 100 times, and the average p-value across repetitions is presented in the heatmap above. The heatmap illustrates that, for most sentence lengths, an increase of as few as four words can significantly alter the distribution of nonconformity scores. Note: despite some values obtaining small p-values on the heatmap, we do know that each sentence length has a different distribution. The heatmap is meant as an illustrative tool to show how easily the differences in distributions are detected even with a limited sample size and, subsequently, limited power. We are not advising the use of this heatmap as a principled statistical test of equal distributions.

\subsection{Proof of Naive conformal prediction sets}

\begin{theorem}\label{thm:naive}
    let $C_{idx}$ and $C_{prob}$ be conformal prediction sets calibrated at levels $1-\alpha$ and $1-\beta$ then the intersection of these two prediction sets obtains a coverage above $1-\alpha - \beta$:

\begin{equation*}
    \mathbb{P}\left( \mathbf{y}_{\text{new}} \in C_{idx}(\mathbf{x}) \cap C_{prob}(\mathbf{x}) \right) \geq 1 - \alpha - \beta
\end{equation*}

\end{theorem}

\begin{proof}
    Let us construct two conformal prediction sets utilizing two distinct non-conformity scores $\mathbf{nc}$ and $\mathbf{nc^*}$ :
    
    \begin{align*} 
        \tau=& Q_{1-\alpha}\left(\{ \mathbf{nc}(\mathbf{y}_i \mid \mathbf{x}_i) \}_{i \in \mathcal{I}_{\text{cal}}} \right) \\
        =& \lceil (1 - \alpha)(1 + |\mathcal{I}_{\text{cal}}|) \rceil^{th}\text{ largest value of $\mathbf{nc}$ in } \mathcal{I}_{\text{cal}}
    \end{align*}
    
    \begin{align*} 
        \tau^*=& Q_{1-\beta}\left(\{ \mathbf{nc^*}(\mathbf{y}_i \mid \mathbf{x}_i) \}_{i \in \mathcal{I}_{\text{cal}}} \right) \\
        =& \lceil (1 - \beta)(1 + |\mathcal{I}_{\text{cal}}|) \rceil^{th}\text{ largest value of $\mathbf{nc^*}$ in } \mathcal{I}_{\text{cal}}
    \end{align*}
    
    Prediction sets for each quantile and non-conformity score are constructed via:
    
    \begin{equation*} \label{eq: prediction set 1}
        C(\mathbf{x},\tau) = \{\mathbf{y} : \mathbf{nc}(\mathbf{y}|\mathbf{x_i}) \leq \tau \}
    \end{equation*}
    
    \begin{equation*} \label{eq: prediction set 2}
        C^*(\mathbf{x},\tau^*) = \{\mathbf{y} : \mathbf{nc}^*(\mathbf{y}|\mathbf{x_i}) \leq \tau^* \}
    \end{equation*}
    
    Via Proposition 1:
    
    \begin{equation*}
        \mathbb{P}(\mathbf{y}_{n+1} \in C(\mathbf{x}_{n+1},\tau)) \geq 1-\alpha
    \end{equation*}
    
    \begin{equation*}
        \mathbb{P}(\mathbf{y}_{n+1} \in C^*(\mathbf{x}_{n+1},\tau^*)) \geq 1-\beta
    \end{equation*}
    
    Following the basic law of probability that $P(A \cap B) \ge P(A) + P(B) - 1$, and then define the events: $A = \{\mathbf{y}_{\text{new}} \in C_{idx}(\mathbf{x})\}$ and $B = \{\mathbf{y}_{\text{new}} \in C_{prob}(\mathbf{x})\}$:
    
    \begin{align*}
        \mathbb{P}&\big[(A\cap B\big] \\
        =&\mathbb{P}(A) + \mathbb{P}(B)  -\mathbb{P}\big[A \cup B\big]  \\
        \geq&\mathbb{P}(A) + \mathbb{P}(B)  - 1  \geq 1-\alpha-\beta
    \end{align*}
\end{proof}

\subsection{Proof of Conditional conformal prediction sets}

\begin{theorem}\label{thm:conditional}
    Let:
    $C_{\text{idx}}(\mathbf{x})$,
    be a conformal prediction set such that the desired coverage is $1-\alpha$. We may then define a second prediction set ($C_{prob}(\mathbf{x})$)such that:
    \[
    \mathbb{P}(\mathbf{y}_{\text{new}} \in C_{\text{prob}}(\mathbf{x}) \mid \mathbf{y}_{\text{new}} \in C_{\text{idx}}(\mathbf{x})) \geq 1 - \beta
    \]
    We propose that the coverage of the intersection of these two sets is bounded by:
    \[
    \mathbb{P}(\mathbf{y}_{\text{new}} \in C_{\text{idx}}(\mathbf{x}) \cap C_{\text{prob}}(\mathbf{x})) \geq (1 - \alpha)(1 - \beta)
    \]

\end{theorem}

\begin{proof}
    Let us construct two conformal prediction sets utilizing a baseline non-conformity score and a secondary non-conformity score. The quantile used to calculate the baseline prediction set is as follows:
    
    \begin{align*} 
        \tau &= Q_{1-\alpha}\left(\{ \mathbf{nc}(\mathbf{y}_i \mid \mathbf{x}_i) \}_{i \in \mathcal{I}_{\text{cal}}} \right)\\ & = \lceil (1 - \alpha)(1 + |\mathcal{I}_{\text{cal}}|) \rceil^{th}\text{ largest value of $\mathbf{nc}$ in } \mathcal{I}_{\text{cal}}
    \end{align*}
    
    Now, define the subset of the calibration data below $\tau$ as $\mathcal{I}_{cal,\tau} = \{i : \mathbf{nc}(\mathbf{y}_i|\mathbf{x}_i) \leq \tau,i \in \mathcal{I}_{cal}\}$. Now define the secondary quantile as follows:
    
    \begin{align*} 
        \tau^*&= Q_{1-\beta}\left(\{ \mathbf{nc}^*(\mathbf{y}_i \mid \mathbf{x}_i) \}_{i \in \mathcal{I}_{cal,\tau}} \right) \\&= \lceil (1 - \beta)(1 + |\mathcal{I}_{cal,\tau}|) \rceil^{th}\text{ largest value of $\mathbf{nc}$ in } \mathcal{I}_{cal,\tau}
    \end{align*}
    
    Prediction sets for each quantile and non-conformity score are constructed via:
    
    \begin{equation} \label{eq: prediction set baseline}
        C(\mathbf{x},\tau) = \{\mathbf{y} : nc(\mathbf{y}|\mathbf{x_i}) \leq \tau \}
    \end{equation}
    
    \begin{equation} \label{eq: prediction set subsequence}
        C^*(\mathbf{x},\tau^*) = \{\mathbf{y} : nc^*(\mathbf{y}|\mathbf{x_i}) \leq \tau^* \}
    \end{equation}
    
    Following Proposition 1:
    \begin{equation*}
        \mathbb{P}(\mathbf{y}_{n+1} \in C(\mathbf{x}_{n+1},\tau)) \geq 1-\alpha
    \end{equation*}
    
    Additionally, by construction:
    
    \begin{align*}
        &\mathbb{P}\bigg[\mathbf{y}_{n+1} \in C^*(\mathbf{x}_{n+1},\tau^*) \bigg|\mathbf{y}_{n+1} \in C(\mathbf{x}_{n+1},\tau)\bigg] \\&= \mathbb{P}\bigg[nc^*(\mathbf{y}_{n+1}|\mathbf{x}_{n+1}) \leq \tau^* \bigg| nc(\mathbf{y}_{n+1}|\mathbf{x}_{n+1}) \leq \tau)\bigg]
    \end{align*}
    
    Now, let $E$ be the event space of all possible NER inputs and outputs $(\mathbf{x},\mathbf{y})$. Let us partition this event space into two regions $\{E_1,E_2\}$, where $E_1 = \{(\mathbf{x},\mathbf{y}) : \mathbf{nc}(\mathbf{y}|\mathbf{x}) \leq \tau\}$ and $E_2 = \{(\mathbf{x},\mathbf{y}) : \mathbf{nc}(\mathbf{y}|\mathbf{x}) > \tau\}$. If the new observation $(\mathbf{x_{n+1}},\mathbf{y_{n+1}})\in E_1$, then $\{\mathbf{nc}^*( \mathbf{x}_i, \mathbf{y}_i)\}_{i\in \mathcal{I}_{\text{cal}}^1\cup[n+1] } =\{\mathbf{nc}^*( \mathbf{x}_i, \mathbf{y}_i)\}_{i\in \mathcal{I}_{\text{cal},\tau}\cup[n+1] }$ is exchangeable. Therefore, by Theorem \ref{thm:stratified}:
    
    \begin{equation*}
        \mathbb{P}\bigg[nc^*(\mathbf{y}_{n+1}|\mathbf{x}_{n+1}) \leq \tau^* \bigg| nc(\mathbf{y}_{n+1}|\mathbf{x}_{n+1}) \leq \tau\bigg] \geq 1-\beta
    \end{equation*}
    
    Theorem~\ref{thm:conditional} now follows from the following basic probability result: for any two events $A$ and $B$,
    $$
    P(A \cap B) = P(A) P(B|A).
    $$
    Where $A = \{\mathbf{y}_{\text{n+1}} \in C(\mathbf{x},\tau)\}$ and $B=\{\mathbf{y}_{\text{n+1}} \in C^*(\mathbf{x},\tau^*)\}$. By definition $P(A) \geq  (1 - \alpha)$ and $P(B|A) \geq (1 - \beta)$.
    
    Now, Let $C_{\text{idx}}(\mathbf{x}) = C(\mathbf{x},\tau)$ and $C_{\text{prob}}(\mathbf{x}) = C^*(\mathbf{x},\tau^*)$. Then it is clear that

    \[
    \mathbb{P}(\mathbf{y}_{\text{new}} \in C_{\text{idx}}(\mathbf{x}) \cap C_{\text{prob}}(\mathbf{x})) \geq (1 - \alpha)(1 - \beta).
    \]

\end{proof}

\section{Supplement for Section 5: Subsequence Conformal Prediction}

\subsection{Additional Subsequence Example}

\begin{figure}[t]
    \centering
    \includegraphics[width=0.8\linewidth]{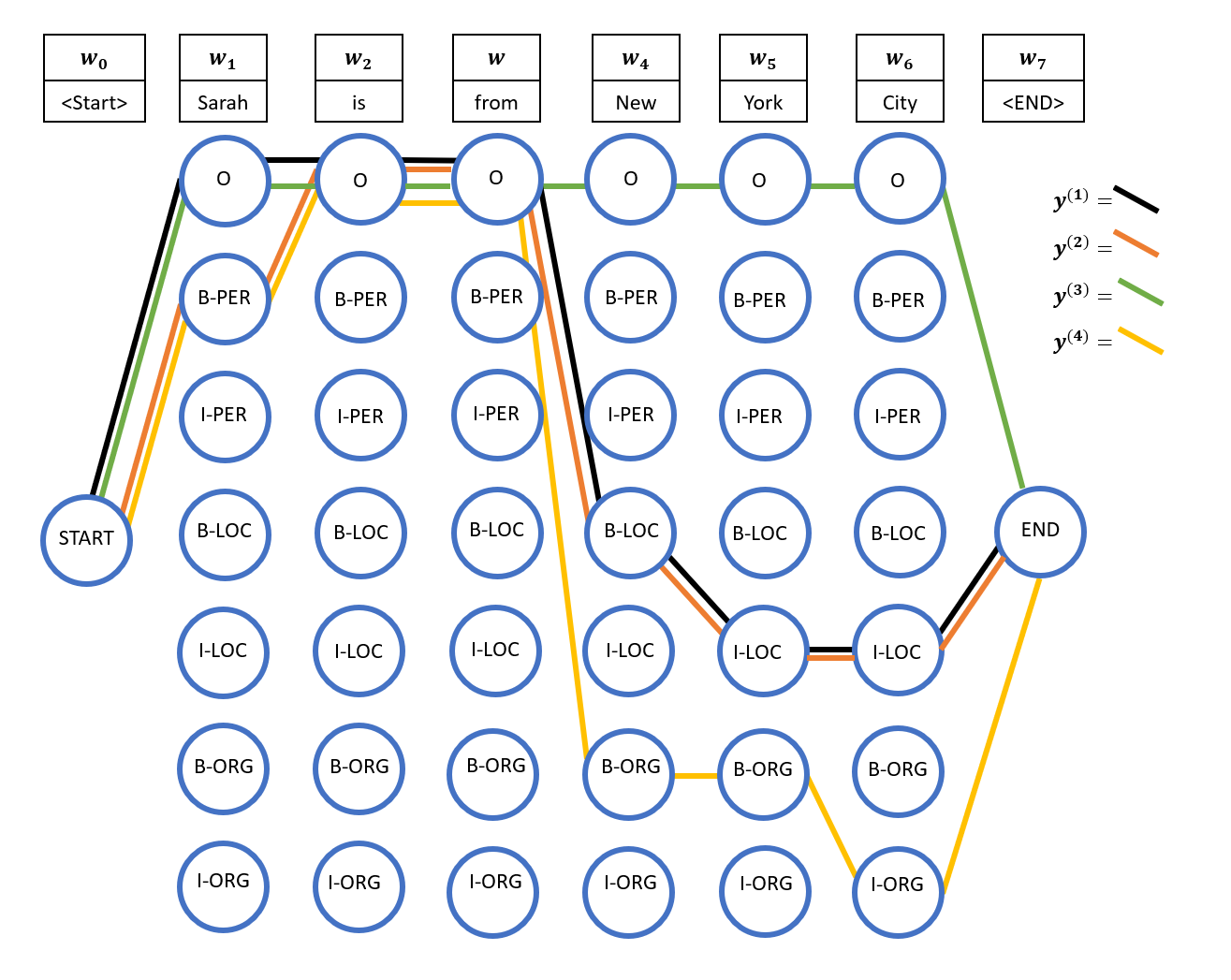}
    \caption{CRF decoding from table \ref{tab:sent_pred_set} represented as paths through a graph. The input words are denoted at the top of the figure, whereas each blue circle denotes the selected label for the given word at each time step. Each of the top four label sequences is depicted via a colored line traversing the graph.}
    \label{fig:decoding_example}
\end{figure}

 Figure \ref{fig:decoding_example} displays a toy example of the decoding strategy utilized in the NER model. The top four decoded sequences (k=4) for the NER output are displayed as different colored lines traveling from label to label. Utilizing the figure, we can see how the probability of any individual subsequence of labels is equal to the sum of the overlapping line probabilities. For demonstration purposes, Figure \ref{fig:decoding_example} utilizes a hyperparameter selection of $k=4$, such that we simplify the probability distribution such that 100\% of the assigned probability is attributed to the first four sequences. The choice of k is determined by the maximum desired coverage of the model on a specific benchmark. In practice, we utilize $k=100$ for the results sections of this paper.

Using the previous example of ``\textit{Sarah is from New York City}'', the probability that the subsequence ``\textit{Sarah is from}'' is labeled as \{\texttt{B-PER} \texttt{O} \texttt{O} \}, is equal to $\mathbf{y}^{(2)} + \mathbf{y}^{(4)}$. The above example has two correct NER entities ``\textit{sarah}'' and ``\textit{New York City}'' with respective labels of \{\texttt{B-PER}\} and \{\texttt{B-LOC} \texttt{I-LOC} \texttt{I-LOC} \}. Using the Figure, it is clear that the model probability assigned to the two correct NER entities is $\mathbf{y}^{(2)} + \mathbf{y}^{(4)}$ and $\mathbf{y}^{(1)} + \mathbf{y}^{(2)}$ respectively.

\subsection{Proof of Subsequence conformal prediction sets}

\begin{theorem}\label{thm: subseq}
    let $C_{ent}(\mathbf{x},\mathbf{\tau}_{\mathcal{W}},a,b)$ be a subsequence conformal prediction set such that:

    $$ C_{ent}(\mathbf{x},\mathbf{\tau}_{\mathcal{W}},a,b) = \bigcup_{w\in \mathcal{W}} C_{w,ent}(x,\tau_w,a,b)$$

    and
    $$
    C_{w,ent}(\mathbf{x}, \tau_w,a,b) = \{ w \text{ if } nc_{ent}(\mathbf{y} \mid \mathbf{x},w,a,b) \leq \tau_w \}.
    $$

    Where $nc_{ent}(\mathbf{y} \mid \mathbf{x},w,a,b)$ is an entity-level non-conformity score and $\tau_w$ is a class-specific cutoff from the $1-\alpha$ quantile of the non-conformity score on $\mathcal{I}_{cal}^w$, the w-class specific entity calibration data. Then the prediction set $C_{ent}(\mathbf{x},\mathbf{\tau}_{\mathcal{W}},a,b)$ is well calibrated such  that:

    $$    \forall \text{ }  w \in \mathcal{W}{}, \text{    } \mathbb{P}\big(\mathbf{y}_{a:a+b} \in C_{ent}(x,\tau_{\mathcal{W}},a,b)| \mathbf{y}_{a:a+b} = w\big) \geq 1-\alpha$$
\end{theorem}

\begin{proof}
    Define a mutually exclusive and exhaustive partition of the subsequence sample space based on the subsequence class label such that $\bigcup_{j=0}^{|\mathcal{W}|} E_j = E, \mathbb{P}(\bigcup_{j=0}^{|\mathcal{W}|} E_j) = 1$ and $\forall j \neq k, E_j \cap E_k = \varnothing, \mathbb{P}(E_j \cap E_k) = 0$. Where $E_0$ represents the non-class subsection and $E_1$ through $E_{|\mathcal{W}|}$ represent a separate partition for each class $w \in |\mathcal{W}|$

    Let $\mathbf{y}_{a:a+b}$ be a entity subsequence of $\mathbf{y}_{n+1}$ .Given that by definition the events $\mathbf{y}_{a:a+b} = w$ and $(\mathbf{x_{n+1}},\mathbf{y_{n+1}})\in E_w$ are equivalent, Theorem~\ref{thm:conditional} has been satisfied and therefore:

    $$ \forall \text{ }  w \in \mathcal{W}{}, \text{    } \mathbb{P}\big(\mathbf{y}_{a:a+b} \in C_{ent}(x,\tau_{\mathcal{W}},a,b)| \mathbf{y}_{a:a+b} = w\big) \geq 1-\alpha$$.
    
\end{proof}

\subsection{Algorithm for subsequence conformal prediction}

The following algorithm depicts subsequence conformal prediction for one subsequence. Applying this algorithm to all possible subsequences simply requires iterating over the set of all subsequences. Notably, significant time is saved due to the top-$k$ approximation, which lets us ignore any subsequence that only maps to the non-entity class within the returned $k$ full-sequences.

\begin{algorithm}[t]
\DontPrintSemicolon
\KwIn{
    Training and calibration data $\mathcal{I}_{\text{train}}, \mathcal{I}_{\text{cal}}$; \\
    Test input $\mathbf{x}_{\text{test}}$; \\
    Confidence level $1-\alpha$; \\
    Non-conformity score $nc_{ent}(\mathbf{x}, w,a,b)$; \\
    Label set $\mathcal{W}$;\\
    Class Mapping function $g$;\\
    Number of considered output sequences $K$; \\
    Subsequence bounds $a,b$
}
\KwOut{Subsequence prediction set $C_{ent}(\mathbf{x},\tau_{\mathcal{W}}, a,b )$ for the subsequence $\mathbf{y}_{a:a+b}$}
Train CRF-based model $M$ on $\mathcal{I}_{\text{train}}$\;

\ForEach{$w \in \mathcal{W}$}{
    Compute class-specific quantile threshold $\tau_w \gets \hat{Q}_{1 - \alpha}(\{nc_{ent}(\mathbf{y}\mid \mathbf{x}, w),a,b\}_{\mathbf{y}_{a:a+b} \in \mathcal{I}_{\text{cal}}^l})$\;
}

Initialize output prediction set $C \gets \emptyset$\;

\ForEach{$w \in \mathcal{W}$}{
    Compute $nc_{ent}(\mathbf{y}_{a:a+b} \mid \mathbf{x}_{\text{test}}, w)$ using top-$K$ sequences\;
    \If{$nc_{ent}(\mathbf{y} \mid \mathbf{x}, w,a,b) \geq \tau_w$}{
        Draw $u \sim \text{Uniform}(0,1)$\;
        Compute adaptive threshold $v \gets V(\mathbf{y}_{a:a+b}, \tau_w, \alpha)$\;
        \If{$u \leq v$}{
            $C \gets C \cup \{ w\}$\;
        }
    }
}

\Return{$\mathcal{C}$}\;
\caption{{\sc Subsequence Conformal Prediction}}
\label{algo:entity}
\end{algorithm}

\section{Supplement for Section 6: Integrated Conformal Prediction}

\subsection{Proof of integrated conformal prediction}

\begin{prop}\label{prop:integrated}
    For any sequence $y^{(i)}$, let $E(\mathbf{y}^{(i)})$ be a set of subsequences such that:
    
    $$ y^{(i)}_{a:a+b} \in E(\mathbf{y}^{(i)}) \text{ iff } \mathbf{y}^{(i)}_{a:a+b} \neq \texttt{Non-Entity}$$
    
    Additionally, let the set of all continuous subsequences be given by: 
    
    \begin{equation*}
        S(\mathbf{x}) = \{(a,b) : a+b \leq len(x), a \geq 1, b \geq 0 \} 
    \end{equation*}  

    And let  the subsequence prediction sets $C_{ent}(\mathbf{x},\mathbf{\tau}_{\mathcal{W}},a,b)$ be constructed with a coverage of $(1-\alpha)^{1/s}$ where $s$ is the number of entities within the full labeled sequence. Then if $s$ is correct, the prediction set
    
    \begin{equation*}
        C_{int}(\mathbf{x},\tau_{\mathcal{W}}) = \{\mathbf{y}^{(i)}: E(\mathbf{y}^{(i)}) \subseteq \bigcup_{(a,b) \in S(\mathbf{x})} C_{ent}(\mathbf{x},\mathbf{\tau}_{\mathcal{W}},a,b)\}
    \end{equation*}

    is valid at a confidence level of $1 -\alpha.$
\end{prop}

\begin{proof}
    Given a new observation  $(\mathbf{x}_{n+1},\mathbf{y}_{n+1})$ with $s$ distinct subsequence entities $\mathbf{y}_{a_i:a_i + b_i}$.

    Via definition and Theorem~\ref{thm: subseq}:

    $$ \mathbb{P}(\mathbf{y}_{a:a + b} \in   C_{ent}(\mathbf{x},\mathbf{\tau}_{\mathcal{W}},a,b)) \geq (1-\alpha)^{1/s}$$

    Also note that by definition $\mathbf{y}^{(i)} \in C_{int}$ iff:

    $$ E(\mathbf{y}_{n+1}) \subseteq \bigcup_{(a,b) \in S(\mathbf{x})} C_{ent}(\mathbf{x},\mathbf{\tau}_{\mathcal{W}},a,b)$$

    which is equivalent to stating that for all $s$ true entity subsequences of $\mathbf{y}_{n+1}$ ($1 \leq i \leq s)$:

    $$ \mathbf{y}_{a_i:a_i + b_i} \in   C_{ent}(\mathbf{x},\mathbf{\tau}_{\mathcal{W}},a_i,b_i) $$

    Therefore, via the intersection of prediction sets:

    $$\mathbb{P}(\mathbf{y}_{n+1} \in C_{int}(\mathbf{x},\tau_{\mathcal{W}})) = \mathbb{P}(\bigcup_{i=1}^{s} \bigg[\mathbf{y}_{a_i:a_i+b_i} \in C_{ent}(\mathbf{x},\mathbf{\tau}_{\mathcal{W}},a_i,b_i)\bigg]) \geq [(1-\alpha)^{1/s}]^s = (1-\alpha) $$.
    
\end{proof}

\subsection{Proof of index-entity conformal prediction}

\begin{prop}\label{prop:integrated-index-entity}
    For any candidate sequence $y^{(i)}$, let $E(\mathbf{y}^{(i)})$ be a set of subsequences such that:
    
    $$ y^{(i)}_{a:a+b} \in E(\mathbf{y}^{(i)}) \text{ iff } \mathbf{y}^{(i)}_{a:a+b} \neq \texttt{Non-Entity}$$
    
    Additionally, let the set of all continuous subsequences be given by: 
    
    \begin{equation*}\label{eq: nontrivial bounds_ref}
        S(\mathbf{x}) = \{(a,b) : a+b \leq len(x), a \geq 1, b \geq 0 \} 
    \end{equation*}  

    And let the subsequence prediction sets $C_{ent,idx}(\mathbf{x},\mathbf{\tau}_{\mathcal{W}},a,b)$ be constructed with a coverage of $(1-\alpha)^{1/s}$ where $s$ is the number of entities within the full labeled sequence. Then, if $s$ is correct, the prediction set
    
    \begin{equation*}
        C_{idx}(\mathbf{x},\tau_{\mathcal{W}}) = \{\mathbf{y}^{(i)}: i \leq max\bigg[ \bigcup_{(a,b) \in S(\mathbf{x})} C_{ent,idx}(\mathbf{x},\mathbf{\tau}_{\mathcal{W}},a,b)\bigg]\}
    \end{equation*}

    is valid at a confidence level of $1 -\alpha.$
\end{prop}

\begin{proof}
     Given a new observation  $(\mathbf{x}_{n+1},\mathbf{y}_{n+1})$ with $s$ distinct subsequence entities $\mathbf{y}_{a_i:a_i + b_i}$.

    Via definition and Theorem~\ref{thm: subseq}:

    $$ \mathbb{P}(i \in   C_{ent,idx}(\mathbf{x},\mathbf{\tau}_{\mathcal{W}},a,b)) \geq (1-\alpha)^{1/s}$$

    Following similar logic to the proof of Proposition~\ref{prop:integrated} we note that by definition $\mathbf{y}^{(i)} \in C_{idx}$ iff:

    $$ i \leq max(\bigcup_{(a,b) \in S(\mathbf{x})} C_{ent,idx}(\mathbf{x},\mathbf{\tau}_{\mathcal{W}},a,b))$$

    Because of the maximum operation, unlike Proposition~\ref{prop:integrated}, $i$ only has to exist in the largest prediction set. However, we can utilize a stricter statement stating that for all $s$ true entity subsequences $(1 \leq j \leq s)$:

    $$ i \in   C_{ent,idx}(\mathbf{x},\mathbf{\tau}_{\mathcal{W}},a_j,b_j) $$

    Therefore, via the intersection of prediction sets, let the correct response $\mathbf{y}_{n+1}$ be the $i'th$ ranked model output:

    $$\mathbb{P}(\mathbf{y}^{(i)} \in C_{idx}(\mathbf{x}_{n+1},\tau_{\mathcal{W}})) \geq \mathbb{P}(\bigcup_{j=1}^{s} \bigg[i \in C_{ent,idx}(\mathbf{x},\mathbf{\tau}_{\mathcal{W}},a_j,b_j)\bigg]) \geq [(1-\alpha)^{1/s}]^s = (1-\alpha) $$.
\end{proof}

\subsection{Additional Figures relating to integrated-conformal prediction }

\begin{figure}[t]
    \centering
    \includegraphics[width=0.8\linewidth]{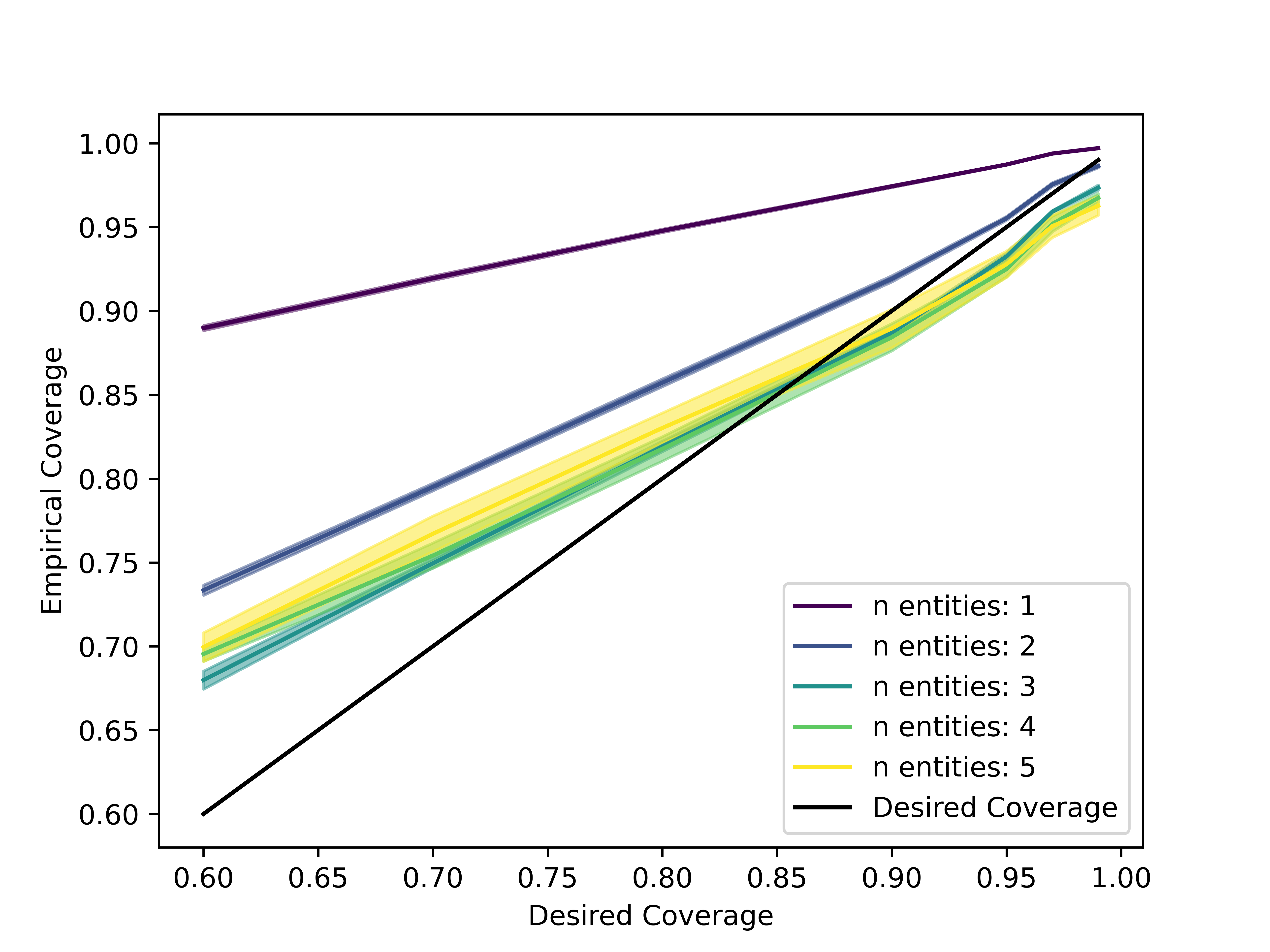}
    \caption{ Calibrated integrated (sentence + entity) method as a function of the number of NER entities in a sentence without the Šidák correction. The empirical coverage displayed is the coverage of the \textbf{nc1} non-conformity score when evaluated on the multilingual WikiNEuRal benchmark and the Bablescape model. }
    \label{fig:n_ent}
\end{figure}

Figure \ref{fig:n_ent} displays the empirical coverage of \textbf{nc1} on the WikiNEuRal benchmark with the Babelscape model for the integrated prediction set without a family-wise error correction. As shown, the coverage for sentences with one NER entity is as desired, but as the number of NER entities within a sentence increases, the empirical coverage decreases relatively.

\begin{figure}[t]
    \centering
    \includegraphics[width=0.8\linewidth]{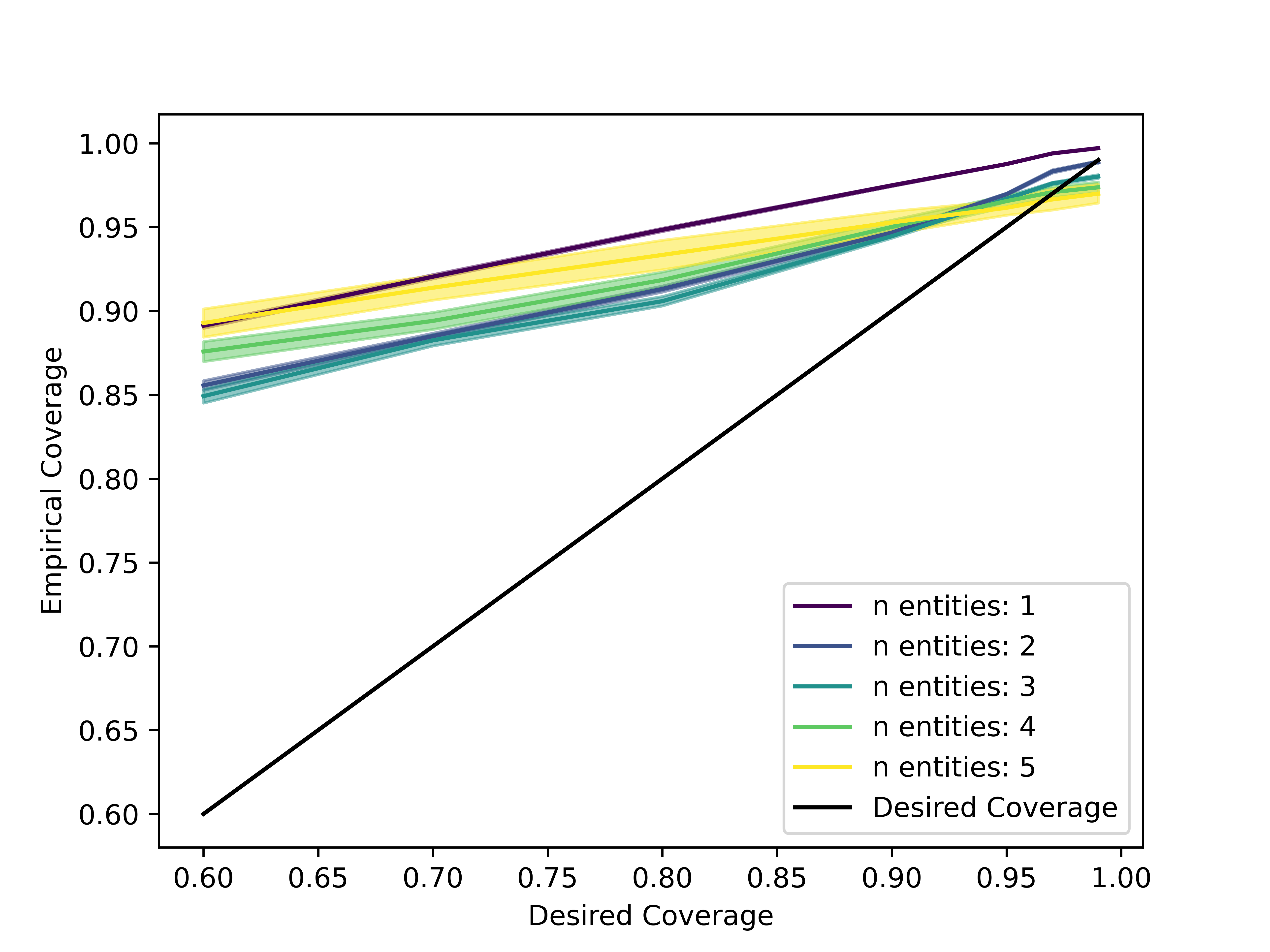}
    \caption{ Calibrated integrated method as a function of the number of NER entities with the Šidák correction. The empirical coverage displayed is the coverage of the \textbf{nc1} non-conformity score when evaluated on the multilingual WikiNEuRal benchmark and the Bablescape model. }   \label{fig:n_ent_sidak}
\end{figure}

Figure~\ref{fig:n_ent_sidak} displays the empirical coverage of the integrated-prediction set method after applying the Šidák correction. The Šidák correction consistently achieves coverage above the desired level, regardless of entity count. However, it is worth noting that the Šidák correction becomes overly conservative as the number of entities increases. In such cases, a less aggressive family-wise error control method may be more appropriate. Coverage lines that drop below the diagonal occur due to the top-$k$ limitation applied to the model.

\begin{figure}[t]
    \centering
    \includegraphics[width=0.8\linewidth]{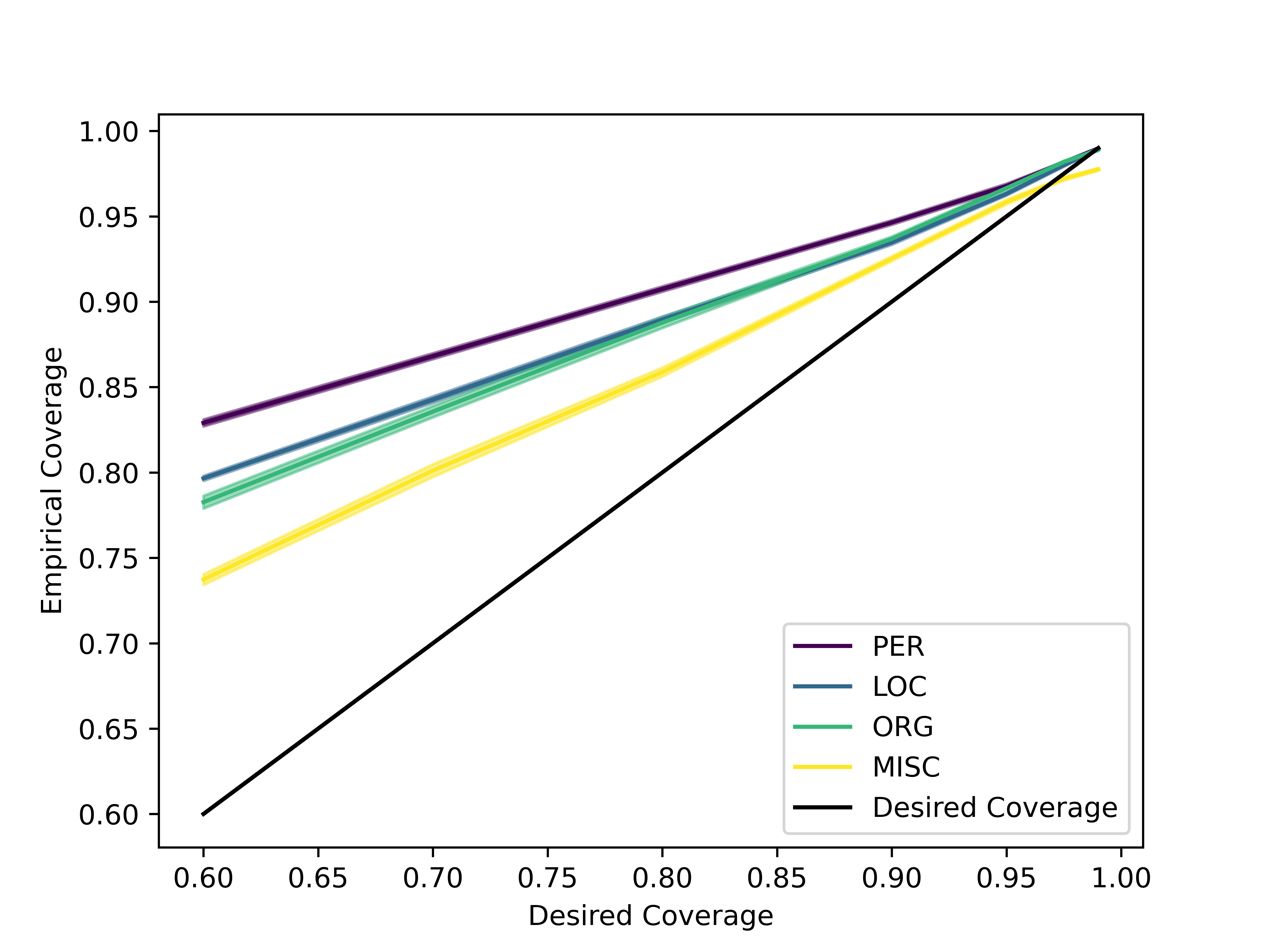}
    \caption{class-conditional calibration of the integrated (sentence + entity) method with \textbf{nc1} and the Šidák correction. Data was computed on the WikiNEuRal benchmark dataset with the Babelscape model}
    \label{fig:Class-calibration}
\end{figure}

Figure~\ref{fig:Class-calibration} illustrates the effect of this integrated method on full-sentence predictions with class calibration. With Figure~\ref{fig:Class-calibration} and Figure~\ref {fig:n_ent_sidak}, we have shown that the integrated method produces valid prediction sets regardless of the sentence length, language, number of entities, and class typing.

\section{Supplement for Section 7: Experiments}
The following supplemental material details more in-depth graphs containing results for coverages of 80\%, 90\%, 95\%, and 97.5\%. Initial results were limited to only 95\% coverage due to paper length. 

\subsection{Supplement for Section 7.3: Calibrating Non-conformity scores}

As previously described in Sections~\ref{sec:class conditional sentence}-\ref{sec:sent+ent}, we proposed multiple methods for calibrating conformal prediction sets to account for confounding variables such as language, sequence length, and number of entities. In those sections, brief graphical results were presented to demonstrate the effectiveness of the calibration methodologies. The following three tables further show the specific changes in prediction set coverage and average size of the prediction sets formed via conformal prediction as a result of calibration. For ease of comparison, all tables were constructed utilizing the Babelscape model on the multilingual WikiNEural benchmark dataset.

\begin{table*}[t]
\centering
\scriptsize
\resizebox{\textwidth}{!}{%
\begin{tabular}{llcccc}
\toprule
\textbf{Length Range} & \textbf{Desired Cov.} 
& \multicolumn{2}{c}{\textbf{No Calibration}} 
& \multicolumn{2}{c}{\textbf{Length Calibration}} \\
\cmidrule(lr){3-4}\cmidrule(lr){5-6}
& &  \textbf{Emp. - Desired Cov.} & \textbf{Average Set Size} & \textbf{Emp. - Desired Cov.} & \textbf{Average Set Size} \\
\midrule
1–10    & 0.8   & 0.1334 & 0.9901 & 0.0530 & \textbf{0.8696} \\
1–10    & 0.9   & 0.0612 & 1.3265 & 0.0289 & \textbf{1.0603} \\
1–10    & 0.95  & 0.0341 & 1.8184 & 0.0150 & \textbf{1.3032} \\
1–10    & 0.975 & 0.0159 & 2.0902 & 0.0072 & \textbf{1.7067} \\
\midrule
11–20   & 0.8   & 0.0861 & 0.9431 & 0.0510 & 0.9071 \\
11–20   & 0.9   & 0.0448 & 1.3098 & 0.0309 & \textbf{1.1792} \\
11–20   & 0.95  & 0.0284 & 1.8396 & 0.0174 & \textbf{1.5684} \\
11–20   & 0.975 & 0.0123 & 2.1584 & 0.0089 & \textbf{2.0120} \\
\midrule
21–30   & 0.8   & 0.0418 & 0.8986 & 0.0506 & 0.9737 \\
21–30   & 0.9   & 0.0231 & 1.2913 & 0.0302 & 1.3452 \\
21–30   & 0.95  & 0.0162 & 1.8445 & 0.0157 & 1.8191 \\
21–30   & 0.975 & 0.0050 & 2.2510 & 0.0059 & \textbf{\textit{2.8533}} \\
\midrule
31–40   & 0.8   & 0.0074 & 0.8668 & 0.0528 & \textbf{\textit{1.0600}} \\
31–40   & 0.9   & 0.0139 & 1.2743 & 0.0315 & \textbf{\textit{1.4416}} \\
31–40   & 0.95  & 0.0016 & 1.8406 & 0.0119 & \textbf{\textit{2.2697}} \\
31–40   & 0.975 & 0.0004 & 2.3104 & 0.0048 & \textbf{\textit{3.5587}} \\
\midrule
40+ & 0.8   & \textbf{-0.0372} & 0.8250 & 0.0492 & \textbf{\textit{1.1949}} \\
40+ & 0.9   & \textbf{-0.0265} & 1.2535 & 0.0240 & \textbf{\textit{1.8985}} \\
40+ & 0.95  & \textbf{-0.0170} & 1.8415 & 0.0076 & \textbf{\textit{6.0854}} \\
40+ & 0.975 & \textbf{-0.0213} & 2.4251 & 0.0009 & \textbf{\textit{24.2288}} \\
\bottomrule
\end{tabular}
}
\caption{Model calibration for five sequence length bins on the WikiNEuRal benchmark. Coverage columns are bolded if the empirical coverage (Emp. Cov.) is less than the desired coverage (Desired Cov.). Set size columns are bolded if there is a decrease of greater than .1 in average set size and bolded + italicized if there is an increase of greater than .1 in average set size.}
\label{tab:length calibratriton_full}
\end{table*}

Table~\ref{tab:length calibratriton_full} shows the effectiveness of calibrating the full-sequence prediction sets based on the length of the full input sequence. In the table, the empirical coverage of length bins `1–10' and `11–20' overshoots the desired coverage to a lesser extent. In doing so, the average prediction-set size decreases significantly. Length bin `21–30', on the other hand, remains relatively unaffected by the calibration procedure, and length bins `31–40' and `40+' demonstrate increased coverages and prediction set sizes. The calibration procedure also flipped the empirical coverage of bin `40+' from being below the desired coverage to above the desired coverage rate.

Importantly, the distribution of non-conformity scores differs between each sequence length. Therefore, if perfectly calibrated responses were desired for all sentence lengths equally, calibration would have to be performed with bins of size one instead of size ten, as displayed in this table. By combining sentences of similar lengths into bins, we ensure that each bin is properly calibrated, accepting that there will be some slight miscalibration within groups between sequences of differing lengths. Ultimately, the bin granularity permitted by the calibration technique is limited by the amount of calibration data that is accessible, with larger datasets allowing for more granular bins. 

\begin{table*}[t]
\centering
\scriptsize
\resizebox{\textwidth}{!}{%
\begin{tabular}{llcccc}
\toprule
\textbf{Language} & \textbf{Desired Cov.} 
& \multicolumn{2}{c}{\textbf{No Calibration}} 
& \multicolumn{2}{c}{\textbf{Language Calibration}} \\
\cmidrule(lr){3-4}\cmidrule(lr){5-6}
 &  & \textbf{Emp. - Desired Cov.} & Average Set Size & \textbf{Emp. - Desired Cov.}& Average Set Size \\
\midrule
Dutch       & 0.8   & 0.1276  & 0.9816 & 0.0483 & \textit{0.8606} \\
Dutch       & 0.9   & 0.0668  & 1.3312 & 0.0338 & \textit{1.0571} \\
Dutch       & 0.95  & 0.0389  & 1.8212 & 0.0152 & \textit{1.2703} \\
Dutch       & 0.975 & 0.0175  & 2.0974 & 0.0094 & \textbf{1.6931} \\
\midrule
English     & 0.8   & \textbf{-0.0575} & 0.8085 & 0.0545 & \textbf{\textit{1.2216}} \\
English     & 0.9   & \textbf{-0.0239} & 1.2526 & 0.0260 & \textbf{\textit{1.7398}} \\
English     & 0.95  & \textbf{-0.0118} & 1.8617 & 0.0124 & \textbf{\textit{3.3362}} \\
English     & 0.975 & \textbf{-0.0128} & 2.4153 & 0.0034 & \textbf{\textit{23.8822}} \\
\midrule
French      & 0.8   & 0.0382  & 0.8958 & 0.0531 & 0.9579 \\
French      & 0.9   & 0.0231  & 1.2866 & 0.0315 & 1.3455 \\
French      & 0.95  & 0.0156  & 1.8462 & 0.0173 & 1.8648 \\
French      & 0.975 & 0.0043  & 2.2375 & 0.0066 & 2.8115 \\
\midrule
German      & 0.8   & 0.0972  & 0.9545 & 0.0522 & 0.8889 \\
German      & 0.9   & 0.0591  & 1.3290 & 0.0313 & \textbf{1.0926} \\
German      & 0.95  & 0.0354  & 1.8307 & 0.0175 & \textbf{1.3981} \\
German      & 0.975 & 0.0181  & 2.1167 & 0.0103 & \textbf{1.7326} \\
\midrule
Italian     & 0.8   & 0.0463  & 0.9027 & 0.0450 & 0.9295 \\
Italian     & 0.9   & 0.0306  & 1.3019 & 0.0256 & 1.2590 \\
Italian     & 0.95  & 0.0184  & 1.8390 & 0.0148 & \textbf{1.7171} \\
Italian     & 0.975 & 0.0029  & 2.2155 & 0.0046 & \textbf{\textit{3.0200}} \\
\midrule
Polish      & 0.8   & \textbf{-0.0118} & 0.8419 & 0.0548 & \textbf{\textit{1.0896}} \\
Polish      & 0.9   & 0.0057  & 1.2633 & 0.0270 & \textbf{\textit{1.5076}} \\
Polish      & 0.95  & 0.0003  & 1.8344 & 0.0108 & \textbf{\textit{2.5685}} \\
Polish      & 0.975 & \textbf{-0.0104} & 2.3595 & 0.0013 & \textbf{\textit{8.5080}} \\
\midrule
Portuguese  & 0.8   & 0.0697  & 0.9325 & 0.0479 & 0.9042 \\
Portuguese  & 0.9   & 0.0331  & 1.3010 & 0.0301 & 1.2724 \\
Portuguese  & 0.95  & 0.0136  & 1.8326 & 0.0132 & 1.8205 \\
Portuguese  & 0.975 & 0.0031  & 2.1758 & 0.0056 & \textbf{\textit{2.8365}} \\
\midrule
Russian     & 0.8   & \textbf{-0.1159} & 0.7441 & 0.0482 & \textbf{\textit{1.3796}} \\
Russian     & 0.9   & \textbf{-0.0507} & 1.2234 & 0.0293 & \textbf{\textit{2.1011}} \\
Russian     & 0.95  & \textbf{-0.0269} & 1.8571 & 0.0110 & \textbf{\textit{6.6575}} \\
Russian     & 0.975 & \textbf{-0.0168} & 2.5331 & 0.0016 & \textbf{\textit{14.7522}} \\
\midrule
Spanish     & 0.8   & 0.1446  & 1.0013 & 0.0580 & \textbf{0.8640} \\
Spanish     & 0.9   & 0.0739  & 1.3325 & 0.0258 & \textbf{0.9653} \\
Spanish     & 0.95  & 0.0419  & 1.8133 & 0.0131 & \textbf{1.1570} \\
Spanish     & 0.975 & 0.0190  & 2.0538 & 0.0073 & \textbf{1.5020} \\
\midrule
\end{tabular}
}
\caption{Model calibration for all languages in the WikiNEuRal benchmark. Coverage columns are bolded if the empirical coverage is less than the desired coverage. Set size columns are bolded if there is a decrease of greater than .1 in average set size and bolded + italicized if there is an increase of greater than .1 in average set size.}
\label{tab: language calibration results_full}
\end{table*}

Similarly to the results on sequence-length calibration, Table~\ref{tab: language calibration results_full} presents the results for language calibration on the multilingual WikiNEuRal benchmark. As observed, the initial under-coverage of the English, Russian, and Polish languages was corrected via language-specific calibration. In doing so, the average prediction-set sizes of all these methods were increased, sometimes drastically. For example, the average prediction set size of English sentences at the 97.5\% coverage level changed from 2.4153 to 23.882 predictions. As a result, the empirical coverage changed from 96.22\% to 97.84\%. Many languages, such as German, had their empirical coverages become closer to the desired coverage, and in doing so, the average prediction set size shrank via language calibration. The overall trend in coverage, increasing/shrinking, is not constant for each language as a function of the desired coverage level. For example, the Italian language saw a slight decrease in average prediction set size for 95\% desired coverage while also experiencing an increase in average prediction set size for a desired coverage of 97.5\%.

\begin{table*}[t]
\centering
\scriptsize

\begin{tabular}{lrrrrrrrrrrrr}
\toprule
\multicolumn{11}{c}{\textbf{full-sequence}} \\
\midrule
\textbf{Desired Coverage}  
& \multicolumn{2}{c}{\textbf{1 Entity}}
& \multicolumn{2}{c}{\textbf{2 Entities}}
& \multicolumn{2}{c}{\textbf{3 Entities}}
& \multicolumn{2}{c}{\textbf{4 Entities}}
& \multicolumn{2}{c}{\textbf{5 Entities}} \\
\cmidrule(lr){2-3}\cmidrule(lr){4-5}\cmidrule(lr){6-7}\cmidrule(lr){8-9}\cmidrule(lr){10-11}
& \textbf{Cov.} & \textbf{Set Size} & \textbf{Cov} & \textbf{Set Size} & \textbf{Cov} & \textbf{Set Size} & \textbf{Cov} & \textbf{Set Size} & \textbf{Cov} & \textbf{Set Size} \\
\midrule
0.8   & 0.9026 & 1.0103 & 0.8310 & 0.9916 & \textbf{0.7891} & 0.9895 & \textbf{0.7911} & 1.0122 & \textbf{0.7982} & 1.0408 \\
0.9   & 0.9493 & 1.2279 & 0.9015 & 1.2744 & \textbf{0.8621} & 1.3029 & \textbf{0.8600} & 1.3342 & \textbf{0.8592} & 1.3595 \\
0.95  & 0.9731 & 1.4527 & \textbf{0.9411} & 1.6003 & \textbf{0.9102} & 1.7559 & \textbf{0.9019} & 1.8736 & \textbf{0.9068} & 2.0519 \\
0.975 & \textbf{0.9845} & 1.7156 & \textbf{0.9628} & 2.1007 & \textbf{0.9392} & 2.6243 & \textbf{0.9299} & 3.2834 & \textbf{0.9364} & 3.6174 \\

\midrule
\multicolumn{11}{c}{\textbf{Integrated without Šidák Correction}} \\
\midrule
\textbf{Desired Coverage}  
& \multicolumn{2}{c}{\textbf{1 Entity}}
& \multicolumn{2}{c}{\textbf{2 Entities}}
& \multicolumn{2}{c}{\textbf{3 Entities}}
& \multicolumn{2}{c}{\textbf{4 Entities}}
& \multicolumn{2}{c}{\textbf{5 Entities}} \\
\cmidrule(lr){2-3}\cmidrule(lr){4-5}\cmidrule(lr){6-7}\cmidrule(lr){8-9}\cmidrule(lr){10-11}
& \textbf{Cov.} & \textbf{Set Size} & \textbf{Cov} & \textbf{Set Size} & \textbf{Cov} & \textbf{Set Size} & \textbf{Cov} & \textbf{Set Size} & \textbf{Cov} & \textbf{Set Size} \\
\midrule
0.8   & 0.8817 & 1.1239 & \textbf{0.7050} & 1.9253 & \textbf{0.6075} & 2.1773 & \textbf{0.5824} & 2.4847 & \textbf{0.5709} & 2.8040 \\
0.9   & 0.9456 & 2.7796 & \textbf{0.8476} & 4.0483 & \textbf{0.7785} & 4.6224 & \textbf{0.7602} & 5.1073 & \textbf{0.7512} & 5.4937 \\
0.95  & 0.9766 & 7.5755 & \textbf{0.9293} & 9.2887 & \textbf{0.8859} & 10.1569 & \textbf{0.8748} & 10.9781 & \textbf{0.8682} & 11.7518 \\
0.975 & 0.9909 & 13.9922 & \textbf{0.9694} & 16.4612 & \textbf{0.9471} & 17.9809 & \textbf{0.9373} & 19.2119 & \textbf{0.9289} & 20.4667 \\

\midrule
\multicolumn{11}{c}{\textbf{Integrated with Šidák Correction}} \\
\midrule
\textbf{Desired Coverage}  
& \multicolumn{2}{c}{\textbf{1 Entity}}
& \multicolumn{2}{c}{\textbf{2 Entities}}
& \multicolumn{2}{c}{\textbf{3 Entities}}
& \multicolumn{2}{c}{\textbf{4 Entities}}
& \multicolumn{2}{c}{\textbf{5 Entities}} \\
\cmidrule(lr){2-3}\cmidrule(lr){4-5}\cmidrule(lr){6-7}\cmidrule(lr){8-9}\cmidrule(lr){10-11}
& \textbf{Cov.} & \textbf{Set Size} & \textbf{Cov} & \textbf{Set Size} & \textbf{Cov} & \textbf{Set Size} & \textbf{Cov} & \textbf{Set Size} & \textbf{Cov} & \textbf{Set Size} \\
\midrule
0.8   & 0.8833 & 1.1716 & 0.8442 & 4.0042 & 0.8387 & 7.1049 & 0.8693 & 10.7663 & 0.8872 & 14.3301 \\
0.9   & 0.9465 & 2.8579 & 0.9227 & 9.2096 & 0.9236 & 15.6326 & 0.9335 & 19.0864 & 0.9374 & 23.1792 \\
0.95  & 0.9773 & 7.6536 & 0.9623 & 16.3684 & 0.9635 & 24.0292 & 0.9629 & 26.1613 & 0.9565 & 28.6008 \\
0.975 & 0.9912 & 14.0604 & 0.9814 & 21.8675 & 0.9755 & 26.5465 & \textbf{0.9699} & 28.8020 & \textbf{0.9643} & 31.2569 \\
\bottomrule
\end{tabular}

\caption{Model calibration per number of named entities within a sentence for three methods, Full-Sequence, Integrated without a Šidák family-wise error control correction, and Integrated with said correction. All prediction sets are calculated on the WikiNEuRal benchmark utilizing the Babelscape model and \textbf{nc1} non-conformity score. Coverage columns are bolded if the empirical coverage is less than the desired coverage. }
\label{tab: n_ent_results_full}
\end{table*}

Finally, we present Table~\ref{tab: n_ent_results_full}, a comparison of how the full-sequence and integrated methods handle model calibration as a function of the number of entities in a sentence. That table shows how the default full-sequence method and baseline integrated method fail to properly calibrate prediction-set coverage for inputs containing multiple entities. By implementing the Šidák family-wise error correction, we can observe how the integrated method produces valid prediction sets for all but 97.5\% desired coverage at 4 and 5 entities. We note that, theoretically, the full-sequence prediction sets could be partitioned by the estimated number of entities in the input. Once partitioned, each group could be calibrated, and we would expect there to be an improvement in the coverage levels for each group. However, at that point, the full-sequence prediction sets would be grouped by language, sentence length, and number of entities: therefore, creating even smaller sample sizes for calibration on each bin combination. On the other hand, the integrated method does not group observations into separate bins based on the number of estimated entities; instead, the coverage of each subsequence prediction set is dynamically increased as the number of estimated entities increases, leading to an overall higher combined coverage when the subsequence prediction sets are merged.

This subsection did not display the results of entity-class calibration on full-sequence, subsequence, or integrated prediction sets. Instead, those results are reported in Table~\ref{tab:hybrid comparisons_full} in Section~\ref{subsec:class cond}.

\subsection{Supplement for Section 7.4: Comparing Index-based Methods}

As mentioned in previous sections, we compared the efficiency of all proposed non-conformity scores for full-sequence, subsequence, and integrated prediction sets. We begin by presenting Table ~\ref{tab:index comparisons_full}, which shows the efficiency of \textbf{nc1}, \textbf{nc2}, and \textbf{nc3} when calculated with the Babelscape model on the multilingual WikiNEuRal benchmark. 

\begin{table*}[t]
\centering

\resizebox{0.8\textwidth}{!}{%
\begin{tabular}{c|cc|cc|cc}
\toprule
\multicolumn{7}{c}{\textbf{full-sequence}} \\
\midrule
\multirow{2}{*}{\textbf{Desired}} 
& \multicolumn{2}{c|}{\textbf{nc1}} 
& \multicolumn{2}{c|}{\textbf{nc2}} 
& \multicolumn{2}{c|}{\textbf{nc3}} \\
\cmidrule(r){2-7}
\textbf{Coverage} & Cov. & Size & Cov. & Size & Cov. & Size \\
\midrule
0.8 & 0.8502 & 0.9090 & 0.9946 & 27.1105 & 0.7844 & 0.8428 \\
0.9 & 0.9276 & 1.2950 & 0.9947 & 36.8596 & 0.8885 & 0.9546 \\
0.95 & 0.9679 & 1.8400 & 0.9948 & 45.4994 & 0.9438 & 1.3345 \\
0.975 & 0.9806 & 2.2226 & 0.9948 & 51.7461 & 0.9742 & 2.5008 \\
\midrule

\multicolumn{7}{c}{\textbf{Subsequence}} \\
\midrule
\multirow{2}{*}{\textbf{Desired}} 
& \multicolumn{2}{c|}{\textbf{nc1}} 
& \multicolumn{2}{c|}{\textbf{nc2}} 
& \multicolumn{2}{c|}{\textbf{nc3}} \\
\cmidrule(r){2-7}
\textbf{Coverage} & Cov. & Size & Cov. & Size & Cov. & Size \\
\midrule
0.8 & 0.959 & 0.9921 & 0.9986 & 1.4225 & 0.8455 & 0.8518 \\
0.9 & 0.9787 & 1.0894 & 0.9991 & 1.6064 & 0.9465 & 0.9714 \\
0.95 & 0.9895 & 1.1815 & 0.9994 & 1.9008 & 0.9812 & 1.1812 \\
0.975 & 0.9951 & 1.2563 & 0.9995 & 2.2071 & 0.9939 & 1.4072 \\
\midrule
\multicolumn{7}{c}{\textbf{Integrated}} \\
\midrule
\multirow{2}{*}{\textbf{Desired}} 
& \multicolumn{2}{c|}{\textbf{nc1}} 
& \multicolumn{2}{c|}{\textbf{nc2}} 
& \multicolumn{2}{c|}{\textbf{nc3}} \\
\cmidrule(r){2-7}
\textbf{Coverage} & Cov. & Size & Cov. & Size & Cov. & Size \\
\midrule
0.8 & 0.9423 & 4.9601 & 0.9951 & 23.6458 & 0.7542 & 13.1623 \\
0.9 & 0.9664 & 8.7681 & 0.995 & 26.699 & 0.9099 & 17.7435 \\
0.95 & 0.9796 & 13.1685 & 0.994 & 34.0406 & 0.9665 & 20.8824 \\
0.975 & 0.9874 & 16.7412 & 0.9945 & 40.9204 & 0.9857 & 22.9789 \\
\bottomrule
\end{tabular}
}
\caption{Average coverage and set size for \textbf{nc1}, \textbf{nc2}, and \textbf{nc3} as evaluated on the multi-lingual WikiNEuRal benchmark with the Babelscape model } 
\label{tab:index comparisons_full}
\end{table*}

In Table \ref{tab:index comparisons_full}, we find that \textbf{nc1} performs significantly better when compared to both \textbf{nc2} and \textbf{nc3} for all high levels of coverage above 95\%. Meanwhile \textbf{nc3} constructs small prediction sets for coverage levels below 95\% while also being slightly under-calibrated due to the adaptive conformal prediction procedure. We expect that the \textbf{nc3} non-conformity score will perform better than \textbf{nc1} and \textbf{nc2} for any application that desires a prediction set with coverage less than its top-1 response accuracy. For all prediction types, \textbf{nc2} performs significantly worse than both \textbf{nc1} and \textbf{nc3}, producing extremely large prediction sets while significantly overstepping the desired coverage levels by a large margin.

\begin{table*}[t]
\centering
\resizebox{\textwidth}{!}{%
\begin{tabular}{c|cc|cc|cc|cc|cc|cc}
 & \multicolumn{4}{c}{full-sequence} & \multicolumn{4}{c}{Subsequence}& \multicolumn{4}{c}{Integrated} \\
\toprule
\multirow{2}{*}{\textbf{Desired}} & \multicolumn{2}{c|}{\textbf{Naive + nc1}} 
& \multicolumn{2}{c|}{\textbf{Naive + nc2}}
& \multicolumn{2}{c|}{\textbf{Naive + nc1}} 
& \multicolumn{2}{c|}{\textbf{Naive + nc2}}
& \multicolumn{2}{c|}{\textbf{Naive + nc1}} 
& \multicolumn{2}{c}{\textbf{Naive + nc2}}\\\\
\cmidrule(r){2-13}
\textbf{Coverage} 
 & Cov. & Size & Cov. & Size & Cov. & Size & Cov. & Size & Cov. & Size & Cov. & Size  \\
\midrule
0.8 & 0.8047 & 0.8247 & 0.8183 & 0.9106 & 0.9058 & 0.7807 & 0.984 & 0.3057 & 0.8563 & 4.9385 & 0.9766 & 23.2816 \\
0.9 & 0.9032 & 0.9917 & 0.9095 & 1.3102 & 0.9548 & 0.9702 & 0.989 & 0.8589 & 0.934 & 11.3335 & 0.9833 & 22.7566 \\
0.95 & 0.9504 & 1.3626 & 0.9569 & 2.7537 & 0.98 & 1.1585 & 0.9945 & 1.2402 & 0.965 & 16.7349 & 0.9893 & 26.1235 \\
0.975 & 0.9744 & 3.6263 & 0.9783 & 7.0064 & 0.9906 & 1.3054 & 0.9974 & 1.5345 & 0.9769 & 20.5522 & 0.9911 & 28.2138 \\

\midrule
\multirow{2}{*}{\textbf{Desired}} & \multicolumn{2}{c|}{\textbf{RAPS + nc1}} 
& \multicolumn{2}{c|}{\textbf{RAPS + nc2}} 
& \multicolumn{2}{c|}{\textbf{RAPS + nc1}} 
& \multicolumn{2}{c|}{\textbf{RAPS + nc2}}
& \multicolumn{2}{c|}{\textbf{RAPS + nc1}} 
& \multicolumn{2}{c}{\textbf{RAPS + nc2}}\\
\cmidrule(r){2-13}
\textbf{Coverage} 
 & Cov. & Size & Cov. & Size & Cov. & Size & Cov. & Size & Cov. & Size & Cov. & Size  \\
\midrule
0.8 & 0.7993 & 0.8183 & 0.8 & 1.1055 & 0.8786 & 0.8102 & 0.9381 & 0.9402 & 0.8033 & 1.4646 & 0.9148 & 3.1428 \\
0.9 & 0.8992 & 1.0017 & 0.8998 & 1.313 & 0.9391 & 0.9419 & 0.9611 & 0.9873 & 0.9108 & 1.7832 & 0.95 & 3.4066 \\
0.95 & 0.9496 & 1.6049 & 0.9495 & 2.1582 & 0.9698 & 1.0443 & 0.9762 & 1.0552 & 0.9504 & 2.6704 & 0.9641 & 4.9603 \\
0.975 & 0.9743 & 3.4693 & 0.9742 & 5.6624 & 0.9848 & 1.1486 & 0.9845 & 1.163 & 0.9704 & 3.6926 & 0.9712 & 6.5276 \\

\midrule
\multirow{2}{*}{\textbf{Desired}} & \multicolumn{2}{c|}{\textbf{Cond. + nc1}} 
& \multicolumn{2}{c}{\textbf{Cond. + nc2.}}
& \multicolumn{2}{c|}{\textbf{Cond. + nc1}} 
& \multicolumn{2}{c|}{\textbf{Cond. + nc2.}}
& \multicolumn{2}{c|}{\textbf{Cond. + nc1}} 
& \multicolumn{2}{c}{\textbf{Cond. + nc2.}}\\
\cmidrule(r){2-13}
\textbf{Coverage} 
 & Cov. & Size & Cov. & Size & Cov. & Size & Cov. & Size & Cov. & Size & Cov. & Size  \\
\midrule
0.8 & 0.804 & 0.8245 & 0.8137 & 0.9163 & 0.9172 & 0.8747 & 0.9668 & 0.9826 & 0.868 & 2.2321 & 0.9562 & 2.3067 \\
0.9 & 0.903 & 0.9971 & 0.9062 & 1.4494 & 0.9601 & 0.992 & 0.9763 & 1.036 & 0.9404 & 3.1087 & 0.969 & 3.2596 \\
0.95 & 0.9518 & 1.42 & 0.9546 & 3.5607 & 0.9817 & 1.1221 & 0.9872 & 1.2178 & 0.969 & 5.9108 & 0.9799 & 6.6276 \\
0.975 & 0.976 & 3.9008 & 0.9769 & 9.4974 & 0.9911 & 1.2114 & 0.9937 & 1.4048 & 0.9789 & 8.2014 & 0.9848 & 10.6308 \\

\bottomrule
\end{tabular}
}
\caption{Average coverage and set size for the naive, RAPS, and conditional methods when utilizing either \textbf{nc1} or \textbf{nc2} as evaluated on the multilingual WikiNEuRal benchmark with the Babelscape model across all types of prediction sets}
\label{tab:indexcomparisons2_full}
\end{table*}

 Table~\ref{tab:indexcomparisons2_full} displays the efficiency and coverage of the six index-based integrated non-conformity scores when evaluated on full-sequence, subsequence, and integrated prediction sets for the WikiNEuRal Benchmark with the Babelscape model. As a general trend, we can observe that all integrated methods, when utilizing \textbf{nc1} perform better than their \textbf{nc2} counterparts in almost all situations. The RAPS procedure also performs slightly better than the conditional and naive methods for most situations except for full-sequence prediction sets at the 90\% and 95\% coverage levels. We believe that part of the disparity between the conditional and RAPS-based prediction sets is the granularity of the grid search performed to estimate the optimal values of $\alpha$ and $\beta$ for the conditional prediction sets. RAPS prediction sets also perform a grid search over the hyperparameter term of $\lambda$, but it is uncertain how much a sparse grid affects the efficiency of the RAPS non-conformity score when compared to the effect of performing a grid search for the conditional non-conformity scores.

Although the Naive methodology constructs valid prediction sets with coverage levels for most scenarios. The integrated prediction sets formed while using the naive methodology and \textbf{nc1} achieve a larger average prediction set size compared to the baseline \textbf{nc1} method presented in Table~\ref{tab:index comparisons_full}. At the same time, the coverage of those integrated prediction sets is closer to the desired coverage level, indicating that some subsections of inputs (e.g., sequences with many entities) may have significantly larger prediction sets while other subsections produce smaller prediction sets on average.

A large motivation for the development of an integrated prediction set methodology was to ensure entity-class conditional coverage of full-sequence prediction sets regardless of the type or number of NER entities within a sentence. Therefore, Table~\ref{tab:hybrid comparisons_full} presents the average prediction set size and coverage for the full-sequence, subsequence, and integrated methods, broken down by class type. These results use the language and length-controlled \textbf{nc1} and $\mathbf{nc1_{ent}}$ non-conformity scores with the Babelscape model on the multilingual WikiNEuRal benchmark. As expected, the full-sequence prediction sets do not achieve class-conditional coverage for the miscellaneous entity type but yield significantly smaller set sizes compared to the integrated level. Importantly, because the integrated prediction sets are well calibrated for each class, their resulting set size is more affected by poorly performing classes (e.g. Miscellaneous). When utilizing the integrated prediction methodology, one poorly performing class will cause a significant increase in prediction set size.  Notably, all subsequence prediction sets, regardless of class type, produce compact prediction sets due to the output space containing only five possible labels. In comparison, the full-sequence and integrated prediction sets may contain up to $m^{|\mathcal{L}|}$ possible sequences, where $m$ is the sentence length and $|\mathcal{L}|$ is the number of word-level labels. 

\begin{table*}[t]
\centering
\resizebox{\textwidth}{!}{%
\begin{tabular}{c|cc|cc|cc|cc}
\multicolumn{9}{c}{\textbf{Sentence Level}} \\
\hline
\textbf{Desired Coverage} & \multicolumn{2}{c|}{\textbf{PER}} & \multicolumn{2}{c|}{\textbf{LOC}} & \multicolumn{2}{c|}{\textbf{ORG}} & \multicolumn{2}{c}{\textbf{MISC}} \\
& Coverage & Set Size & Coverage & Set Size & Coverage & Set Size & Coverage & Set Size \\
\hline
0.8 & 0.8554 & 0.9129 & 0.8373 & 0.8964 & 0.8168 & 0.8774 & \textbf{0.7332} & 0.8037 \\
0.9 & 0.9291 & 1.2876 & 0.9194 & 1.2812 & 0.9058 & 1.2695 & \textbf{0.8592} & 1.2392 \\
0.95 & 0.9685 & 1.8325 & 0.9637 & 1.8347 & 0.9572 & 1.8417 & \textbf{0.9303} & 1.861 \\
0.975 & 0.9818 & 2.2374 & 0.9788 & 2.2655 & 0.9749 & 2.3215 & \textbf{0.9551} & 2.4555 \\
\end{tabular}}

\vspace{0.5cm}

\resizebox{\textwidth}{!}{%
\begin{tabular}{c|cc|cc|cc|cc}
\multicolumn{9}{c}{\textbf{Entity Level}} \\
\hline
\textbf{Desired Coverage} & \multicolumn{2}{c|}{\textbf{PER}} & \multicolumn{2}{c|}{\textbf{LOC}} & \multicolumn{2}{c|}{\textbf{ORG}} & \multicolumn{2}{c}{\textbf{MISC}} \\
& Coverage & Set Size & Coverage & Set Size & Coverage & Set Size & Coverage & Set Size \\
\hline
0.8 & 0.9729 & 0.9853 & 0.9629 & 1.0018 & 0.9474 & 0.9895 & 0.946 & 0.9266 \\
0.9 & 0.9837 & 1.0471 & 0.9785 & 1.1149 & 0.9727 & 1.1305 & 0.9787 & 1.0333 \\
0.95 & 0.9895 & 1.0994 & 0.9884 & 1.2159 & 0.9882 & 1.2669 & 0.9917 & 1.1198 \\
0.975 & 0.9946 & 1.1466 & 0.9951 & 1.2901 & 0.9957 & 1.3668 & 0.9949 & 1.2061 \\
\end{tabular}}

\vspace{0.5cm}

\resizebox{\textwidth}{!}{%
\begin{tabular}{c|cc|cc|cc|cc}
\multicolumn{9}{c}{\textbf{Integrated}} \\
\hline
\textbf{Desired Coverage} & \multicolumn{2}{c|}{\textbf{PER}} & \multicolumn{2}{c|}{\textbf{LOC}} & \multicolumn{2}{c|}{\textbf{ORG}} & \multicolumn{2}{c}{\textbf{MISC}} \\
& Coverage & Set Size & Coverage & Set Size & Coverage & Set Size & Coverage & Set Size \\
\hline
0.8 & 0.9451 & 5.9275 & 0.9305 & 6.1636 & 0.9221 & 6.0232 & 0.9149 & 7.8172 \\
0.9 & 0.9677 & 10.5229 & 0.9579 & 10.5338 & 0.9561 & 10.9653 & 0.9554 & 13.2076 \\
0.95 & 0.9798 & 15.415 & 0.9756 & 15.285 & 0.9785 & 16.2023 & 0.9732 & 18.2391 \\
0.975 & 0.988 & 19.6862 & 0.9877 & 19.576 & 0.9884 & 20.7705 & 0.9785 & 21.9911 \\
\end{tabular}}
\caption{Average coverage and prediction set size by conformal prediction methodology with the \textbf{nc1} non-conformity score. Non-conformity scores were calculated using the Babelscape model on the multilingual WikiNEuRal benchmark dataset. Empirical coverages below the desired coverage threshold are bolded.  \label{tab:hybrid comparisons_full}}
\end{table*}

\subsection{Supplement for Section 7.5: Benchmark Comparisons}

Table~\ref{tab:entity_performance_full} presents class-specific results for the subsequence prediction set method. In this approach, prediction sets are generated for all identified spans, with each set containing up to five possible labels—one for each entity class and an additional label for non-entity spans. In the table, results for the Conll\_reduced benchmark are reported as the 'General-Entity' class. As we can see, the average set size is nearly one for all models. Despite this, the models are not over-generating NER entities, as we can see from Table~\ref{tab:non-entity} that the average set size of non-entity spans is near zero, indicating that the conformal prediction procedure is able to differentiate between valid and erroneous entity spans. 

For the CoNLL and WikiNEuRal benchmarks, the subsequence prediction sets consistently achieve coverage above the desired threshold while maintaining an average prediction set size of approximately 1.1 to 2.5 NER types per identified entity span. Consistent with the trends observed in Table~\ref{tab:sentence results}, the Babelscape model continues to outperform its counterparts on the WikiNEuRal benchmark while under-performing on the CoNLL benchmark compared to all other methods. Dslim produced the smallest prediction sets for the person entity category on CoNLL, while the Jean model produced slightly smaller prediction sets for all other class categories.

Across all models, performance is notably worse when identifying organizations and locations in the WikiNEuRal and CoNLL benchmark compared to persons or miscellaneous entities. This suggests that the task of recognizing organizations may involve more challenging decision boundaries, leading to greater confusion with other entity types. Additionally, all models—except Babelscape—show reduced performance on the miscellaneous entity class in the WikiNEuRal benchmark compared to person. This decline may be attributed to the inherently ambiguous and heterogeneous nature of the ‘miscellaneous’ class, which varies significantly across the training datasets used by the respective models (WikiNEuRal, CoNLL, or OntoNotes in the case of TNER). TNER, relative to its full-sequence performance, produces good results for both CoNLL and WikiNEuRal benchmarks across all class types. This may indicate that the poor performance of TNER in the full-sequence task is due to its hallucination of false positives, as indicated by Table~\ref{tab:non-entity}. Because the full-sequence prediction task requires all labels within a sentence to be correct, any false positive entities will lead to an overall incorrect result, even if all true positives are identified correctly.

\begin{table*}[t]
\centering

\begin{center}
   \textbf{ Person }
\end{center}

\resizebox{\textwidth}{!}{%
\begin{tabular}{llllll}
\toprule
Benchmark & Metric & Babelscape & Dslim & Jean & Tner \\
\midrule
\multirow{2}{*}{CoNLL}
& Coverage & 0.982 $\pm$ 0.002 & 0.986 $\pm$ 0.004 & 0.985 $\pm$ 0.002 & 0.978 $\pm$ 0.004 \\
& Set Size & 1.61 $\pm$ 0.024 & \textbf{1.096 $\pm$ 0.017} & 1.139 $\pm$ 0.011 & 1.355 $\pm$ 0.017 \\
\midrule
\multirow{2}{*}{WikiNeuRal\_en}
& Coverage & 0.978 $\pm$ 0.004 & 0.978 $\pm$ 0.002 & 0.980 $\pm$ 0.004 & 0.976 $\pm$ 0.003 \\
& Set Size & \textbf{1.178 $\pm$ 0.012} & 1.285 $\pm$ 0.019 & 1.202 $\pm$ 0.003 & 1.275 $\pm$ 0.010 \\
\bottomrule
\end{tabular}
}

\begin{center}
   \textbf{ Location}
\end{center}

\resizebox{\textwidth}{!}{%
\begin{tabular}{llllll}
\toprule
Benchmark & Metric & Babelscape & Dslim & Jean & Tner \\
\midrule
\multirow{2}{*}{CoNLL}
& Coverage & 0.980 $\pm$ 0.007 & 0.985 $\pm$ 0.006 & 0.982 $\pm$ 0.006 & 0.984 $\pm$ 0.006 \\
& Set Size & 2.459 $\pm$ 0.045 & 1.941 $\pm$ 0.023 & \textbf{1.482 $\pm$ 0.011} & 1.992 $\pm$ 0.019 \\
\midrule
\multirow{2}{*}{WikiNeuRal\_en}
& Coverage & 0.982 $\pm$ 0.004 & 0.985 $\pm$ 0.003 & 0.986 $\pm$ 0.004 & 0.984 $\pm$ 0.002 \\
& Set Size & \textbf{1.431 $\pm$ 0.010} & 1.475 $\pm$ 0.016 & 1.548 $\pm$ 0.009 & 1.535 $\pm$ 0.015 \\
\bottomrule
\end{tabular}
}

\begin{center}
   \textbf{ Organization }
\end{center}

\resizebox{\textwidth}{!}{%
\begin{tabular}{llllll}
\toprule
Benchmark & Metric & Babelscape & Dslim & Jean & Tner \\
\midrule
\multirow{2}{*}{CoNLL}
& Coverage & 0.978 $\pm$ 0.005 & 0.978 $\pm$ 0.001 & 0.984 $\pm$ 0.005 & 0.984 $\pm$ 0.002 \\
& Set Size & 2.06 $\pm$ 0.013 & 1.162 $\pm$ 0.028 & \textbf{1.069 $\pm$ 0.007} & 1.871 $\pm$ 0.034 \\
\midrule
\multirow{2}{*}{WikiNeuRal\_en}
& Coverage & 0.984 $\pm$ 0.004 & 0.982 $\pm$ 0.005 & 0.980 $\pm$ 0.003 & 0.984 $\pm$ 0.003 \\
& Set Size &\textbf{ 1.344 $\pm$ 0.021} & 1.492 $\pm$ 0.012 & 1.474 $\pm$ 0.019 & 1.487 $\pm$ 0.014 \\
\bottomrule
\end{tabular}
}

\begin{center}
   \textbf{ Miscellaneous }
\end{center}

\resizebox{\textwidth}{!}{%
\begin{tabular}{llllll}
\toprule
Benchmark & Metric & Babelscape & Dslim & Jean & Tner \\
\midrule
\multirow{2}{*}{CoNLL}
& Coverage & 0.987 $\pm$ 0.013 & 0.976 $\pm$ 0.002 & 0.985 $\pm$ 0.005 & 0.976 $\pm$ 0.008 \\
& Set Size & 1.383 $\pm$ 0.014 & 1.076 $\pm$ 0.002 & \textbf{1.034 $\pm$ 0.009} & 1.316 $\pm$ 0.027 \\
\midrule
\multirow{2}{*}{WikiNeuRal\_en}
& Coverage & 0.994 $\pm$ 0.002 & 0.989 $\pm$ 0.001 & 0.993 $\pm$ 0.002 & 0.988 $\pm$ 0.001 \\
& Set Size & \textbf{1.156 $\pm$ 0.015} & 1.399 $\pm$ 0.013 & 1.381 $\pm$ 0.022 & 1.429 $\pm$ 0.005 \\
\bottomrule
\end{tabular}
}

\begin{center}
   \textbf{ General-Entiity }
\end{center}

\resizebox{\textwidth}{!}{%
\begin{tabular}{llllll}
\toprule
Benchmark & Metric & Babelscape & Dslim & Jean & Tner \\
\midrule
\multirow{2}{*}{CoNLL\_Reduced}
& Coverage & 0.994 $\pm$ 0.002 & 0.986 $\pm$ 0.001 & 0.988 $\pm$ 0.001 & 0.993 $\pm$ 0.001 \\
& Set Size & 0.992 $\pm$ 0.002 & \textbf{0.961 $\pm$ 0.003} & \textbf{0.964 $\pm$ 0.004} & 0.986 $\pm$ 0.002 \\
\bottomrule
\end{tabular}
}
\caption{Coverage and Set Size of subsequence prediction sets with 95\% confidence for each model and entity type in CoNLL and the English partition of WikiNEuRal.The best overall prediction set size is bolded in each benchmark.}
\label{tab:entity_performance_full}

\end{table*}

\end{document}